\documentclass[runningheads]{llncs}

\usepackage{eccv}

\usepackage{eccvabbrv}

\usepackage{graphicx}
\usepackage{booktabs}
\usepackage[table]{xcolor}
\usepackage{array}
\usepackage{tabularx}
\usepackage{makecell}
\usepackage{siunitx}
\usepackage{multirow,colortbl}
\usepackage{caption}

\usepackage[accsupp]{axessibility}  

\usepackage[pagebackref,breaklinks,colorlinks,citecolor=eccvblue]{hyperref}
\usepackage{hyperref}

\usepackage{orcidlink}

\usepackage{algorithmic}
\usepackage{algorithm}

\begin{document}

\title{Three Creates All: You Only Sample 3 Steps} 

\titlerunning{Three Creates All}


\author{Yuren Cai\inst{1} \and
Guangyi Wang\inst{1} \and
Zongqing Li\inst{1} \and 
Li Li\inst{2} \and
Zhihui Liu\inst{3} \and
Songzhi Su\inst{1}
}

\authorrunning{Cai et al.}

\institute{School of Informatics, Xiamen University \and
	School of Artificial Intelligence, Shenzhen Polytechnic University \and
    Algorithm Research Department, Truesight}

\maketitle

\begin{abstract} Diffusion models deliver high-fidelity generation but remain slow at inference time due to many sequential network evaluations. We find that standard timestep conditioning becomes a key bottleneck for few-step sampling. Motivated by layer-dependent denoising dynamics, we propose \textbf{Multi-layer Time Embedding Optimization (MTEO)}, which freeze the pretrained diffusion backbone and distill a small set of step-wise, layer-wise time embeddings from reference trajectories. MTEO is plug-and-play with existing ODE solvers, adds no inference-time overhead, and trains only a tiny fraction of parameters. Extensive experiments across diverse datasets and backbones show state-of-the-art performance in the few-step sampling and substantially narrow the gap between distillation-based and lightweight methods. Code will be available.  \end{abstract}

\section{Introduction}
\label{sec:intro}

Diffusion models~\cite{DDPM,DDPM2015,SMLD,SDE,SD} have become a leading paradigm for generative modeling, delivering strong results across a wide range of modalities, including image synthesis, audio/speech generation~\cite{speech}, and text-conditioned generation such as text-to-image~\cite{SD} and text-to-video~\cite{sora}.
A key reason behind their success is the simplicity and generality of learning a denoising function that reverses a gradual noising process.
However, this advantage comes with a well-known bottleneck at inference time: generating a single sample requires an iterative denoising procedure, which translates into many sequential function evaluations of a large neural network, limiting practical deployment and real-time applications.

A substantial body of work has aimed to accelerate diffusion sampling process.
\textbf{Training-free} approaches~\cite{DDIM,DPM,DPM++,DPM-v3,unipc,deis&ipndm,pndm,PFdiff,OptimizedTimeSteps,Alignyoursteps}
improve the numerical solver and/or the sampling schedule without retraining the diffusion model, \eg, by using more efficient non-Markovian trajectories~\cite{DDIM}, dedicated higher-order solvers for diffusion ODEs~\cite{DPM,DPM++,DPM-v3,unipc,deis&ipndm,pndm}, or optimized timestep schedules~\cite{OptimizedTimeSteps,Alignyoursteps}.
While these methods introduce little to no training cost, their sample quality typically degrades rapidly in the extremely few-step regime (\eg, 3--5 NFE), where large step sizes amplify discretization errors.

\textbf{Training-based} acceleration methods mitigate this issue by introducing an additional training stage tailored to few-step inference.
Existing works can be roughly grouped into:
(i) \emph{lightweight training-based techniques}~\cite{AMED,AutoDiffusion,Trajectory,afs,timestepAligner,LD3,PAS},
which make minimal modifications (\eg, plugin predictors, schedule/search refinements, or auxiliary parameterizations) and often achieve strong performance around $\sim$5 NFE; and
(ii) \emph{distillation-based approaches}~\cite{progressiveDistillation,GuidedPD,cm,ctm,ect,sfd,dmd},
which can reach high-fidelity generation at 3--4 NFE (or even one-step in some settings) but usually require substantially heavier training and updating a large fraction of model parameters.
Despite this progress, an important gap remains: achieving \emph{distillation-level} few-step quality in the 3--5 NFE regime without paying distillation-level training cost.

In this paper, we revisit a component that is ubiquitous yet often treated as fixed during sampling: \emph{time conditioning}.
Modern diffusion backbones condition each denoising step through a timestep embedding, which is injected into network blocks to control feature-wise modulation (e.g., FiLM/AdaLN).
By carefully analyzing the full pathway from the scalar time variable to layer-wise feature modulation (see~\cref{sec:background}), we identify three coupled limitations that become pronounced under few-step sampling:
(1) the conventional single time-variable design is misaligned with the effective optimal update time in large-step solvers (see~\cref{subsec:inefficiency});
(2) different network layers exhibit distinct feature trajectories and thus prefer different effective time conditioning (see~\cref{subsec:feature_traj});
and (3) the standard timestep embedding has limited expressive degrees of freedom, which can severely constrain the modulation capacity of FiLM (see~\cref{subsec:vartoemb}).

To address these issues, we propose \textbf{Multi-layer Time Embedding Optimization (MTEO)}, a training-based yet \emph{lightweight} framework for fast diffusion sampling.
Instead of retraining (or distilling) the entire diffusion model, MTEO freezes the pretrained backbone and optimizes only a small set of \emph{step-wise, layer-wise} time embeddings via trajectory distillation (see~\cref{subsec:train}).
This design (i) decouples time conditioning across layers, (ii) unlocks richer layer-wise feature modulation, and (iii) introduces \emph{no} additional inference-time overhead, while requiring only a tiny fraction of trainable parameters (typically $<0.2\%$) and modest training data (a small set of trajectories).
Across diverse benchmarks and backbones, MTEO achieves state-of-the-art performance in the 3--6 NFE regime and significantly narrows the quality gap to distillation-based methods under extreme acceleration (see \cref{sec:exp}).



\section{Background}

We review only the components directly used by MTEO; additional background (\textit{Overview of Diffusion Models, Trajectory Distillation}) is deferred to Appendix.
\label{sec:background}

\subsection{Procedure of Time Information Injected into Diffusion Network}
\label{subsec:procedure}
A wide range of neural network architectures have been adopted for diffusion models. Despite differences in specifics, most models are built on backbones such as U-Net~\cite{Unet} or Transformer~\cite{DiT}, which typically follow a layer-wise design with multi-scale feature hierarchies and skip connections. Regardless of the architectural details, the core computational workflow is similar. At each time step $t$, the noisy input $x_t$ is fed into the network (see~\cref{fig:overview}) to predict the output: the denoised $x_0$ or the noise residual $\epsilon$. Crucially, the network is conditioned on the time step through a \textit{time embedding}.

In practice, time information is injected into each layer of the network through the following process: First, the current time step $t$ is encoded using a sinusoidal positional encoding and then passed through an MLP to produce a time embedding vector. As feature information flows through the network layer by layer, the time embedding vector is injected into specific layers, allowing the model to modulate its computation based on the current noise level. This enables each layer to process both the intermediate feature map and the corresponding time embedding, effectively integrating both the feature and temporal information throughout the denoising process. At each individual layer, time information specifically modulates the processing of feature information through Feature-wise Linear Modulation (FiLM).




\subsection{Feature-wise Linear Modulation (FiLM)}
\label{subsec:film}
Feature-wise Linear Modulation (FiLM) is a conditional mechanism that adjusts intermediate features by applying a learnable, feature-wise affine transformation (i.e., scaling and shifting) as a function of the conditioning signal. Concretely, let $s_l \in \mathbb{R}^{c \times w \times h}$ denote the feature tensor at layer $l$, where $c$ is the number of channels and each channel corresponds to a $w \times h$ feature map. FiLM modulates $s_l$ as
\begin{equation}
    s_{\text{modulated}}=\alpha_l \odot s_l + \beta_l ,
\end{equation}
where $\alpha_l,\beta_l \in \mathbb{R}^{c}$ are channel-wise scaling and bias vectors (broadcast over spatial dimensions), and $\odot$ denotes element-wise multiplication. This operation can be viewed as applying a per-channel linear transformation to the feature representation at each layer.

\begin{figure*}[htbp]
    \centering
    \includegraphics[width=0.95\linewidth]{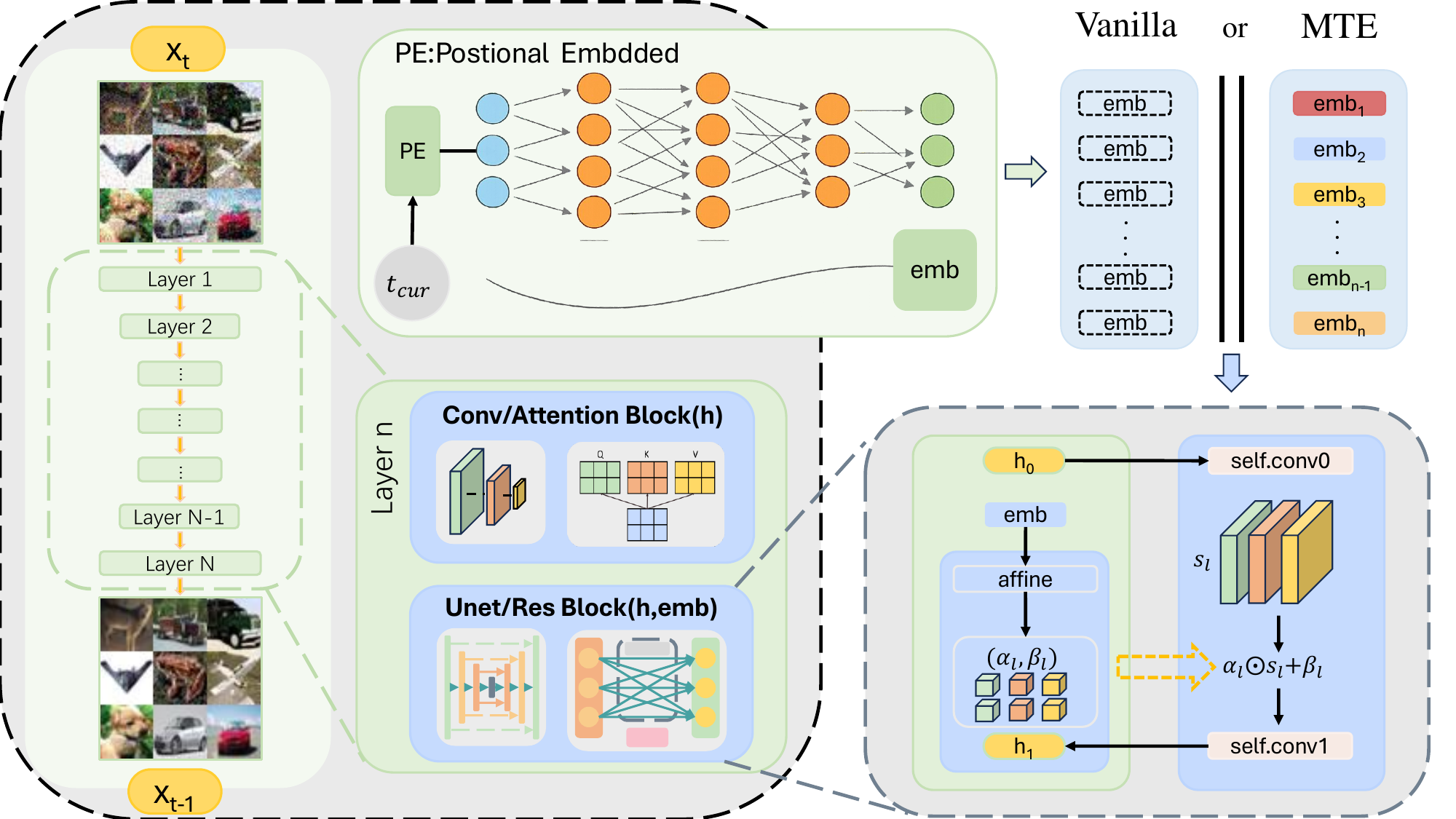}
    \caption{Overview. (left) The visualization of single sampling step. (mid top) Procedure from $t_{cur}$ to time embedding. (mid bottom) Two type of diffusion inner blocks. (right top) The visualization of multi-layer and vanilla embedding. Different colors represent independence across the layers, dashed box denote the embedding are produced by broadcasting. (right bottom) The inner display of FiLM exert on feature maps.}
    \label{fig:overview}
\end{figure*}

In the U-Net framework, the time embedding vector injected into layer $l$ is first passed through a layer-specific \texttt{affine} projection to produce the corresponding FiLM parameters $(\alpha_l,\beta_l)$, which are then used to modulate $s_l$ as above.
In the DiT framework, the overall procedure is conceptually the same: the time embedding is provided to each Transformer block and mapped by the block's \texttt{adaLN\_modulation} module (which plays a role analogous to the affine projection in U-Net) to generate $(\alpha_l,\beta_l)$ for feature modulation.~\cref{fig:overview} (right,bottom) provides a visual illustration of this process.




\section{Method}



\subsection{Suboptimality of Conventional Single Time Variable Design}
\label{subsec:inefficiency}
We begin by examining why the conventional practice of conditioning the denoiser on a \emph{single global time variable} can be suboptimal in the few-step regime.
Most diffusion backbones encode the scalar timestep using a sinusoidal embedding (inspired by the positional encoding in Transformers~\cite{transformer}) and use it to condition the network on the current noise level.

The issue becomes pronounced when the sampling budget is extremely small.
With few steps, each solver update spans a large time interval from $t_{\text{cur}}$ to $t_{\text{next}}$, yet the standard sampler queries the network using the endpoint time $t_{\text{cur}}$.
As a result, a single network evaluation must serve as a coarse proxy for the model behavior over the entire interval, which can introduce systematic discretization errors.

\begin{figure*}[tb]

    \centering
    
    \begin{subfigure}[b]{0.32\textwidth}
        \centering
        \includegraphics[width=\textwidth]{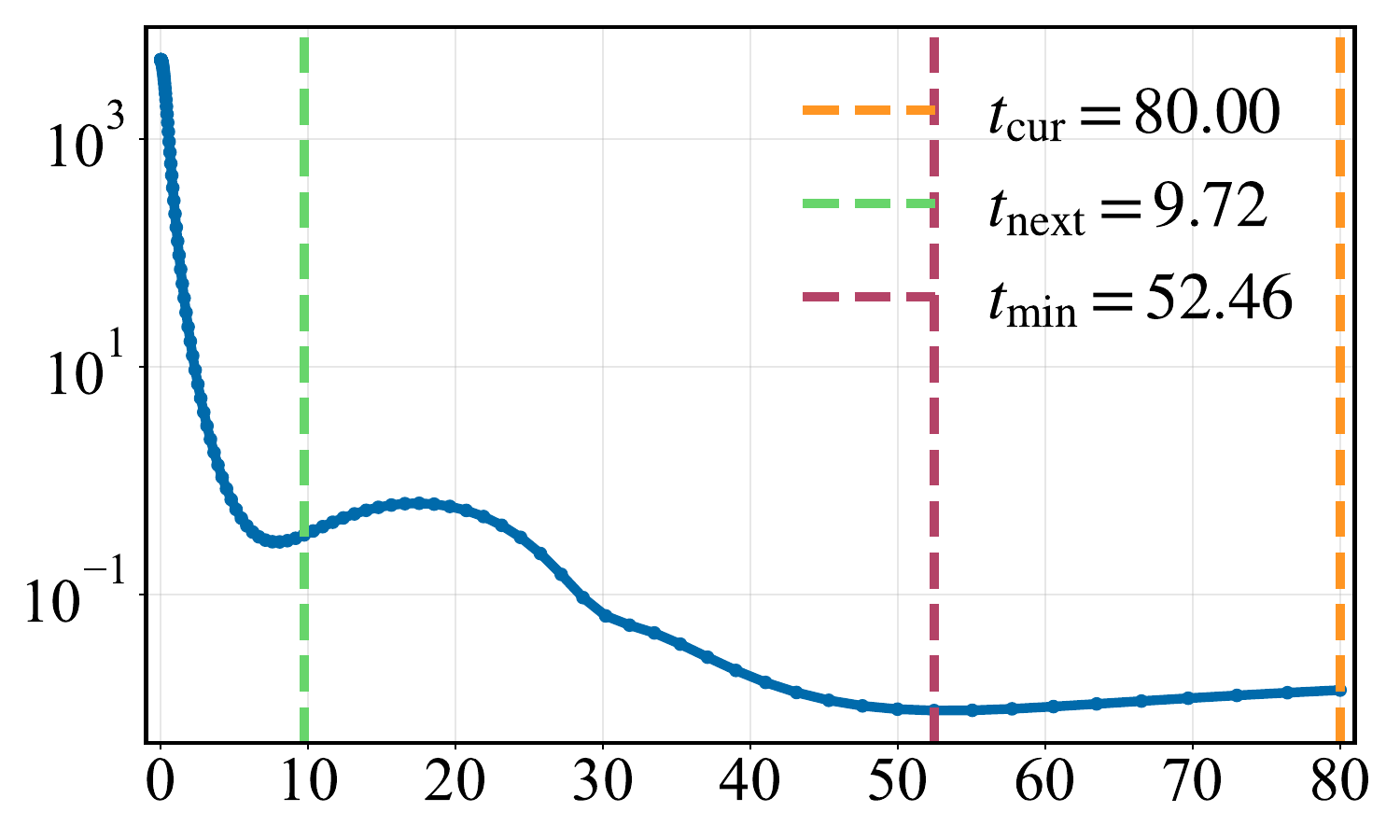}
        \caption{4 steps}
    \end{subfigure}
    \hfill
    \begin{subfigure}[b]{0.32\textwidth}
        \centering
        \includegraphics[width=\textwidth]{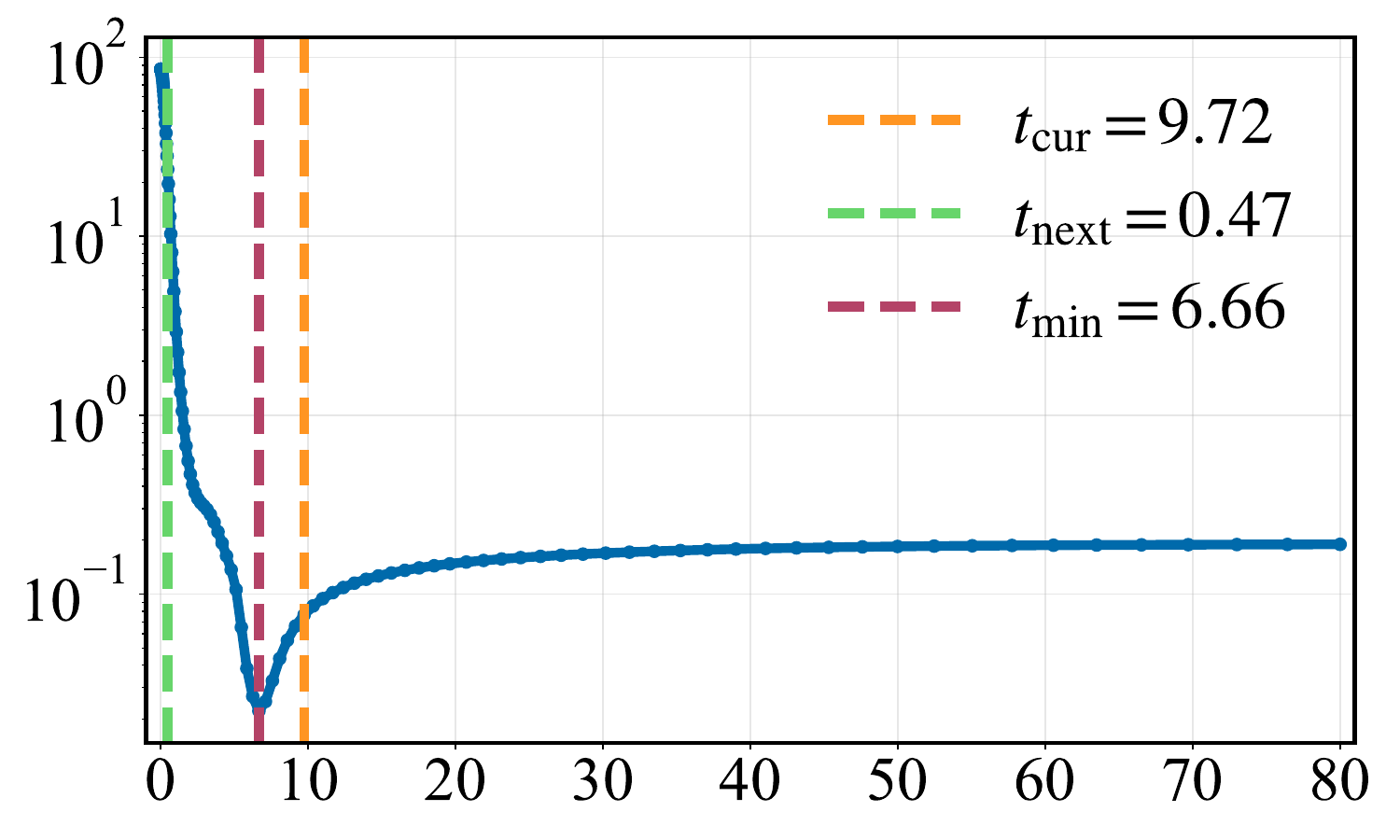}
        \caption{4 steps}
    \end{subfigure}
    \hfill
    \begin{subfigure}[b]{0.32\textwidth}
        \centering
        \includegraphics[width=\textwidth]{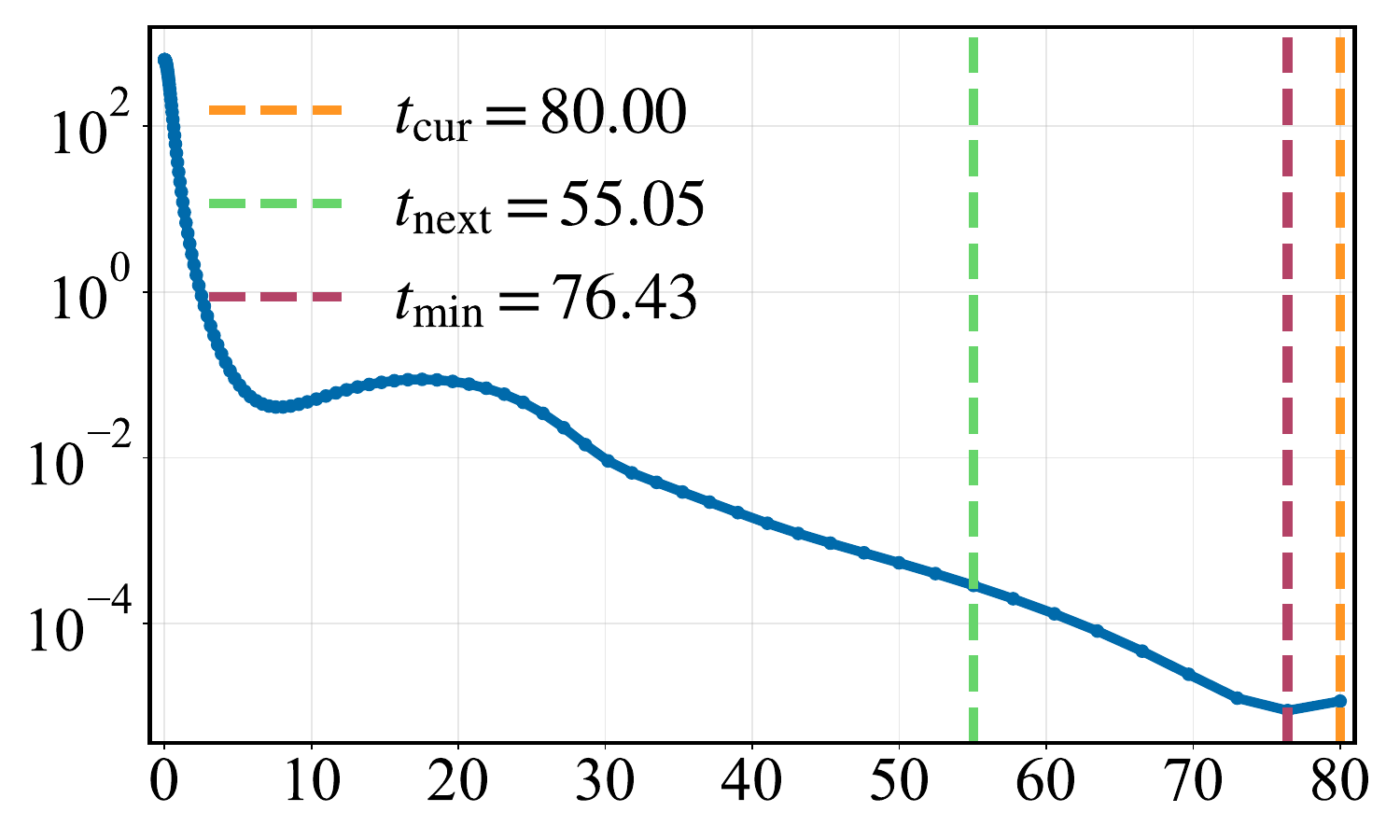}
        \caption{16 steps}
    \end{subfigure}
    
    \vskip\baselineskip 
    \begin{subfigure}[b]{0.32\textwidth}
        \centering
        \includegraphics[width=\textwidth]{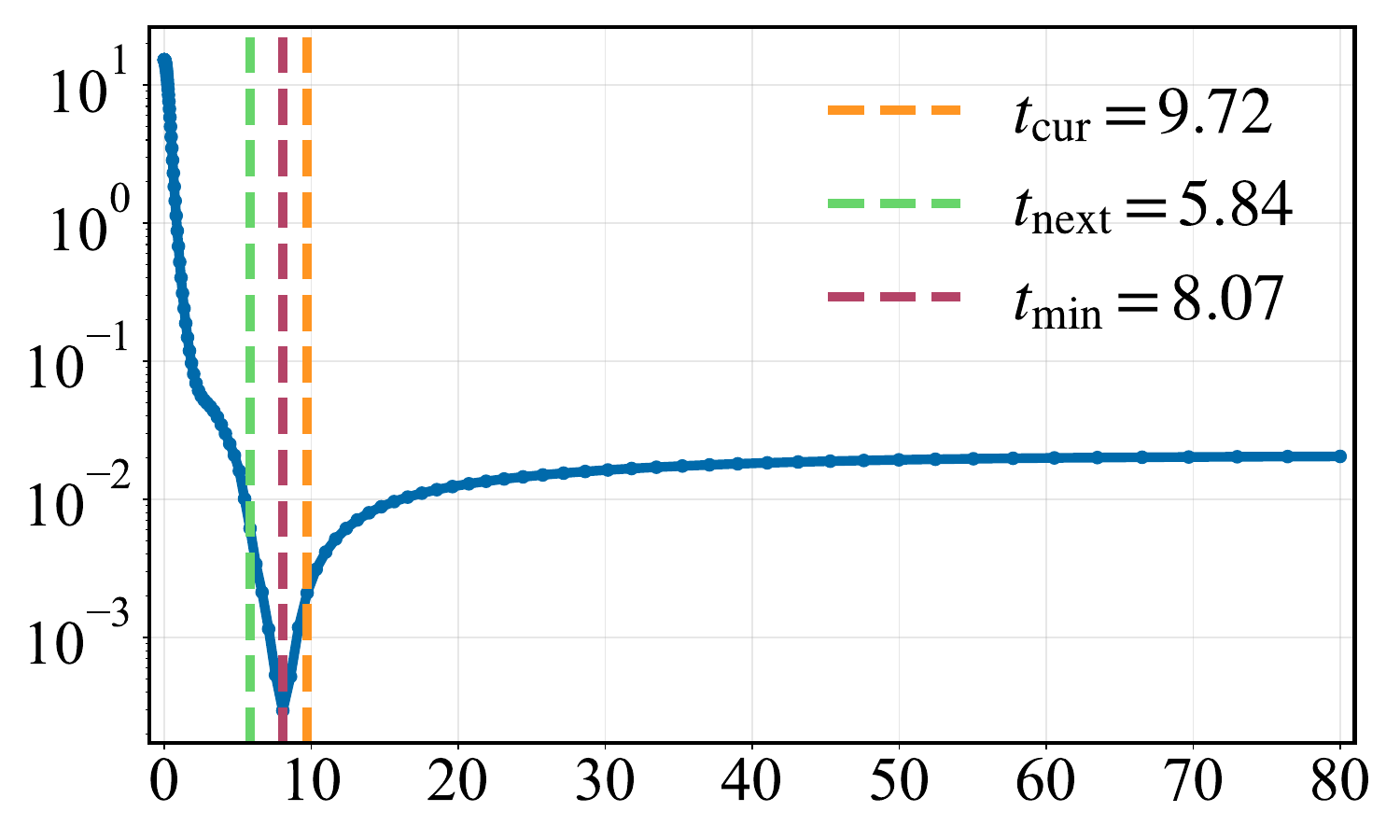}
        \caption{16 steps}
    \end{subfigure}
    \hfill
    \begin{subfigure}[b]{0.32\textwidth}
        \centering
        \includegraphics[width=\textwidth]{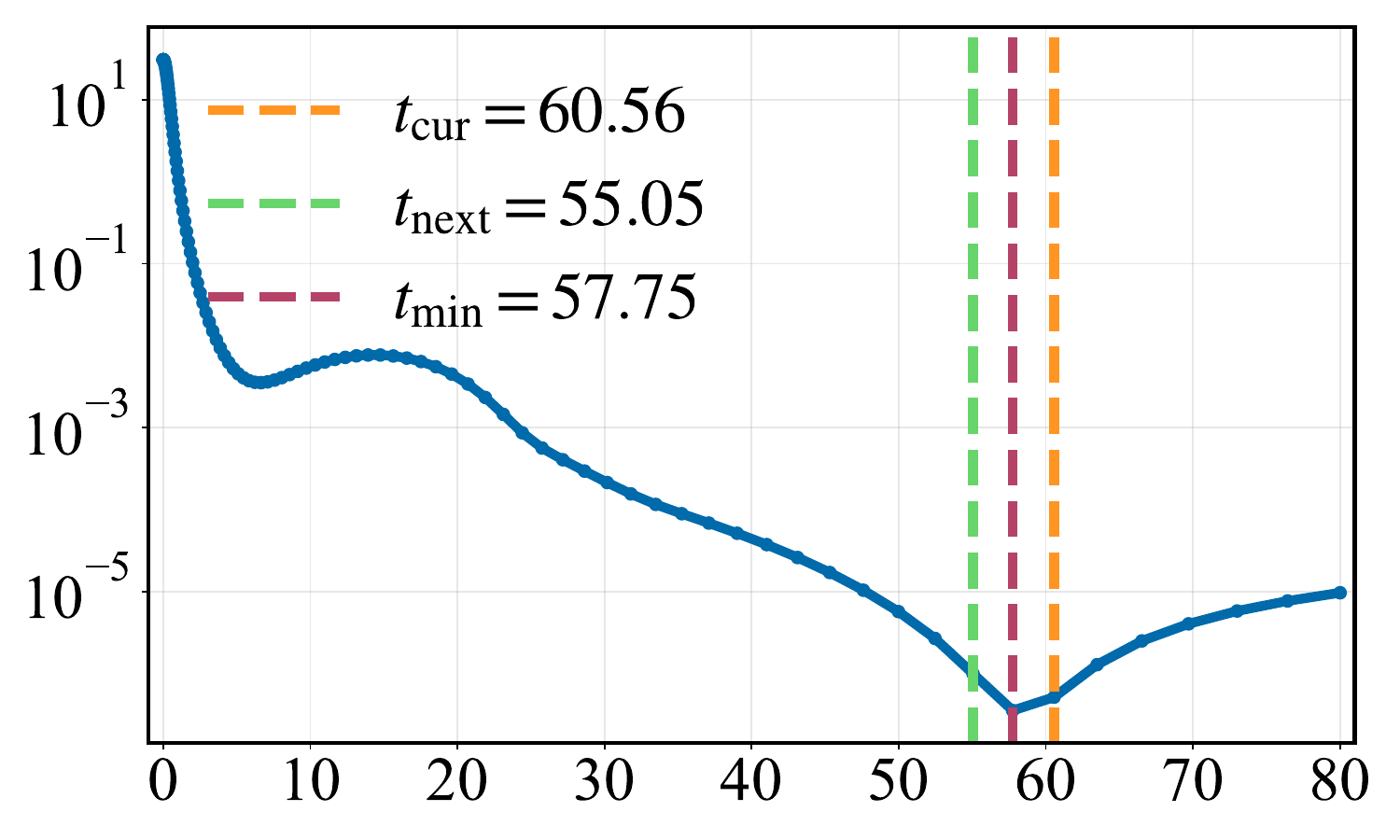}
        \caption{61 steps}
    \end{subfigure}
    \hfill
    \begin{subfigure}[b]{0.32\textwidth}
        \centering
        \includegraphics[width=\textwidth]{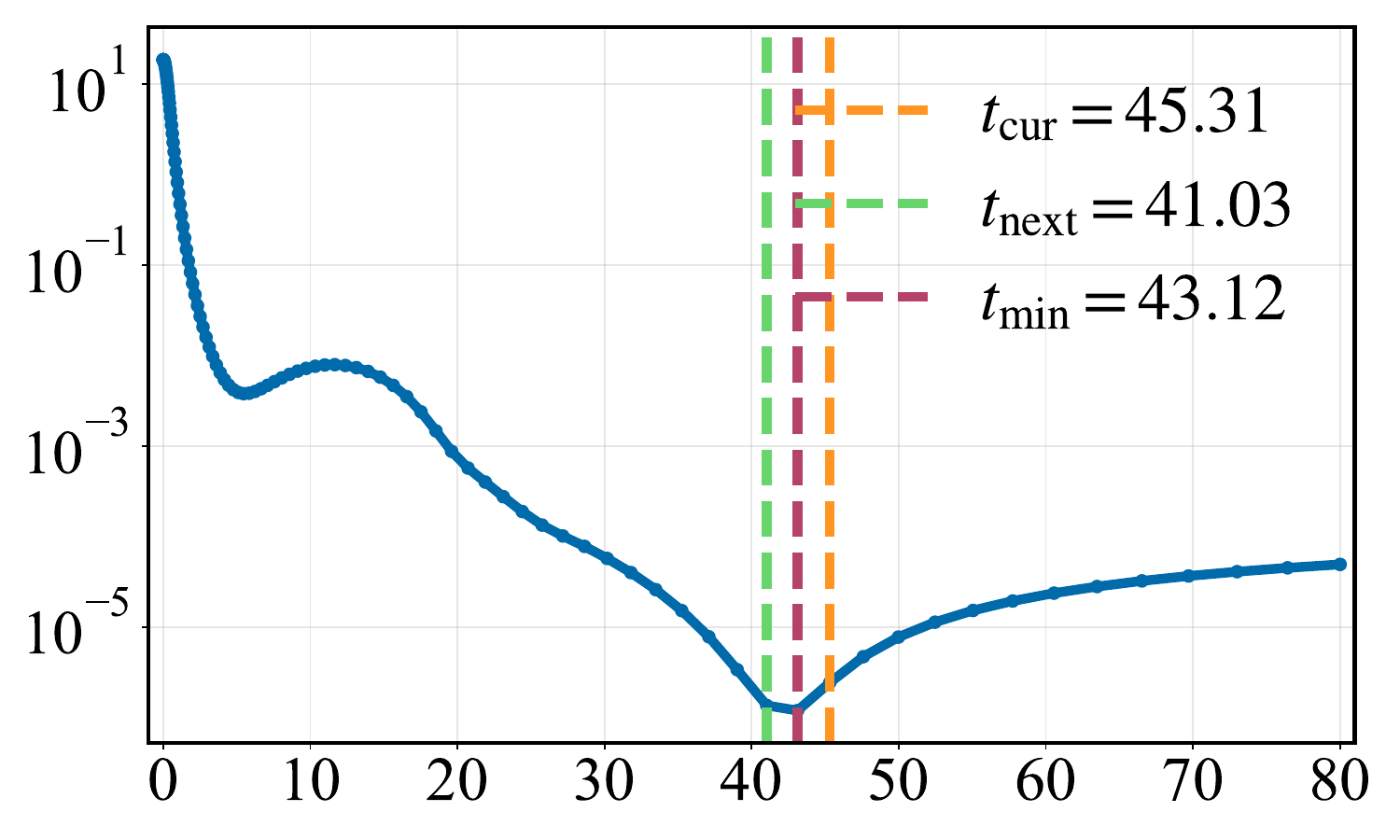}
        \caption{61 steps}
    \end{subfigure}

    \caption{L2 distance between different sampling step of DDIM and ground-truth trajectories on CIFAR-10. We generate the ground-truth trajectory with iPNDM, 61 steps, schedule\_type=\textit{polynomial}, schedule\_rho=\textit{7}. The $\tau$ is generated by 121 steps.}
    \label{fig:motivation-1}
\end{figure*}

To quantify this mismatch, we generate a high-fidelity reference trajectory using a large number of sampling steps, and construct additional trajectories with fewer steps (``medium-step'' and ``low-step'').
For each solver step from $t_{\text{cur}}$ to $t_{\text{next}}$ along a low-step trajectory, we treat the conditioning time as a free variable: we sweep $\tau$ from 80.00 to 0.002 while keeping the solver interval fixed, and compare the resulting output against the reference.
 As~\cref{fig:motivation-1}, we find that the time value yielding the best match (denoted $t_{\min}$) consistently lies \emph{inside} the interval $(t_{\text{next}},\, t_{\text{cur}})$, rather than exactly at $t_{\text{cur}}$ as assumed by the standard sampler.
Moreover, $t_{\min}$ deviates further from $t_{\text{cur}}$ as the step size increases, while it naturally approaches $t_{\text{cur}}$ when the step size becomes small.

\begin{figure*}[tbp]

\begin{minipage}[t]{0.49\textwidth}
    \centering
    \includegraphics[width=1.\linewidth]{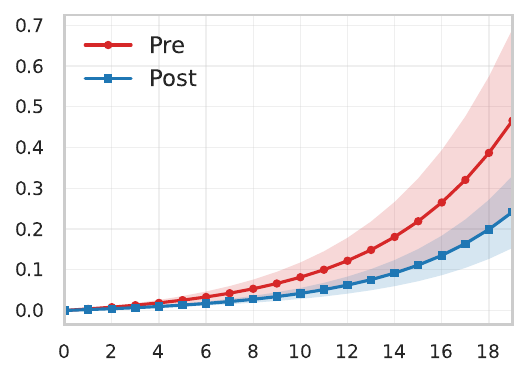}
    \captionof{figure}{The L1 distance of FiLM. We sample 64 ground-truth trajectories and extract the feature maps inside the first block. Then we simulate the first evaluation of a 4-step sample from $t=80.00$ (idx=0) to $t=9.72$ (idx=19). We calculate the L1 distance between $t=80.00$ and $t=9.72$ before (Pre) and after (Post) adjusting the FiLM and present the mean and variance. See more in the appendix.}
    \label{fig:film}
\end{minipage}
\hfill
\begin{minipage}[t]{0.49\textwidth}
    \centering
    \raisebox{30pt}{
    \begin{tabular}{lS[table-format=2.2]S[table-format=2.2]S[table-format=2.2]S[table-format=2.2]S[table-format=2.2]}
        \toprule
        \textbf{Index} & {PC$_1$} & {PC$_{1:2}$} & {PC$_{1:3}$} & {PC$_{1:4}$} & {PC$_{1:5}$} \\
        \midrule
        \multicolumn{6}{l}{\textbf{Shallow}} \\  \midrule
        \addlinespace[2pt]
        0  & \textbf{90.13} & 99.33 & 99.82 & 99.94 & 99.98 \\
        1  & 73.99 & 88.59 & \textbf{94.94} & 98.36 & 99.29 \\
        2  & 67.04 & 82.43 & \textbf{90.70} & 96.40 & 98.30 \\
        32 & 53.16 & 75.76 & 85.36 & \textbf{91.48} & 94.68 \\ 
        33 & 74.25 & 89.34 & \textbf{93.85} & 96.68 & 97.95 \\ \midrule
        \addlinespace[4pt]
        \multicolumn{6}{l}{\textbf{Deep}} \\ \midrule
        \addlinespace[2pt]
        12 & 45.63 & 70.34 & 82.45 & \textbf{90.45} & 94.41 \\
        13 & 44.99 & 69.76 & 81.85 & \textbf{90.07} & 94.26 \\
        14 & 42.10 & 68.67 & 83.16 & \textbf{91.31} & 94.92 \\ 
        16 & 44.46 & 69.35 & 81.80 & 89.77 & \textbf{93.77} \\
        20 & 47.87 & 70.88 & 81.75 & 89.79 & \textbf{93.39} \\
        \bottomrule
    \end{tabular}
    }
    \captionof{table}{Cumulative explained variance (\%) of the first five principal components for trajectory 0. $PC_{1:n}$ denote first n principal components. Complete results are list in the appendix.}
    \label{tab:featuretraj}
\end{minipage}

\end{figure*}
These observations indicate that, in the few-step setting, using a single fixed endpoint time to condition the entire network is generally misaligned with the effective time that best matches the reference trajectory, and is therefore suboptimal.

\begin{figure*}[tb]

    \centering
    
    \begin{subfigure}[b]{0.32\textwidth}
        \centering
        \includegraphics[width=\textwidth]{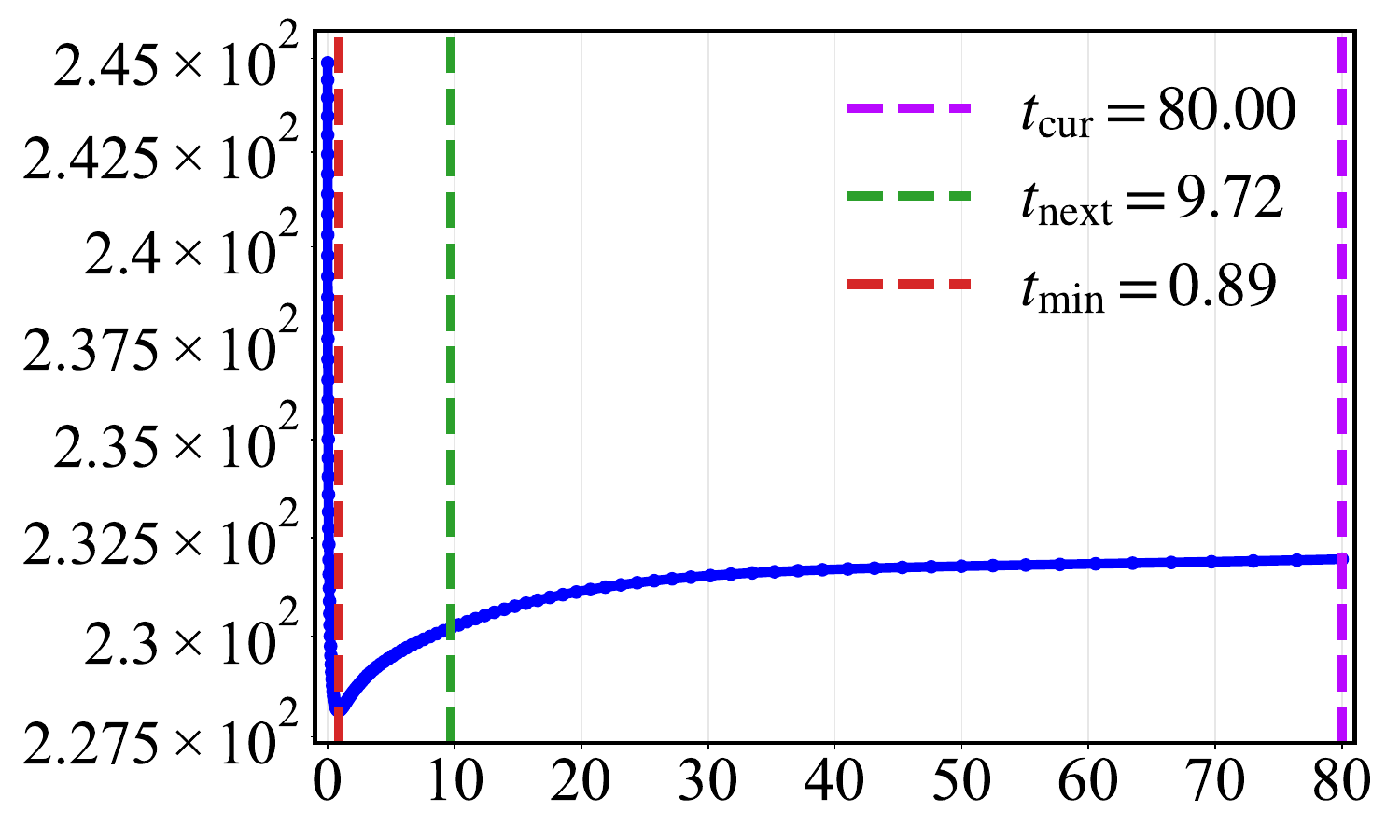}
        \caption{4 steps}
    \end{subfigure}
    \hfill
    \begin{subfigure}[b]{0.32\textwidth}
        \centering
        \includegraphics[width=\textwidth]{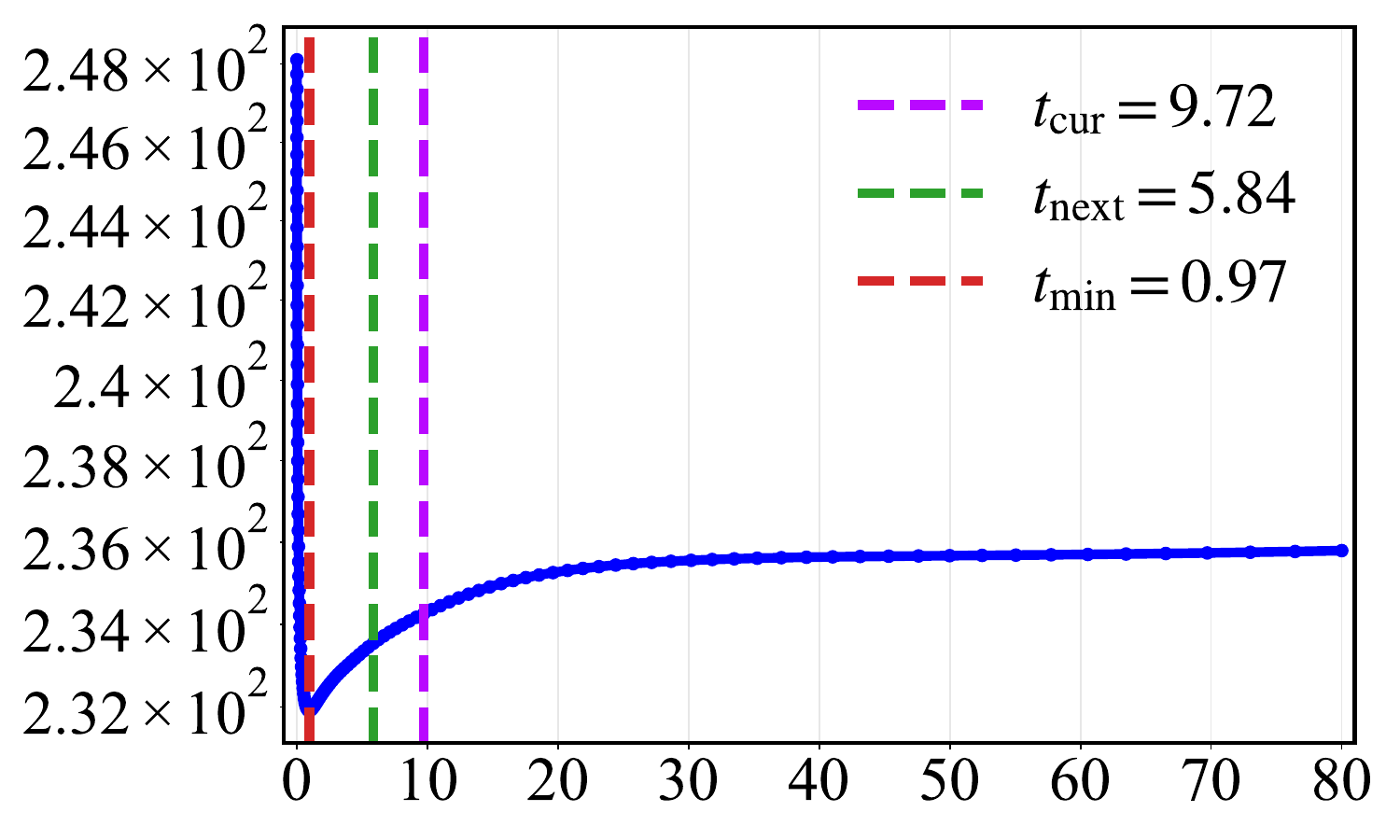}
        \caption{16 steps}
    \end{subfigure}
    \hfill
    \begin{subfigure}[b]{0.32\textwidth}
        \centering
        \includegraphics[width=\textwidth]{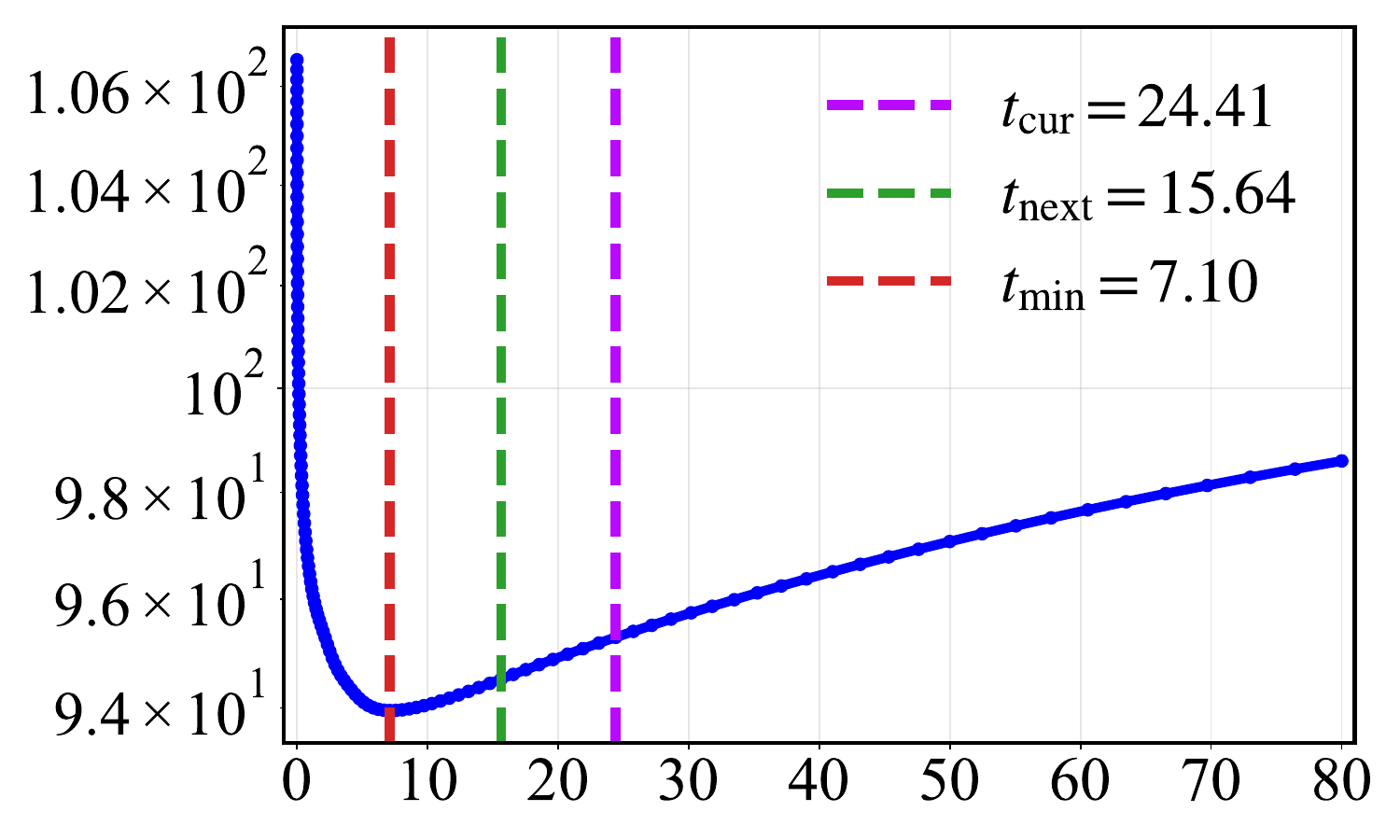}
        \caption{16 steps}
    \end{subfigure}
    
    \vskip\baselineskip 
    \begin{subfigure}[b]{0.32\textwidth}
        \centering
        \includegraphics[width=\textwidth]{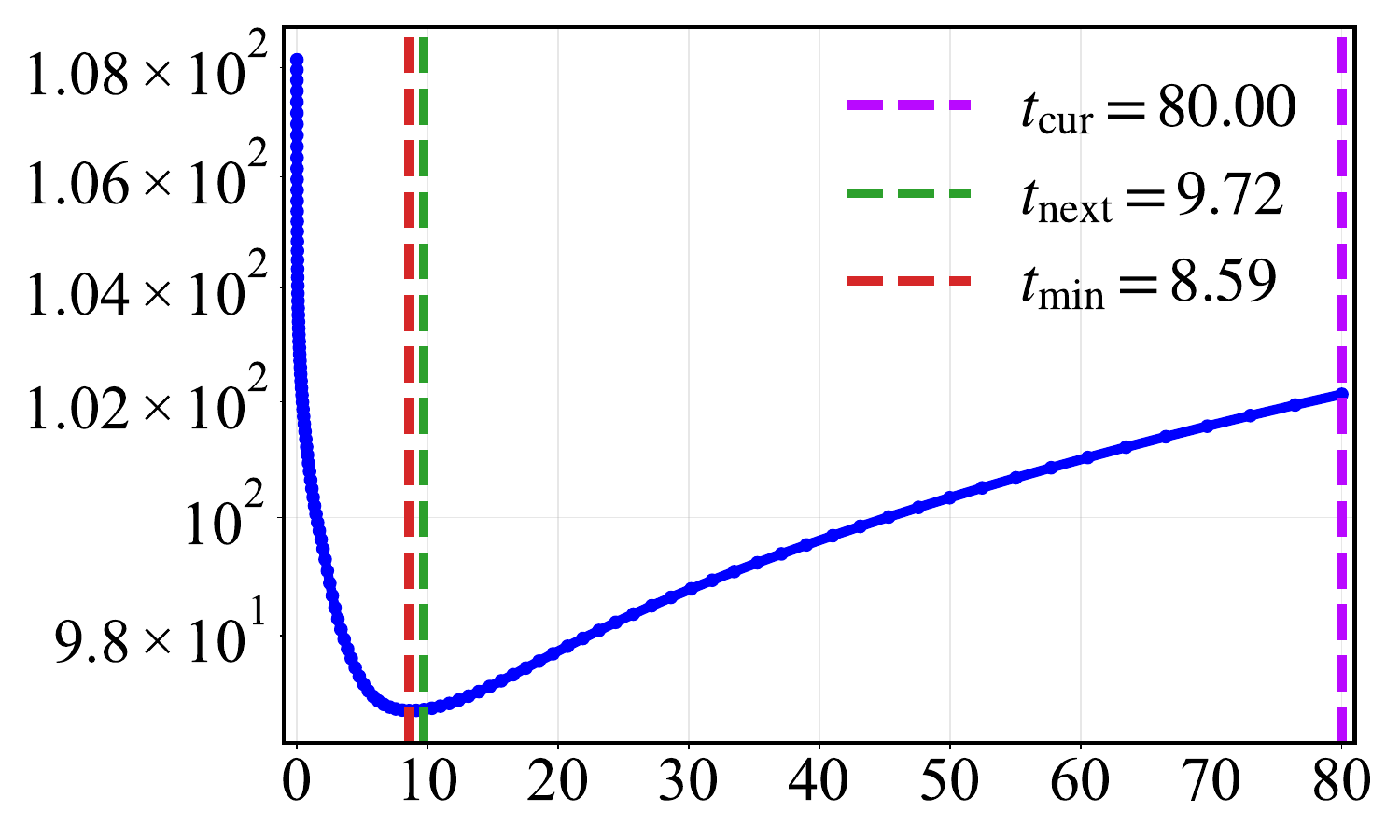}
        \caption{4 steps}
    \end{subfigure}
    \hfill
    \begin{subfigure}[b]{0.32\textwidth}
        \centering
        \includegraphics[width=\textwidth]{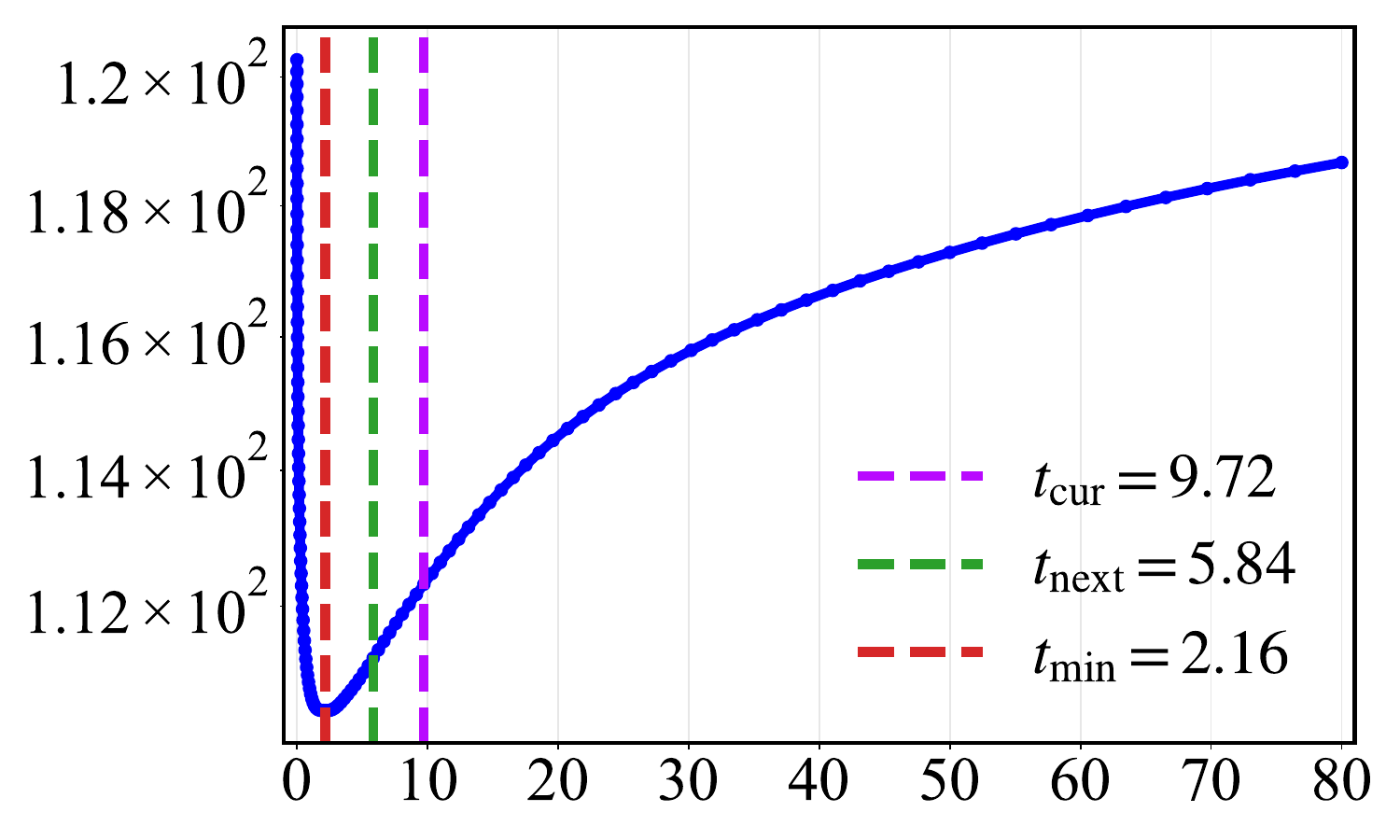}
        \caption{16 steps}
    \end{subfigure}
    \hfill
    \begin{subfigure}[b]{0.32\textwidth}
        \centering
        \includegraphics[width=\textwidth]{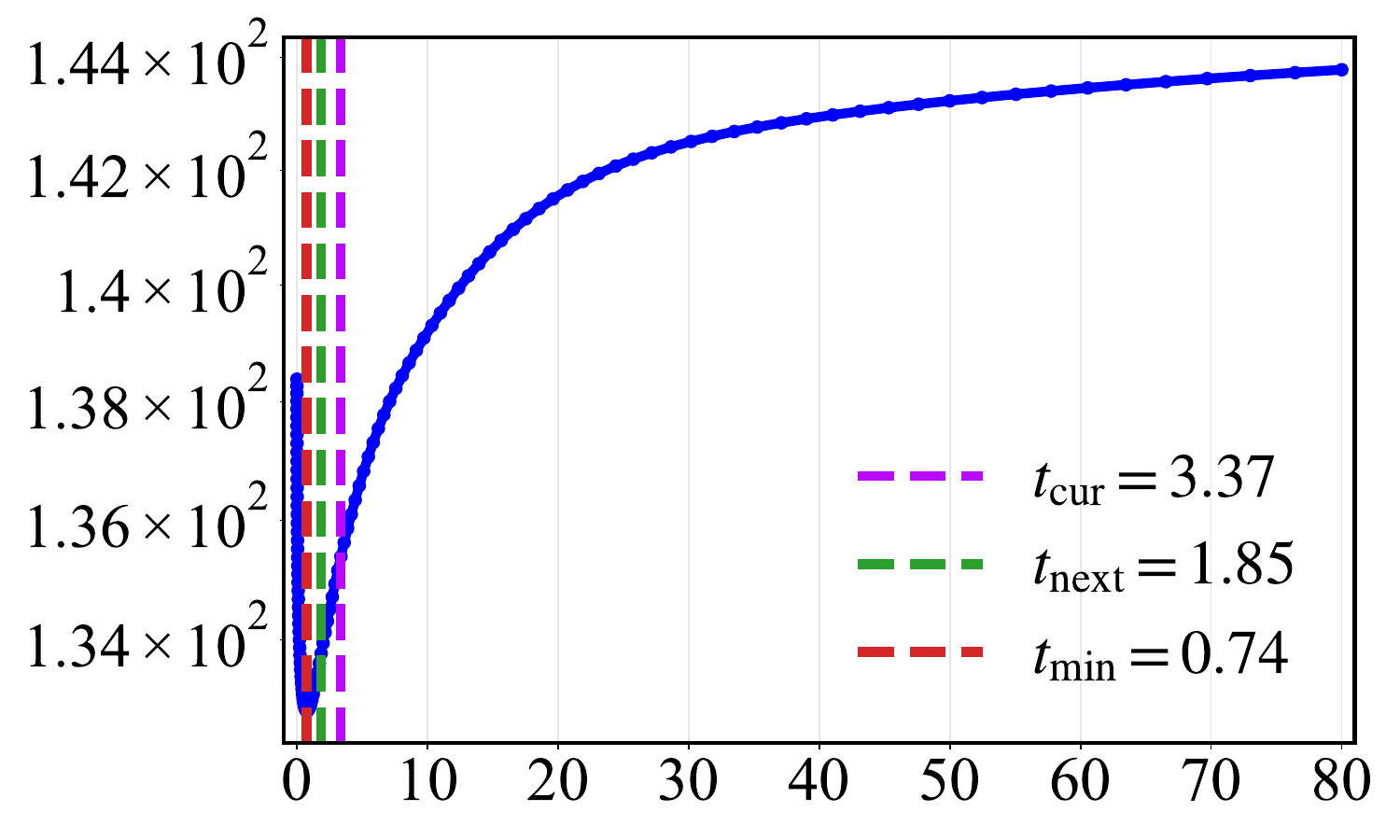}
        \caption{16 steps}
    \end{subfigure}

    \caption{L2 distance between different feature trajectories. The setting is same as~\cref{fig:motivation-1}.}
    \label{fig:moti-layer}
\end{figure*}

\subsection{Distinct Feature Trajectories Across Layers}
\label{subsec:feature_traj}
It has been observed in prior work~\cite{AMED} that the sampling trajectory of a diffusion model often lies nearly in a two-dimensional subspace of the high-dimensional state space.
Motivated by this observation, we hypothesize that, analogous to the model’s overall sampling trajectory, each layer of the network traces its own \emph{feature trajectory} over the course of denoising. To analyze these differences, we captured a reference denoising trajectory and recorded the feature map at every layer along this trajectory. Following the approach of previous work~\cite{AMED}, we applied principal component analysis (PCA) to the sequence of feature maps from each layer and visualized each layer’s trajectory in its principal component subspace. This figure (Appendix) shows examples of these trajectories.

As shown in figure (Appendix), these layer-specific trajectories differ from the global one, and from each other, in two major ways: (1) the dimensionality of the low-dimensional subspace that a given layer’s trajectory occupies varies across layers; and (2) although all layers are driven by the same time variable $t$, the curvature (i.e., the degree of turning) of feature trajectories differs substantially across layers.


\cref{tab:featuretraj} quantifies the dimensionality of these trajectories: in early layers, the first 2 or 3 principal components explain over 90\% of the variance, indicating that the feature trajectory lies essentially in a planar (2D or 3D) subspace. In deeper layers, the variance is distributed across more dimensions: 4 or 5 components are required to reach 90\% explained variance.

Taken the perspective on the global sampling path, these inter-layer discrepancies suggest that the \emph{effective} time that best aligns with each layer’s feature trajectory may differ across layers. Similar to the experiment in \cref{fig:motivation-1}, we apply the same time sweeping procedure to layer-wise feature trajectories. As shown in~\cref{fig:moti-layer}, the optimal conditioning time $t_{\min}$ for a given layer often deviates from the current point $t_{\text{cur}}$, and moreover varies across layers. This empirically indicates that layer-wise feature dynamics are not strongly coupled to a single shared input time, and motivates assigning each layer its own effective time $t_l$ (equivalently, layer-specific time conditioning) during sampling.

\begin{figure}[tbp]
  \centering
  \begin{subfigure}[b]{0.47\textwidth}
    \includegraphics[width=\textwidth]{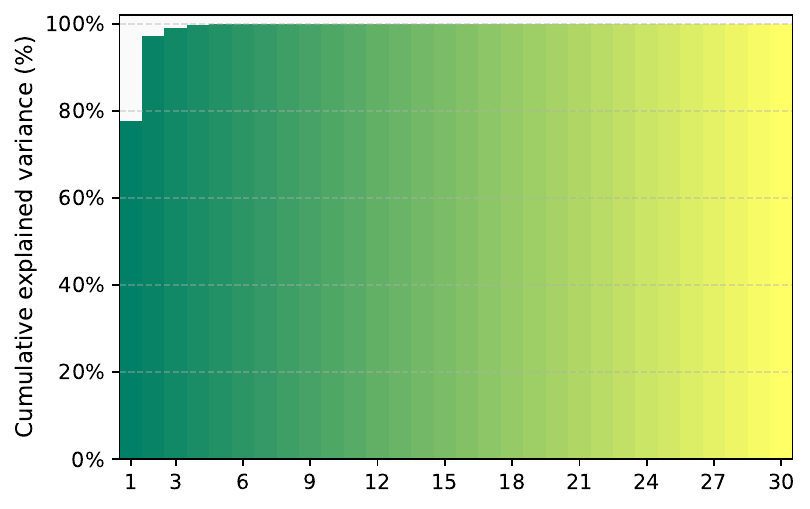}
    \caption{Vanilla time embedding}
  \end{subfigure}
  \hfill
  \begin{subfigure}[b]{0.47\textwidth}
    \includegraphics[width=\textwidth]{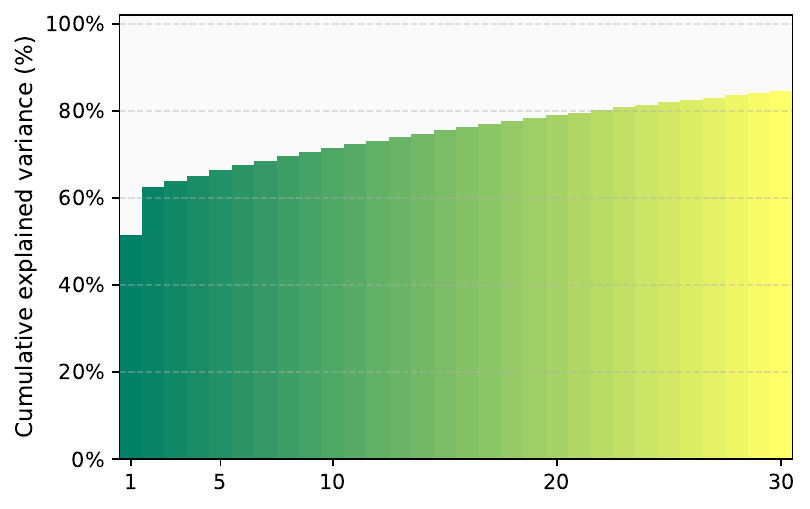}
    \caption{MTE time embedding}
  \end{subfigure}
  
  
  \begin{subfigure}[b]{0.47\textwidth}
    \includegraphics[width=\textwidth]{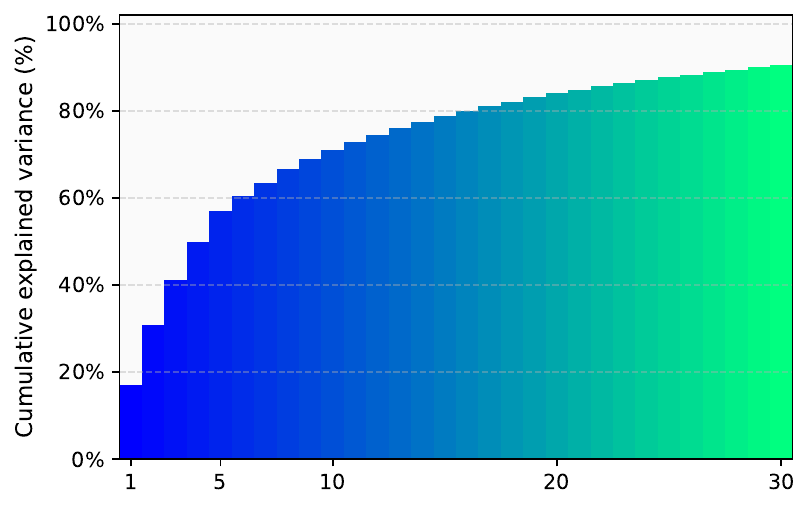}
    \caption{Vanilla FiLM parameters}
  \end{subfigure}
  \hfill
  \begin{subfigure}[b]{0.47\textwidth}
    \includegraphics[width=\textwidth]{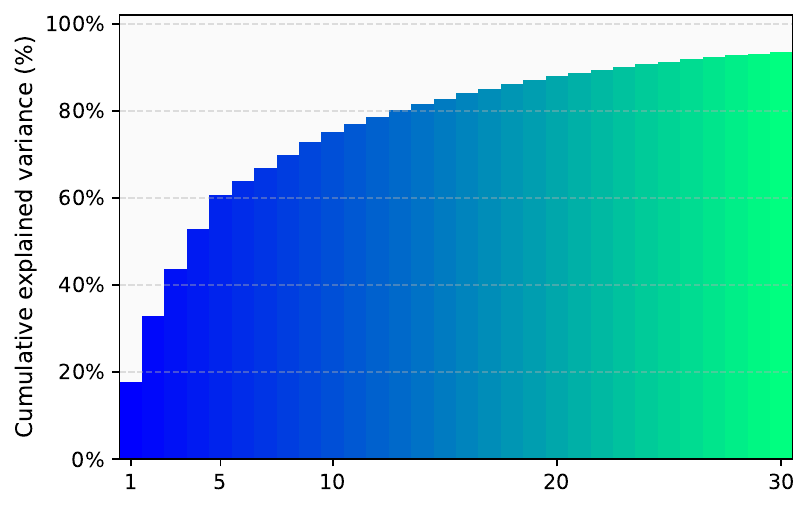}
    \caption{MTE FiLM parameters}
  \end{subfigure}
  \caption{PCA analysis on conventional embedding design and MTE.  The x-axis denote first n principal components. \textcolor{green}{Green}: time embedding. \textcolor{blue}{Blue}: FiLM parameters.}
  \label{fig:pca_emb}
\end{figure}

\subsection{From Time Variable to Time Embedding}
\label{subsec:vartoemb}
Previously, we argued that during sampling, different layers of a diffusion network can exhibit a certain degree of independence with respect to the time variable, i.e., different layers may prefer different effective times. We now further examine the design that maps the scalar time variable to the time embedding injected into the network.

As discussed earlier, the scalar time $t$ is first encoded via a sinusoidal embedding and then passed through an MLP to produce a time embedding vector. As the data flows through the network, this time embedding is injected into each layer and projected by a layer-specific affine transform to generate the FiLM modulation parameters, the channel-wise scaling and shift vectors $(\alpha_l,\beta_l)\in\mathbb{R}^{c}$, which directly modulate intermediate features.

This raises an important question: does the resulting time embedding provide sufficient expressive power to generate appropriate layer-wise $(\alpha_l,\beta_l)$, or does it instead impose a representational bottleneck? To probe this, we collect the time embeddings from a 121-step schedule and perform PCA. As shown in~\cref{fig:pca_emb}(left), the first two principal components explain \textbf{77.72\%} and \textbf{19.41\%} of the variance, respectively, accounting for \textbf{97.14\%} in total. This indicates that the time embeddings across timesteps lie close to a very low-dimensional subspace.

In contrast, when we perform PCA on the FiLM parameters $(\alpha_l,\beta_l)$ across time steps and layers, we find substantially higher intrinsic dimensionality: the first two principal components explain only \textbf{19.41\%} of the variance, and even the first 30 components explain only \textbf{90.49\%}. This mismatch suggests that the conventional time-embedding design can severely restrict the degrees of freedom available to FiLM modulation, thereby limiting its representational capacity. We also conduct the same PCA procedure on MTE~\cref{fig:pca_emb}(right), which shows the explosive degree of freedom, futher strengthen our claim. We continue the discussion in Appendix.

Finally, we empirically verify whether FiLM itself has sufficient modulation capacity to correct layer-wise feature processing when provided with appropriate parameters. We compare a high-step sampling trajectory (iPNDM-61) with an extremely few-step trajectory (DDIM-4). For one layer at time $t$ along the few-step trajectory, we record the feature maps before and after FiLM as $s^t$ and $s^t_{\text{modulated}}$, respectively. For the corresponding teacher trajectory at time $\tau$, we denote the analogous quantities as $\hat{s}^\tau$ and $\hat{s}^\tau_{\text{modulated}}$. We define the $\ell_1$ discrepancy as
\begin{equation}
    \mathcal{L}
    = \left\lVert \hat{s}^\tau_{\text{modulated}} - s^t_{\text{modulated}} \right\rVert_1
    = \left\lVert \hat{s}^\tau_{\text{modulated}} - (\alpha_l \odot s^t + \beta_l) \right\rVert_1 .
\end{equation}
In principle, by choosing suitable scaling and bias vectors $(\alpha_l,\beta_l)$, FiLM can reduce this discrepancy and align the modulated features. Following~\cite{AMED}, for a solver step spanning $(t_{\text{next}}, t_{\text{cur}})$, there exists an intermediate time $t_{\min}\in (t_{\text{next}}, t_{\text{cur}})$ whose direction provides the best update. Therefore, we scan the teacher trajectory within $(t_{\text{next}}, t_{\text{cur}})$ (to approximate $t_{\min}$), compute the original $\ell_1$ loss, and then optimize $(\alpha_l,\beta_l)$ to minimize the loss for each $\tau$, reporting the mean and variance. As shown in~\cref{fig:film}, FiLM modulation can substantially reduce the feature discrepancy, demonstrating its strong corrective capability. This also suggests that such capability is not fully exploited under the conventional time-embedding pipeline, leaving layers insufficiently modulated and leading to pronounced quality degradation in few-step sampling.




\subsection{Multi-layer Time Embedding}
\label{subsec:mte}

Our analyses in \cref{subsec:feature_traj,subsec:inefficiency,subsec:vartoemb} point to a consistent message: \emph{time conditioning in diffusion networks is inherently layer-dependent in the few-step regime}.
Specifically, (i) different layers exhibit distinct feature trajectories with different geometric complexity, suggesting that their dynamics may align best with different effective times; and (ii) the conventional shared timestep embedding provides limited degrees of freedom for generating diverse layer-wise FiLM modulation parameters.
Together, these findings indicate that forcing all layers to share the same time embedding at each solver step can be unnecessarily restrictive, and may under-utilize FiLM's modulation capacity.

Motivated by this, we propose \emph{Multi-layer Time Embedding (MTE)}.
In standard diffusion samplers, a single time embedding computed from the solver time (\eg, $t_{\text{cur}}$) is broadcast to all layers.
By contrast, MTE maintains a \emph{layer-specific} embedding for each sampling step:
for step $i$ (with solver time $t_i$), we use a set of learnable vectors
$\Phi_i=\{\phi_{i,\ell}\}_{\ell=1}^{L}$, where $\phi_{i,\ell}\in\mathbb{R}^{d}$ is used exclusively to condition layer $\ell$.
During sampling, each layer consumes its own $\phi_{i,\ell}$ to produce its FiLM modulation parameters, while the backbone denoiser weights remain unchanged.
This decoupling removes the implicit synchronization of time conditioning across layers, allowing each block to adapt its modulation to its own feature trajectory rather than being constrained by a shared embedding.
\cref{fig:overview} (right top) provides a conceptual illustration of the MTE design. 
It also naturally raises the next question: how can we obtain MTE for a fixed few-step sampling schedule?

\begin{algorithm}[!h]
    \renewcommand{\algorithmicrequire}{\textbf{Input:}}
    \renewcommand{\algorithmicensure}{\textbf{Output:}}
    \caption{Training MTE via Trajectory Distillation}
    \label{alg:train}
    \begin{algorithmic}[1]
        \REQUIRE Teacher trajectory $\{\hat{x}_{t_i}\}_{i=0}^{N-1}$, time schedule $\{t_i\}_{i=0}^{N-1}$, ODE solver $S$,
        frozen denoiser $\epsilon_\theta$, distance metric $\mathcal{D}$,
        layer-wise embeddings $\{\Phi_i\}_{i=0}^{N-1}$ with $\Phi_i=\{\phi_{i,\ell}\}_{\ell=1}^{L}$.
        \ENSURE Optimized embeddings $\{\Phi_i\}_{i=0}^{N-1}$.
        \STATE $x_{t_0} \leftarrow \hat{x}_{t_0}$
        \FOR{$i = 0$ \textbf{to} $N-2$}
            \REPEAT
                \STATE $x_{t_{i+1}} \leftarrow S(x_{t_i},\, \epsilon_\theta,\, t_i,\, t_{i+1},\, \Phi_i)$
                \STATE $\mathcal{L} \leftarrow \mathcal{D}\!\big(x_{t_{i+1}},\, \hat{x}_{t_{i+1}}\big)$
                \STATE Update $\Phi_i$ using $\nabla_{\Phi_i}\mathcal{L}$
            \UNTIL{converged}
            \STATE $x_{t_{i+1}} \leftarrow S(x_{t_i},\, \epsilon_\theta,\, t_i,\, t_{i+1},\, \Phi_i)$
        \ENDFOR
        \RETURN $\{\Phi_i\}_{i=0}^{N-1}$
    \end{algorithmic}
\end{algorithm}

\subsection{Learning MTE via Trajectory Distillation}
\label{subsec:train}

The analysis in \cref{subsec:vartoemb} suggests that FiLM has sufficient capacity to correct intermediate representations when provided with appropriate modulation parameters.
We therefore learn MTE by a trajectory-matching objective: we freeze the pretrained diffusion model and optimize only the layer-wise time embeddings so that a few-step sampler can track a high-fidelity teacher trajectory.

Concretely, we first generate a teacher trajectory $\{\hat{x}_{t_i}\}_{i=0}^{N-1}$ using a high-quality sampler with a large number of function evaluations.
We then run a student trajectory $\{x_{t_i}\}_{i=0}^{N-1}$ using a fixed $N$-step schedule $\{t_i\}_{i=0}^{N-1}$, but with MTE enabled.
Let $\Phi_i=\{\phi_{i,\ell}\}_{\ell=1}^{L}$ denote the collection of learnable embedding vectors at step $i$ across all layers.
We initialize $x_{t_0}=\hat{x}_{t_0}$ and, for each step $i$, perform one solver update
\begin{equation}
    x_{t_{i+1}} = S\!\left(x_{t_i},\, \epsilon_\theta,\, t_i,\, t_{i+1},\, \Phi_i\right),
\end{equation}
where $\epsilon_\theta$ is the frozen denoiser and $S(\cdot)$ is the chosen ODE solver (sampler step).
We optimize $\Phi_i$ to match the teacher state at the same time using an $\ell_2$ objective:
\begin{equation}
    \mathcal{L}_i = \left\lVert x_{t_{i+1}} - \hat{x}_{t_{i+1}} \right\rVert_2^2 .
\end{equation}
In practice, we adopt a stage-wise training scheme and optimize $\Phi_i$ step by step, which avoids excessive overhead and stabilizes optimization.

To efficiently train all embeddings, we use an early-stopping criterion based on the \emph{relative} improvement in loss,
\begin{equation}
    \mathcal{L}_{\mathrm{rel}} = \frac{\mathcal{L}_{\mathrm{prev}} - \mathcal{L}_{\mathrm{cur}}}{\mathcal{L}_{\mathrm{prev}}} .
\end{equation}
Given a threshold $\varepsilon$, patience $p$, and a maximum epoch budget $E_{\max}$, we stop updating $\Phi_i$ if
$\mathcal{L}_{\mathrm{rel}} < \varepsilon$ for $p$ consecutive checks, or when $E_{\max}$ epochs are reached.
For later denoising steps (smaller $t$), we progressively tighten the threshold $\varepsilon$.
For the last step, we disable early stopping and always train up to $E_{\max}$ epochs.
The full algorithm, including the optimization schedule, is provided in the Appendix.

\section{Experiments}
\label{sec:exp}
\subsection{Experimental Settings}

\subsubsection{Datasets and Checkpoints.}
We evaluate MTEO under two representative diffusion backbones: (i) EDM-style U-Net models and (ii) Diffusion Transformers (DiT).
Under the EDM setting, we consider CIFAR-10 (32$\times$32)~\cite{cifar10}, FFHQ (64$\times$64)~\cite{ffhq}, ImageNet-64 (64$\times$64)~\cite{imagenet}, and LSUN Bedroom (256$\times$256)~\cite{lsun}.
We further evaluate text-to-image generation on MS-COCO (512$\times$512)~\cite{coco} using Stable Diffusion v1.5~\cite{SD}.
Under the DiT setting, we report results on ImageNet-256 (256$\times$256)~\cite{imagenet}.
Regarding conditioning, CIFAR-10, FFHQ, and LSUN-Bedroom use unconditional generation checkpoints, while ImageNet-64 and ImageNet-256 are class-conditional.
For MS-COCO, we use text prompts and the corresponding text encoder in Stable Diffusion v1.5.
All pretrained diffusion checkpoints (and any associated configuration details) are summarized in the Appendix.
\subsubsection{Training Configuration.}
We train MTEO following the trajectory-distillation procedure described in \cref{subsec:train} (see also \cref{alg:train}). As default, We use 256 teacher trajectories(seeds 50000-50255), batch size=64, $\epsilon=0.01$, $E_{max}=300$, $p=10$, and teacher's NFE=21,20,20,24 corresponding to student's NFE=3,4,5,6.
For all EDM experiments (across datasets and solvers), we enable the early-stopping and convergence-control strategy in \cref{subsec:train} to reduce training overhead while maintaining stable optimization.
For DiT, we disable early stopping and instead train for a fixed number of epochs for simplicity and stability.
The full set of training hyperparameters for every experiment, and additional training diagnostics (\eg., loss curves), intermediate visualizations are provided in Appendix. The training procedure is highly reproducible.
\subsubsection{Evaluation Protocol.}
We adopt the Fr\'echet Inception Distance (FID)~\cite{fid} as the primary metric.
Unless otherwise stated, we generate 50{,}000 samples to compute FID, using a unique random seed per image (seeds 0--49{,}999).
For transparency and reproducibility, the reference statistic files used for FID computation on each dataset are listed in the Appendix. 
We additionally report Inception Score (IS)~\cite{IS}, sFID~\cite{sfid}, Precision/Recall~\cite{precisionRecall}, and CLIP Score~\cite{clip,clipscore}. All results are highly reproducible.

\subsection{Main Results}
\subsubsection{Comparison with Lightweight Samplers.}
We compare MTEO against a broad range of lightweight fast methods, including DDIM~\cite{DDIM}, UniPC~\cite{unipc}, DEIS~\cite{SDE}, iPNDM~\cite{deis&ipndm}, DPM-Solver variants~\cite{DPM,DPM++,DPM-v3}, AMED~\cite{AMED}, and the recent EPD~\cite{epd}. The 3,5 NFE results of AMED and EPD are directly borrow from original papers while others runs on our machine, following the same configuration recommended by authors. Across 3--6 NFE, MTEO achieves sota performance, with pronounced gains at the lowest budget (3 NFE).
For instance, at 3 NFE, the previous sota EPD reports FID of 10.40, 19.02, and 18.28 on CIFAR-10, FFHQ, and ImageNet-64, respectively, whereas MTEO achieves 3.85, 5.46, and 7.42, indicating substantially improved few-step sampling quality.

\begin{table*}[!t]
    \caption{FID evaluation on CIFAR-10 and FFHQ Datasets. We compare MTEO against lightweight and training-free methods. Details are presented in the appendix.}
    \label{tab:cf-lw}
    \centering
    \renewcommand{\arraystretch}{0.8}

    
    \begin{tabularx}{\textwidth}{l|*{4}{>{\centering\arraybackslash}X}|*{4}{>{\centering\arraybackslash}X}}

    \toprule
    Method & \multicolumn{4}{c|}{CIFAR-10} & \multicolumn{4}{c}{FFHQ} \\
    \cmidrule{2-9}
           & \multicolumn{8}{c}{NFE} \\
    \cmidrule{2-9}
           & 3 & 4 & 5 & 6 & 3 & 4 & 5 & 6 \\
    \midrule
    DDIM & 93.36 & 67.40 & 49.66 & 36.08 & 78.16 & 57.37 & 43.85 & 35.15 \\
    iPNDM & 47.98 & 24.81 & 13.59 & 7.05 & 45.90 & 28.21 & 17.14 & 10.00 \\
    UniPC & 109.60 & 45.54 & 23.98 & 11.51 & 86.34 & 44.73 & 21.36 & 12.82 \\
    DEIS & 56.00 & 25.64 & 14.37 & 9.39 & 54.45 & 28.30 & 17.39 & 12.29 \\ \midrule
    \multicolumn{9}{l}{\textbf{DPM}} \\
    \quad +Solver-2 & 155.70 & 145.91 & 57.30 & 59.97 & 266.00 & 238.61 & 87.10 & 83.15 \\
    \quad +Solver++(3M) & 110.00 & 46.52 & 24.97 & 11.99 & 86.45 & 45.94 & 22.51 & 13.74 \\

    \multicolumn{9}{l}{\textbf{AMED}} \\
    \quad +Solver & 18.49 & 17.18 & 7.59 & 7.04 & 47.33 & 31.19 & 14.76 & 11.43 \\
    \quad +Plugin & 10.81 & 10.43 & 6.61 & 6.67 & 26.90 & 24.07 & 12.47 & 9.95 \\ \midrule
    \multicolumn{9}{l}{\textbf{EPD}} \\
    \quad +Solver & 10.40 & -- & 4.33 & -- & 21.74 & -- & 7.84 & -- \\
    \quad +Plugin & 10.54 & -- & 4.47 & -- & 19.02 & -- & 7.97 & -- \\
    \midrule
    \rowcolor{gray!12}
    \multicolumn{9}{l}{\textbf{MTEO (Ours)}} \\
    \rowcolor{gray!12}
    \quad +DDIM & \textbf{3.85} & \textbf{2.83} & \textbf{2.62} & \textbf{2.50} & 5.46 & 3.66 & \textbf{3.17} & \textbf{2.99} \\
    \rowcolor{gray!12}
    \quad +iPNDM & 4.83 & 3.52 & 3.01 & 2.74 & 7.46 & 5.63 & 4.31 & 3.72 \\
    \rowcolor{gray!12}
    \quad +DPM-Solver++(3M) & 3.91 & 3.11 & 2.98 & 2.66 & \textbf{5.29} & \textbf{3.65} & 3.37 & 3.13 \\
    \bottomrule
    
    \end{tabularx}

\end{table*}

Detailed FID results on CIFAR-10, FFHQ, ImageNet-64, and LSUN Bedroom are reported in~\cref{tab:cf-lw} and Appendix(ImageNet-64, LSUN Bedroom), We report the best results of MTEO. Additional metrics are provided in the Appendix. Results on MS-COCO (FID and CLIP Score) are summarized in~\cref{tab:coco-fid}.
For DiT on ImageNet-256, we evaluate MTEO on top of DDIM, which serves as a strong and widely used deterministic baseline. The metrics are compute by \textit{evaluation} file provided by~\cite{guide}.
The corresponding results are reported in~\cref{tab:dit_fid}.


\begin{table*}[tb]
\centering
\renewcommand{\arraystretch}{0.8}
\caption{FID and CLIP evaluation on MS-COCO Dataset. We omitted some of the results due to the lack of corresponding pretrained checkpoints and settings.}
\setlength{\tabcolsep}{2pt}
\begin{tabular}{l|cccc|cccc}
\toprule
Method & \multicolumn{4}{c|}{FID $\downarrow$} & \multicolumn{4}{c}{CLIP $\uparrow$} \\
\cmidrule(lr){2-9}
Step (1 Step = 2 NFE) & 3 & 4 & 5 & 6 & 3 & 4 & 5 & 6 \\
\midrule
DDIM & 36.04 & 24.02 & 19.73 & 17.69 & 25.87 & 28.35 & 29.32 & 29.79 \\
DPM-Solver++(2M) & 35.03 & 21.57 & 17.45 & 15.85 & 25.95 & 28.50 & 29.42 & 29.83 \\
\midrule

AMED-Plugin & $-$ & 18.92 & $-$ & 14.84 & $-$ & $-$ & $-$ & $-$ \\
EPD-Solver & $-$ & 16.46 & $-$ & \textbf{13.14} & $-$ & $-$ & $-$ & $-$ \\
\midrule

\rowcolor{gray!15} +DDIM & \textbf{14.93} & \textbf{12.92} & 13.65 & 13.53 & \textbf{28.86} & \textbf{29.65} & 29.85 & 30.01 \\
\rowcolor{gray!15} +DPM++(2M) & 16.12 & 13.39 & \textbf{13.18} & 13.57 & 28.61 & 29.56 & \textbf{29.89} & \textbf{30.05} \\
\bottomrule
\end{tabular}
\label{tab:coco-fid}
\end{table*}




\begin{table}[tb]
  \centering
  \renewcommand{\arraystretch}{0.84}
  \begin{minipage}[t]{0.62\textwidth}
    \centering
    \caption{IS, FID, sFID, Precision and Recall evaluation on ImageNet-256 Dataset. We implement MTEO on DiT-XL-2.}
        \label{tab:dit_fid}
        \begin{tabular}{llrrrrr}
        \toprule
        \textbf{Method} & \textbf{Metric} & \textbf{NFE=3} & \textbf{4} & \textbf{5} & \textbf{6} \\
        \midrule
        \multirow{5}{*}{DDIM} & IS & 12.82 & 32.22 & 59.53 & 90.81 \\
         & FID & 108.08 & 77.82 & 52.52 & 34.36 \\
         & sFID & 94.32 & 63.94 & 41.64 & 27.72 \\
         & Precision & 11.13\% & 23.46\% & 36.05\% & 47.37\% \\
         & Recall & 36.04\% & 42.05\% & 44.85\% & 46.30\% \\
        \midrule
        \multirow{5}{*}{MTEO} & IS & 133.65 & 181.52 & 197.59 & 211.00 \\
         & FID & 15.75 & 7.22 & 5.17 & 4.13 \\
         & sFID & 12.51 & 6.92 & 5.42 & 5.00 \\
         & Precision & 60.57\% & 70.38\% & 73.02\% & 75.47\% \\
         & Recall & 54.83\% & 58.00\% & 59.16\% & 59.51\% \\
        \bottomrule
        \end{tabular}
  \end{minipage}
  \hfill
  \begin{minipage}[t]{0.34\textwidth}
    \centering
        \caption{MTEO+DDIM employ on few-step and sampling with medium-step.}
        \label{tab:cifar10-skip}
        \begin{tabular}{c@{\hspace{0.7em}}cr}
            \toprule
            Steps & Training Steps & FID \\
            \midrule
            4 & -- & 93.36 \\
            5 & -- & 67.40 \\
            6 & -- & 49.66 \\
            7 & -- & 36.08 \\
            11 & -- & 15.68 \\
            13 & -- & 11.89 \\ \midrule
            13 & 4 & 3.52 \\
            13 & 5 & 2.40 \\
            11 & 6 & 2.33 \\
            13 & 7 & 2.26 \\
            \bottomrule
        \end{tabular}
  \end{minipage}
\end{table}

\subsubsection{Comparison with Distillation-Based Acceleration.}
We further compare MTEO (applied to DDIM) with state-of-the-art distillation-based acceleration methods on CIFAR-10, ImageNet-64, and LSUN Bedroom, including SFD/SFD-v~\cite{sfd}, CTM~\cite{ctm}, DMD-cond~\cite{dmd}, and ECM~\cite{ect}. As summarized in~\cref{tab:cifar-db} and Appendix(ImageNet-64, LSUN Bedroom), we report not only FID but also the model size, the number of trainable parameters, the percentage of trainable parameters relative to the original model, and the wall-clock training time (measured on solo RTX 3090). Generating 256 teacher trajectories with NFE=21 costs 0.17 minutes on RTX3090, which is <0.003\% of embedding optimization time. The comparison highlights a clear operating regime where MTEO is particularly attractive: while distillation methods typically require updating a large fraction (or the entirety) of the model parameters and incur substantial training cost, MTEO fine-tunes only a tiny set of time-embedding parameters (often well below 0.2\% of the backbone size) and can reach competitive quality with markedly lower training overhead. In our experiments, MTEO reduces training time by a large margin (ranging from $\sim$2$\times$ to $\sim$1000$\times$, depending on the baseline and target NFE), while remaining lightweight by design.
We compute model sizes and trainable-parameter sizes directly from the official pretrained checkpoints and the embedding files produced by MTEO.
The Appendix details (i) how we normalize training time across different GPU types when necessary and (ii) the sources used for distillation baselines’ runtime statistics.

\begin{table*}[tb]
\caption{Comparison MTEO against Distillation-Based acceleration on CIFAR-10 Dataset. We report +DDIM as main results. Hours denote solo RTX3090 GPU hour. The Model Size denote the size of original teacher model and The Parameter indicates the parameter count used for distillation in the corresponding method.}
\label{tab:cifar-db}
\centering

\renewcommand{\arraystretch}{0.8}
\begin{tabularx}{\textwidth}{l >{\centering\arraybackslash}X >{\centering\arraybackslash}X >{\centering\arraybackslash}X r r r}

\toprule
\textbf{Method} & \textbf{Model Size}  & \textbf{Parameter} & \textbf{Percent}  & \textbf{NFE} & \textbf{FID} & \textbf{Hours}  \\
\midrule
DMD & 212.83MB & 1464.32MB  & 688\% & 1 & 2.66 & - \\
CTM & 212.83MB & 460.80MB   & 216\% & 1 & 1.98 & 153.70 \\
ECM & 212.83MB & 212.83MB  & 100\% & 1 & 3.60 & 709.52 \\
ECM & 212.83MB & 212.83MB  & 100\% & 2 & 2.11 & 709.52 \\ 
SFD & 212.83MB & 212.83MB  & 100\% & 2 & 4.53 & 1.19 \\
SFD & 212.83MB & 212.83MB   & 100\% & 3 & 3.58 & 1.69 \\
SFD & 212.83MB & 212.83MB  & 100\% & 4 & 3.24 & 2.17 \\
SFD & 212.83MB & 212.83MB   & 100\% & 5 & 3.06 & 2.63 \\
SFD-v & 212.83MB & 212.83MB   & 100\% & 2 & 4.28 & 7.84 \\
SFD-v & 212.83MB & 212.83MB   & 100\% & 3 & 3.50 & 7.84 \\ 
SFD-v & 212.83MB & 212.83MB   & 100\% & 4 & 3.18 & 7.84 \\
SFD-v & 212.83MB & 212.83MB   & 100\% & 5 & 2.95 & 7.84 \\ \midrule
MTEO & 212.83MB & 0.27MB  & 0.125\%  & 3 & 3.85 & 0.13 \\
MTEO & 212.83MB & 0.33MB  & 0.156\%  & 4 & 2.83 & 0.18 \\
MTEO & 212.83MB & 0.40MB  & 0.188\%  & 5 & 2.62 & 0.19 \\
MTEO & 212.83MB & 0.47MB  & 0.219\%  & 6 & 2.50 & 0.22 \\
\bottomrule
\end{tabularx}

\end{table*}

\subsection{Ablation Study}
This section include follows: \textit{Batch Size, Training Set Size, Training Heuristics, Sensibility to Steps of Teacher, Comparison with a Shared (Single) Time Embedding, and Which Time Steps Matter}. Additional ablations are reported in Appendix; here we focus on step-importance and Shared (Single) Time Embedding, which directly supports our design.

\subsubsection{Which Time Steps Matter.}

\begin{figure}[tb]
  \centering
  \begin{minipage}[r]{0.48\textwidth}
    \centering
    \begin{subfigure}[t]{0.48\linewidth}
      \includegraphics[width=\linewidth]{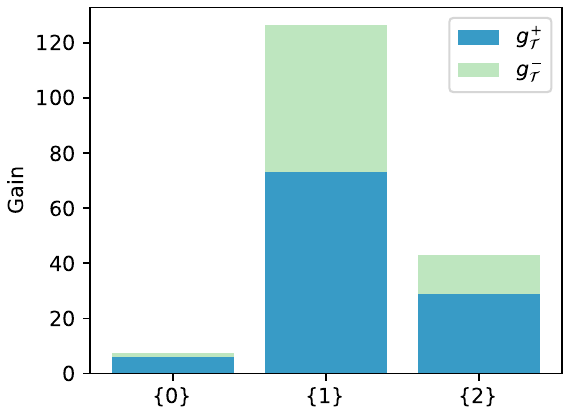}
      \caption{Step 4}
    \end{subfigure}\hfill
    \begin{subfigure}[t]{0.48\linewidth}
      \includegraphics[width=\linewidth]{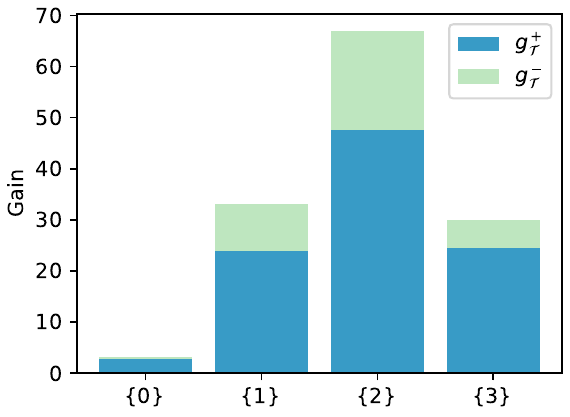}
      \caption{Step 5}
    \end{subfigure}   
    \begin{subfigure}[t]{0.48\linewidth}
      \includegraphics[width=\linewidth]{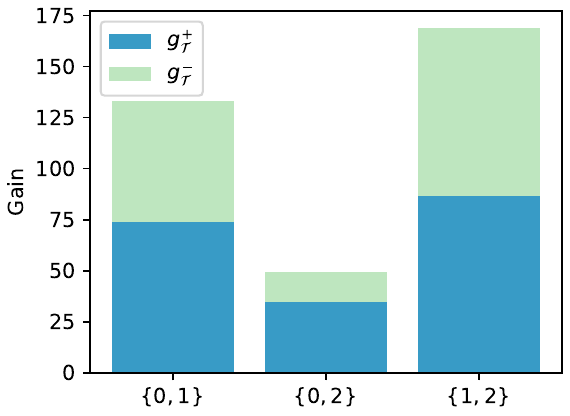}
      \caption{Step 4}
    \end{subfigure}\hfill
    \begin{subfigure}[t]{0.48\linewidth}
      \includegraphics[width=\linewidth]{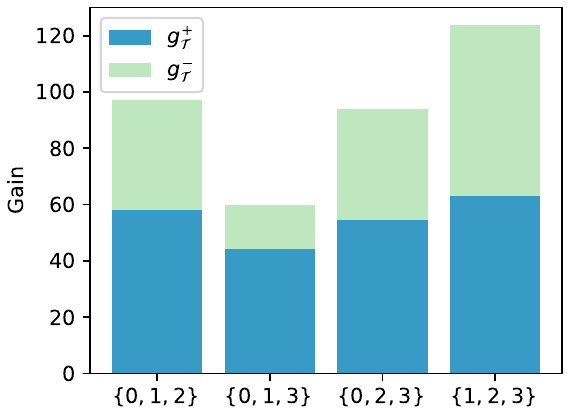}
      \caption{Step 5}
    \end{subfigure}
  \end{minipage}
  \hfill
  \begin{minipage}[c]{0.47\textwidth}
    \centering
    \begin{subfigure}[t]{\linewidth}
      \includegraphics[width=\linewidth]{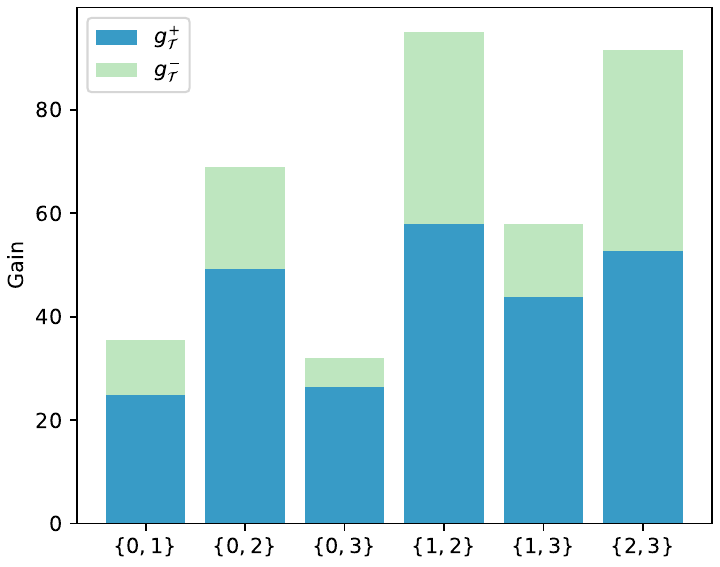}
      \caption{Step 5}
    \end{subfigure}
  \end{minipage} 
  \caption{visualization of both Gain $g^{+}_{\mathcal{T}}$ and Drop $g^{-}_{\mathcal{T}}$ view. The x axis denote subset $\mathcal{T}$.}
  \label{fig:gain}
\end{figure}

Given a sampling trajectory with N time steps indexed by 
$$\{0,1,\dots,N-1\}$$, we select a subset of time-step indices 
$\mathcal{T} \subseteq \mathbb{U}=\{0,1,\dots,N-2\}$ for optimization. 
Each element in $\mathcal{T}$ corresponds to a distinct time step at which additional optimization is performed. We analyze the importance (and necessity) of step-specific embeddings by evaluating MTEO from two complementary perspectives.
\paragraph{Gain view.} We enable MTE only on a subset of steps and use the original time embedding for all other steps, measuring how much this partial MTE improves over the baseline sampler.
\paragraph{Drop view.} Starting from the full MTE setting, we \emph{remove} MTE from a subset of steps $\mathcal{T}$ while keeping MTE on the remaining steps $\bar{\mathcal{T}}=\mathbb{U} \setminus \mathcal{T}$, and measure the resulting degradation relative to full MTE.
Formally, we define
\begin{equation}
    g^{+}_{\mathcal{T}} = FID_{\emptyset} - FID_{\mathcal{T}},
    \qquad
    g^{-}_{\mathcal{T}} = FID_{\bar{\mathcal{T}}} - FID_{\mathbb{U}},
\end{equation}
where $FID_{\emptyset}$ denotes the baseline sampler without MTE, $FID_{\mathbb{U}}$ denotes full MTEO, and $FID_{\mathcal{T}}$ denote using MTE on the corresponding step subsets.
Under both views, larger $g^{+}_{\mathcal{T}}$ or $g^{-}_{\mathcal{T}}$ indicates that the subset $\mathcal{T}$ contributes more positively to sampling quality.
The results in~\cref{fig:gain} show consistent positive contributions across steps under both views, supporting that MTEO draws benefits from \emph{multiple} steps rather than relying on a single critical step.
Finally, we test whether MTE trained on a few-step schedule can transfer to a denser multi-step schedule that includes the same time points.
As shown in~\cref{tab:cifar10-skip}, few-step embeddings still provide improvements, suggesting that MTE captures transferable correction signals rather than overfitting to a single step count. Details are list in Appendix.


\begin{table}[!b]
  \begin{minipage}[t]{0.42\textwidth}
  \vspace{2pt}
    \subsubsection{Comparison with a Shared Time Embedding.}
    To isolate the effect of the \emph{multi-layer} design, we train a strong single-embedding baseline under the same trajectory-distillation configuration.
    Concretely, at each sampling step, we learn \emph{one} time embedding vector and broadcast it to all layers (i.e., the embedding is step-specific but not layer-specific). The comparison is reported in~\cref{tab:cifar-single}.
    
  \end{minipage}
  \hfill
  \begin{minipage}[t]{0.49\textwidth}
  {\setlength{\belowcaptionskip}{0.01pt}
  \captionsetup{width=0.9\textwidth}
        \caption{Ablation of single-layer}
        \centering
        \renewcommand{\arraystretch}{0.4}
        \newcommand{\mteo}[1]{\cellcolor{gray!15}\textbf{#1}}
        \newcommand{\single}[1]{\cellcolor{gray!5}#1}
        
        \begin{tabular}{l c rrr}
        \toprule
        \multirow{2}{*}{Method} & \multirow{2}{*}{NFE} &
        \multicolumn{3}{c}{FID $\downarrow$}  \\
        \cmidrule(lr){3-5}
         & & vanilla & +single & +MTEO \\
        \midrule
        \multirow{4}{*}{DDIM}
         & 3 & 93.36 & \single{15.60}  & \mteo{5.29}\\
         & 4 & 67.40 & \single{10.43}  & \mteo{3.91}\\
         & 5 & 49.66 & \single{6.81}   & \mteo{3.02}\\
         & 6 & 36.08 & \single{5.18}   & \mteo{2.87}\\
        \midrule
        \multirow{4}{*}{iPNDM}
         & 3 & 47.98 & \single{18.75} & \mteo{6.63}\\
         & 4 & 24.81 & \single{10.06} & \mteo{4.81}\\
         & 5 & 13.59 & \single{8.14}  & \mteo{4.20}\\
         & 6 & 7.05 & \single{5.30}   & \mteo{3.26}\\
        \midrule
        \multirow{4}{*}{DPM++3M}
         & 3 & 110.00 & \single{24.77} & \mteo{5.54} \\
         & 4 & 46.52 & \single{12.95} & \mteo{3.87} \\
         & 5 & 24.97 & \single{8.18}  & \mteo{3.38} \\
         & 6 & 11.99 & \single{5.34}  & \mteo{3.60} \\
        \bottomrule
        \end{tabular}
        \label{tab:cifar-single}
        }
  \end{minipage}
  
\end{table}





%
%
\bibliographystyle{splncs04}
\bibliography{main}

\clearpage

\appendix

\section{Complementary Background}
\subsection{Overview of Diffusion Models}
We consider a continuous-time diffusion process $\{x_t\}_{t\in[0,1]}$ that gradually perturbs an initial data sample $x_0 \sim p_{\text{data}}(x)$ into pure noise as time $t$ increases. The forward diffusion (noise-adding process) can be formalized by the $It\bar{o}$ SDE~\cite{SDE}:
\begin{equation}
    \label{eq:forward_sde}
    \mathrm{d}x_t = f(x_t,t)\mathrm{d}t + g(t)\mathrm{d}w_t,
\end{equation}
where $f(x_t,t)$ is a drift term, $g(t)\ge 0$ is the diffusion coefficient, and $w_t$ is a standard Wiener process. Let $p_t(x)$ denote the marginal density of $x_t$. Under mild regularity conditions, one can derive a corresponding reverse-time SDE that transforms noise back into data:
\begin{equation}
    \label{eq:reverse_sde}
    \mathrm{d}x_t = \Big[f(x_t,t) - g(t)^2\nabla_x \log p_t(x_t)\Big]\mathrm{d}t + g(t)\mathrm{d}\bar w_t,
\end{equation}
where $\bar w_t$ is a standard Wiener process running backward in time, and $\nabla_x \log p_t(x_t)$ is the \textit{score function} of the distribution. In practice, since $p_t(x)$ is intractable, training a neural network $s_\theta(x_t, t)$ to approximate the score $\nabla_x \log p_t(x_t)$ via denoising score matching on noisy data is a solution. At inference time, one can generate new samples by starting from $x_{T} \sim \mathcal{N}(0,I)$ and numerically integrating either the reverse SDE \cref{eq:reverse_sde} (a stochastic generative process) or an equivalent ODE that shares the same marginals. This deterministic counterpart, known as the \textit{probability-flow ODE}~\cite{SDE}, is given by
\begin{equation}
    \label{eq:pf_ode}
    \frac{\mathrm{d}x_t}{\mathrm{d}t} = f(x_t,t)-\frac{1}{2}g(t)^2s_\theta(x_t,t),
\end{equation}
and yields trajectories with the same distribution as the SDE. Solving this ODE corresponds to a deterministic sampling procedure. For example, the Denoising Diffusion Implicit Models (DDIM)~\cite{DDIM} sampler can be viewed as a first-order discretization of \cref{eq:pf_ode}. In either case, the integration from $x_{T}$ to $x_0$ is carried out in a finite number of steps, each requiring an evaluation of the network $s_\theta$. We refer to the total number of model evaluations as the \textbf{number of function evaluations (NFE)}. Achieving high sample fidelity typically requires a large NFE (often hundreds of evaluations), which leads to slow sampling. To mitigate this issue, a variety of \textit{training-free} acceleration techniques have been explored. These methods focus on improving the sampler itself without modifying the learned model. For example, one can apply higher-order numerical solvers or use optimized time-step schedules to integrate \cref{eq:reverse_sde} or \cref{eq:pf_ode} more efficiently. The DDIM sampler is a popular first-order method that can reduce sampling time compared to the original diffusion process, but it still usually needs on the order of hundreds of steps for high-quality results. More recently, dedicated high-order solvers such as \textit{DPM-Solver}~\cite{DPM} have been developed to further cut down the required steps, often achieving acceptable image quality with only about $10$--$20$ function evaluations. Nevertheless, even these advanced samplers suffer noticeable degradation in fidelity when the NFE budget is extremely low (e.g. $\sim$10), due to accumulating integration error. This limitation motivates \textit{training-based} approaches that learn to preserve diffusion model performance under very low NFE regimes. 

\subsection{Trajectory Distillation}
Maintaining high generation quality with very few sampling steps has prompted the development of several \textit{training-based} acceleration methods that augment the diffusion model through an extra training phase. Notable examples include \textit{progressive distillation}~\cite{progressiveDistillation}, which gradually distills a model to use fewer steps by repeatedly halving the number of steps, and \textit{consistency distillation}~\cite{cm,ctm}, which trains the model to produce identical outputs whether one uses many small steps or a single large step. Despite their differing formulations, these methods are all built upon the core concept of \textbf{trajectory distillation}. Trajectory distillation aims to use a high-NFE ``teacher'' trajectory to guide a low-NFE ``student'' model. In essence, one first generates a reference trajectory of intermediate states using a large number of steps, and then trains a student model so that when it uses a much smaller number of steps, its intermediate results mimic the teacher’s trajectory. Concretely, let $\{\hat{x}_{t_i}\}_{i=0}^{N-1}$ denote the states along a high-NFE trajectory (with $N$ steps) obtained from the teacher process. The student model is optimized such that for a reduced number of steps $N$ (much smaller integration intervals), it produces states ${x_{t_i}}$ that stay close to the teacher’s states ${\hat{x}_{t_i}}$ at the corresponding time points. By aligning the student’s denoising trajectory with the teacher’s trajectory, the final sample quality can be preserved even under aggressive step reduction.~\cref{alg:trajectory distillation} provides a pseudocode of the trajectory distillation training procedure. 

\begin{algorithm}[!h]
\renewcommand{\algorithmicrequire}{\textbf{Input:}}
\renewcommand{\algorithmicensure}{\textbf{Output:}}
\caption{Trajectory Distillation}
\label{alg:trajectory distillation}
\begin{algorithmic}[1]
\REQUIRE Teacher trajectory $\{\hat{x}_{t_i}\}_{i=0}^{N-1}$, ODE solver $S$, student model $s_\theta$, number of steps $N$, loss function $\mathcal{L}$, distance metric $\mathcal{D}$.
\STATE Initialize $x_{t_0} = \hat{x}_{t_0}$
\FOR{$i = 0$ to $N-2$}
\REPEAT
\STATE $x_{t_{i+1}} = S\big(x_{t_i},s_\theta,t_i,t_{i+1}\big)$
\STATE $\mathcal{L} = \mathcal{D}\big(x_{t_{i+1}},\hat{x}_{t_{i+1}}\big)$
\STATE Update $\theta$ by taking $\nabla_{\theta}\mathcal{L}$ (optimize student)
\UNTIL{converged}
\STATE $x_{t_{i+1}} = S\big(x_{t_i},s_\theta,t_i,t_{i+1}\big)$ \hfill // advance to next step
\ENDFOR
\RETURN Trained student model $s_\theta$
\end{algorithmic}
\end{algorithm}








\section{Experiments}
 We present all remaining experimental results. In \cref{app:exp-a} we report comparisons on the remaining datasets (ImageNet-64 and LSUN Bedroom). In \cref{app:exp-b} we provide more detailed results, including additional evaluation metrics. In \cref{app:config} we give the full experimental configurations and hyperparameter details, include trainging loss curve and intermediate visualization.

\subsection{Additional Results}
\label{app:exp-a}
We provide the comparation of lightweight methods for ImageNet-64 and LSUN Bedroom in~\cref{tab:ib-lw}, and for comparation of distillation-based acceleration, are in ~\cref{tab:imagenet64-db} and~\cref{tab:lsun-bedroom-db}, respectively.

\begin{table*}[htb]
    \caption{FID evaluation on ImageNet-64 and LSUN Bedroom Datasets}
    \label{tab:ib-lw}
    \centering

    \setlength{\tabcolsep}{2pt} 

    \begin{tabularx}{\textwidth}{l|*{4}{>{\centering\arraybackslash}X}|*{4}{>{\centering\arraybackslash}X}}

    \toprule
    Method & \multicolumn{4}{c|}{ImageNet-64} & \multicolumn{4}{c}{LSUN Bedroom} \\
    \cmidrule{2-9}
           & \multicolumn{8}{c}{NFE} \\
    \cmidrule{2-9}
           & 3 & 4 & 5 & 6 & 3 & 4 & 5 & 6 \\
    \midrule
    DDIM & 75.91 & 52.53 & 43.86 & 24.33 & 86.13 & 54.53 & 34.34 & 25.26 \\
    iPNDM & 52.17 & 28.95 & 15.61 & 10.60 & 80.99 & 43.89 & 26.65 & 20.71 \\
    UniPC & 85.31 & 49.69 & 20.87 & 12.02 & 112.3 & 49.00 & 23.20 & 11.64 \\
    DEIS & 40.31 & 20.26 & 12.45 & 8.94 & 52.46 & 23.65 & 14.42 & 10.74 \\ \midrule
    \multicolumn{9}{l}{\textbf{DPM}} \\
    \quad +Solver-2 & 140.20 & 129.74 & 42.41 & 43.48 & 210.60 & 210.41 & 80.60 & 80.19 \\
    \quad +Solver++(3M) & 91.52 & 56.31 & 25.49 & 15.05 & 111.90 & 49.46 & 23.15 & 12.28 \\

    \multicolumn{9}{l}{\textbf{AMED}} \\
    \quad +Solver & 38.10 & 32.69 & 10.74 & 10.63 & 58.21 & -- & 13.20 & -- \\
    \quad +Plugin & 28.06 & 23.55 & 13.83 & 12.05 & 101.50 & -- & 25.68 & -- \\ \midrule
    \multicolumn{9}{l}{\textbf{EPD}} \\
    \quad +Solver & 18.28 & -- & 6.35 & -- & 13.21 & -- & 7.52 & -- \\
    \quad +Plugin & 19.89 & -- & 8.17 & -- & 14.12 & -- & 8.26 & -- \\
    \midrule
    \rowcolor{gray!12}
    \multicolumn{9}{l}{\textbf{MTEO (Ours)}} \\
    \rowcolor{gray!12}
    \quad +DDIM & 7.42 & \textbf{4.87} & \textbf{4.11} & \textbf{3.81} & \textbf{12.99} & \textbf{7.01} & \textbf{6.03} & \textbf{4.44} \\
    \rowcolor{gray!12}
    \quad +iPNDM & 9.47 & 5.46 & 4.72 & 4.12 & 23.71 & 13.43 & 8.46 & 6.25 \\
    \rowcolor{gray!12}
    \quad +DPM-Solver++(3M) & \textbf{7.07} & 5.10 & 4.66 & 4.19 & 20.77 & 14.94 & 6.56 & 6.45 \\
    \bottomrule

    \end{tabularx}

\end{table*}

\begin{table*}[htbp]
\caption{Comparison MTEO against Distillation-Based acceleration on ImageNet-64 Dataset. We report +DDIM as main results. Hours denote solo RTX3090 GPU hour.}
\label{tab:imagenet64-db}
\centering
\renewcommand{\arraystretch}{0.65}
\setlength{\tabcolsep}{2pt}
\begin{tabularx}{\textwidth}{l >{\centering\arraybackslash}X >{\centering\arraybackslash}X >{\centering\arraybackslash}X r r r}

\toprule
\textbf{Method} & \textbf{Model Size} & \textbf{Parameter} & \textbf{Percent} & \textbf{NFE} & \textbf{FID} & \textbf{Hours} \\
\midrule
DMD-cond & -- & -- & -- & 1 & 2.62 & -- \\
CTM & 1126.4MB & 1228.8MB & 109\% & 1 & 2.06 & 1660.86 \\
ECM-S & 1126.4MB & 1126.4MB & 100\% & 1 & 5.51 & 76.50 \\
ECM-S & 1126.4MB & 1126.4MB & 100\% & 2 & 3.18 & 76.50 \\
SFD & 1126.4MB & 1126.4MB & 100\% & 2 & 10.25 & 6.15 \\
SFD & 1126.4MB & 1126.4MB & 100\% & 3 & 6.35 & 8.53 \\
SFD & 1126.4MB & 1126.4MB & 100\% & 4 & 4.99 & 11.07 \\
SFD & 1126.4MB & 1126.4MB & 100\% & 5 & 4.33 & 13.17 \\
SFD-v & 1126.4MB & 1126.4MB & 100\% & 2 & 9.47 & 43.49 \\
SFD-v & 1126.4MB & 1126.4MB & 100\% & 3 & 5.78 & 43.49 \\ 
SFD-v & 1126.4MB & 1126.4MB & 100\% & 4 & 4.72 & 43.49 \\
SFD-v & 1126.4MB & 1126.4MB & 100\% & 5 & 4.21 & 43.49 \\ \midrule
MTEO  & 1126.4MB & 0.42MB & 0.037\% & 3 & 7.42 & 3.32 \\
MTEO  & 1126.4MB & 0.53MB & 0.047\% & 4 & 4.87 & 3.65 \\
MTEO  & 1126.4MB & 0.63MB & 0.056\% & 5 & 4.11 & 4.06 \\
MTEO  & 1126.4MB & 0.74MB & 0.066\% & 6 & 3.81 & 4.32 \\
\bottomrule
\end{tabularx}

\end{table*}
\begin{table*}[htbp]
\caption{Comparison MTEO against Distillation-Based acceleration on LSUN Bedroom Dataset. We report +DDIM as main results.}
\label{tab:lsun-bedroom-db}
\centering
\renewcommand{\arraystretch}{0.65}
\setlength{\tabcolsep}{2pt}
\begin{tabularx}{\textwidth}{l >{\centering\arraybackslash}X >{\centering\arraybackslash}X >{\centering\arraybackslash}X r r}

\toprule
\textbf{Method} & \textbf{Model Size} & \textbf{Parameter} & \textbf{Percent} & \textbf{NFE} & \textbf{FID} \\
\midrule
SFD & 2007.04MB & 2007.04MB & 100\% & 2 & 10.39 \\
SFD & 2007.04MB & 2007.04MB & 100\% & 3 & 6.42 \\
SFD & 2007.04MB & 2007.04MB & 100\% & 4 & 5.26 \\
SFD & 2007.04MB & 2007.04MB & 100\% & 5 & 4.73 \\
SFD-v & 2007.04MB & 2007.04MB & 100\% & 2 & 9.25 \\
SFD-v & 2007.04MB & 2007.04MB & 100\% & 3 & 5.36 \\
SFD-v & 2007.04MB & 2007.04MB & 100\% & 4 & 4.63 \\
SFD-v & 2007.04MB & 2007.04MB & 100\% & 5 & 4.33 \\ \midrule
MTEO  & 2007.04MB & 0.58MB & 0.029\% & 3 & 7.02 \\
MTEO  & 2007.04MB & 0.72MB & 0.036\% & 4 & 5.32 \\
MTEO  & 2007.04MB & 0.87MB & 0.043\% & 5 & 5.93 \\
MTEO  & 2007.04MB & 1.01MB & 0.050\% & 6 & 5.98 \\
\bottomrule
\end{tabularx}

\end{table*}

\begin{table}[!htbp]
\centering
\caption{Detailed FID results on LSUN Bedroom}
\label{tab:bedroom-detail}
\renewcommand{\arraystretch}{0.65}
\begin{tabular}{l|l>{\hspace{2.0em}}r<{\hspace{2.0em}}|rrrr}
\toprule
Method   & Schedule & $\rho$ & \multicolumn{4}{c}{NFE} \\
\cmidrule(lr){4-7}
 & & & 3 & 4 & 5 & 6 \\
\midrule
DDIM & polynomial & 7 & 86.13 & 54.53 & 34.34 & 25.26 \\
iPNDM & polynomial & 7 & 80.99 & 43.89 & 26.65 & 20.71 \\
DPM-Solver++(3M) & logsnr & -- & 111.90 & 49.46 & 23.15 & 12.28 \\ \midrule

\rowcolor{gray!15}
\multicolumn{7}{l}{\textbf{MTEO (Ours)}} \\

\rowcolor{gray!15}
\quad +DDIM & polynomial & 7 & 12.99 & 7.01 & 6.03 & 4.44 \\

\rowcolor{gray!15}
\quad +iPNDM & polynomial & 7 & 23.71 & 13.43 & 8.46 & 6.25 \\

\rowcolor{gray!15}
\quad +DPM-Solver++(3M) & logsnr & -- & 20.77 & 14.94 & 6.56 & 6.45 \\
\midrule

\rowcolor{gray!15}
\multicolumn{7}{l}{\textbf{MTEO (Ours)}} \\

\rowcolor{gray!15}
\quad +DDIM & time\_uniform & 1 & 7.02 & 5.32 & 5.93 & 5.98 \\

\rowcolor{gray!15}
\quad +iPNDM & time\_uniform & 1 & 13.94 & 9.57 & 7.41 & 6.02 \\

\rowcolor{gray!15}
\quad +DPM-Solver++(3M) & time\_uniform & 1 & 6.23 & 5.37 & 5.93 & 7.03 \\
\bottomrule
\end{tabular}
\end{table}

\subsection{Detailed Results}
\label{app:exp-b}
We provide the detailed comparation of CIFAR-10, FFHQ, ImageNet-64 and LSUN Bedroom in~\cref{tab:cifar-detailed},~\cref{tab:ffhq-detail},~\cref{tab:imagenet64-detail} and~\cref{tab:bedroom-detail}, respectively.
\begin{table}[htbp]
\centering
\captionsetup{width=32em}
\caption{Detailed FID results on CIFAR-10. We report different types of schedule on +DDIM, +iPNDM and +DPM++(3M). The time\_uniform2 schedule perform best which also report as main result.}
\label{tab:cifar-detailed}
\begin{tabular}{l|l>{\hspace{2.0em}}r<{\hspace{2.0em}}|rrrr}
\toprule
Method   & Schedule & $\rho$ & \multicolumn{4}{c}{NFE} \\
\cmidrule(lr){4-7}
 & & & 3 & 4 & 5 & 6 \\
\midrule
DDIM & polynomial & 7 & 93.36 & 67.40 & 49.66 & 36.08 \\
iPNDM & polynomial & 7 & 47.98 & 24.81 & 13.59 & 7.05 \\
DPM-Solver++(3M) & logsnr & -- & 110.00 & 46.52 & 24.97 & 11.99 \\  

\rowcolor{gray!15}\multicolumn{7}{>{\columncolor{gray!15}}l}{\textbf{MTEO (Ours)}} \\

\rowcolor{gray!15}
\quad +DDIM & polynomial & 7 & 5.29 & 3.91 & 3.02 & 2.87 \\

\rowcolor{gray!15}
\quad +iPNDM & polynomial & 7 & 6.63 & 4.81 & 4.20 & 3.26 \\

\rowcolor{gray!15}
\quad +DPM-Solver++(3M) & logsnr & -- & 5.54 & 3.87 & 3.38 & 3.60 \\ 
\midrule
DDIM & logsnr & -- & 121.70 & 73.20 & 53.53 & 38.19 \\
iPNDM & logsnr & -- & 88.36 & 35.58 & 19.86 & 10.68 \\
DPM-Solver++(3M) & polynomial & 7 & 70.03 & 50.39 & 31.66 & 17.89 \\ 

\rowcolor{gray!15}\multicolumn{7}{>{\columncolor{gray!15}}l}{\textbf{MTEO (Ours)}} \\

\rowcolor{gray!15}
\quad +DDIM & logsnr & -- & 6.40 & 4.06 & 3.52 & 3.17 \\

\rowcolor{gray!15}
\quad +iPNDM & logsnr & -- & 6.15 & 3.75 & 3.09 & 2.80 \\

\rowcolor{gray!15}
\quad +DPM-Solver++(3M) & polynomial & 7 & 5.93 & 4.85 & 3.96 & 3.51 \\
\midrule
DDIM & time\_uniform & 2 & 74.51 & 42.62 & 28.79 & 21.21 \\
iPNDM & time\_uniform & 2 & 39.91 & 15.35 & 9.02 & 5.26 \\
DPM-Solver++(3M) & time\_uniform & 2 & 64.98 & 29.92 & 15.47 & 9.41 \\

\rowcolor{gray!15}\multicolumn{7}{>{\columncolor{gray!15}}l}{\textbf{MTEO (Ours)}} \\

\rowcolor{gray!15}
\quad +DDIM & time\_uniform & 2 & 3.85 & 2.83 & 2.62 & 2.50 \\

\rowcolor{gray!15}
\quad +iPNDM & time\_uniform & 2 & 4.83 & 3.52 & 3.01 & 2.74 \\

\rowcolor{gray!15}
\quad +DPM-Solver++(3M) & time\_uniform & 2 & 3.91 & 3.11 & 2.98 & 2.66 \\
\bottomrule
\end{tabular}
\end{table}

\begin{table}[htbp]
\centering
\captionsetup{width=32em}
\caption{Detailed FID results on FFHQ. We report different types of schedule on +DDIM, +iPNDM and +DPM++(3M). The time\_uniform schedule perform best which also report as main result.}
\label{tab:ffhq-detail}
\begin{tabular}{l|l>{\hspace{2.0em}}r<{\hspace{2.0em}}|rrrr}
\toprule
Method & Schedule & $\rho$ & \multicolumn{4}{c}{NFE} \\
\cmidrule(lr){4-7}
 & & & 3 & 4 & 5 & 6 \\
\midrule
DDIM & polynomial & 7 & 78.16 & 57.37 & 43.85 & 35.15 \\
iPNDM & polynomial & 7 & 45.90 & 28.21 & 17.14 & 10.00 \\
DPM-Solver++(3M) & logsnr & -- & 86.45 & 45.94 & 22.51 & 13.74 \\
\rowcolor{gray!12}
\multicolumn{7}{l}{\textbf{MTEO (Ours)}} \\
\rowcolor{gray!12}
\quad +DDIM & polynomial & 7 & 9.84 & 6.05 & 4.45 & 3.76 \\
\rowcolor{gray!12}
\quad +iPNDM & polynomial & 7 & 12.97 & 9.82 & 7.95 & 5.83 \\
\rowcolor{gray!12}
\quad +DPM-Solver++(3M) & logsnr & -- & 9.46 & 9.17 & 5.01 & 4.43 \\
\midrule
\rowcolor{gray!12}
\multicolumn{7}{l}{\textbf{MTEO (Ours)}} \\
\rowcolor{gray!12}
\quad +DDIM & time\_uniform & 2 & 5.46 & 3.66 & 3.17 & 2.99 \\
\rowcolor{gray!12}
\quad +iPNDM & time\_uniform & 2 & 7.46 & 5.63 & 4.31 & 3.72 \\
\rowcolor{gray!12}
\quad +DPM-Solver++(3M) & time\_uniform & 2 & 5.29 & 3.65 & 3.37 & 3.13 \\
\bottomrule
\end{tabular}
\end{table}
\begin{table}[htbp]
  \centering
  \renewcommand{\arraystretch}{0.65}
  \caption{IS, FID, sFID, Precision and Recall evaluation on ImageNet-64 Dataset.}
  \label{tab:imagenet64-detail}
  \begin{tabular}{llr@{\hspace{1.2em}}lrrrr}
  \toprule
  \textbf{Method} & \textbf{schedule} & $\rho$    & \textbf{Metric} & \textbf{NFE=3} & \textbf{4} & \textbf{5} & \textbf{6} \\
  \midrule
  \multirow{5}{*}{DDIM} & \multirow{5}{*}{polynomial} & \multirow{5}{*}{7} & IS & 11.68 & 16.25 & 20.49 & 24.40 \\
   &  &  & FID & 75.91 & 52.53 & 43.86 & 24.33 \\
   &  &  & sFID & 70.94 & 47.43 & 33.80 & 26.53 \\
   &  &  & Precision & 29.25\% & 38.20\% & 45.81\% & 51.15\% \\
   &  &  & Recall & 30.24\% & 38.81\% & 44.21\% & 48.50\% \\
  \midrule
  \multirow{5}{*}{iPNDM} & \multirow{5}{*}{polynomial} & \multirow{5}{*}{7} & IS & 15.37 & 23.54 & 31.34 & 36.06 \\
   &  &  & FID & 52.17 & 28.95 & 15.61 & 10.60 \\
   &  &  & sFID & 43.28 & 22.94 & 12.27 & 10.76 \\
   &  &  & Precision & 37.92\% & 50.17\% & 60.43\% & 64.33\% \\
   &  &  & Recall & 41.77\% & 51.65\% & 58.27\% & 59.67\% \\
  \midrule
  \multirow{5}{*}{DPM++(3M)} & \multirow{5}{*}{logsnr} & \multirow{5}{*}{--} & IS & 10.45 & 16.25 & 25.56 & 31.64 \\
   &  &  & FID & 91.52 & 56.31 & 25.49 & 15.05 \\
   &  &  & sFID & 65.35 & 33.70 & 15.23 & 9.58 \\
   &  &  & Precision & 36.59\% & 41.53\% & 55.91\% & 62.25\% \\
   &  &  & Recall & 25.63\% & 41.79\% & 54.37\% & 59.14\% \\
  \midrule
  \multicolumn{8}{l}{\textbf{MTEO (Ours)}} \\
  \multirow{5}{*}{\quad +DDIM} & \multirow{5}{*}{polynomial} & \multirow{5}{*}{7} & IS & 40.69 & 46.10 & 46.99 & 46.73 \\
   &  &  & FID & 11.61 & 6.94 & 5.37 & 4.82 \\
   &  &  & sFID & 7.82 & 5.55 & 4.73 & 4.39 \\
   &  &  & Precision & 63.67\% & 67.92\% & 69.68\% & 69.82\% \\
   &  &  & Recall & 58.07\% & 61.00\% & 62.57\% & 63.19\% \\
  \midrule
  \multirow{5}{*}{\quad +iPNDM} & \multirow{5}{*}{polynomial} & \multirow{5}{*}{7} & IS & 37.54 & 42.78 & 45.28 & 47.11 \\
   &  &  & FID & 14.45 & 10.07 & 7.86 & 5.87 \\
   &  &  & sFID & 10.11 & 8.08 & 6.68 & 5.71 \\
   &  &  & Precision & 61.78\% & 66.09\% & 67.49\% & 69.44\% \\
   &  &  & Recall & 56.81\% & 58.65\% & 61.08\% & 62.19\% \\
  \midrule
  \multirow{5}{*}{\quad +DPM++(3M)} & \multirow{5}{*}{logsnr} & \multirow{5}{*}{--} & IS & 39.53 & 41.41 & 41.34 & 40.52 \\
   &  &  & FID & 16.45 & 13.36 & 10.72 & 7.11 \\
   &  &  & sFID & 8.12 & 5.42 & 4.74 & 4.87 \\
   &  &  & Precision & 63.32\% & 66.60\% & 67.69\% & 67.09\% \\
   &  &  & Recall & 58.12\% & 61.87\% & 62.86\% & 63.51\% \\
  \midrule
  \multirow{5}{*}{\quad +DDIM} & \multirow{5}{*}{time\_uniform} & \multirow{5}{*}{2} & IS & 43.90 & 46.31 & 46.34 & 46.89 \\
   &  &  & FID & 7.42 & 4.87 & 4.11 & 3.81 \\
   &  &  & sFID & 5.98 & 4.76 & 4.55 & 4.30 \\
   &  &  & Precision & 67.05\% & 69.74\% & 70.73\% & 71.03\% \\
   &  &  & Recall & 59.41\% & 61.53\% & 62.26\% & 62.93\% \\
  \midrule
  \multirow{5}{*}{\quad +iPNDM} & \multirow{5}{*}{time\_uniform} & \multirow{5}{*}{2} & IS & 40.51 & 44.09 & 44.73 & 45.86 \\
   &  &  & FID & 9.47 & 5.46 & 4.72 & 4.12 \\
   &  &  & sFID & 7.53 & 5.57 & 5.64 & 4.98 \\
   &  &  & Precision & 64.96\% & 68.72\% & 69.75\% & 70.15\% \\
   &  &  & Recall & 58.67\% & 61.97\% & 62.04\% & 62.88\% \\
  \midrule
  \multirow{5}{*}{\quad +DPM++(3M)} & \multirow{5}{*}{time\_uniform} & \multirow{5}{*}{2} & IS & 43.28 & 43.98 & 43.03 & 43.80 \\
   &  &  & FID & 7.07 & 5.10 & 4.66 & 4.19 \\
   &  &  & sFID & 5.80 & 4.83 & 4.65 & 4.34 \\
   &  &  & Precision & 67.69\% & 70.16\% & 70.59\% & 70.32\% \\
   &  &  & Recall & 58.98\% & 61.40\% & 63.06\% & 63.58\% \\
  \bottomrule
  \end{tabular}
\end{table}

\subsection{Configuration}
\label{app:config}

We provide the full hyperparameters, checkpoint and stats file in~\cref{tab:exp-set}, ~\cref{tab:checkpoints} and~\cref{tab:status}.

\begin{table*}[htbp]
\caption{Experimental settings for each dataset.}
\centering
\resizebox{\textwidth}{!}{%
\begin{tabular}{llrrrrrrrrrrrrr}
\toprule
Dataset & Sampler & Steps & Schedule & $\rho$ & $lr$ & $lr_{min}$ & $\varepsilon$ & $\varepsilon_{min}$ & Patience & $E_{max}$ & Train Seeds & Batch Size & Tea Sampler & Tea Steps \\
\midrule
\multirow{12}{*}{\textbf{CIFAR10/FFHQ}} 
&  \multirow{4}{*}{DDIM}  
& \cellcolor{blue!12}4 & \cellcolor{blue!12}poly/uni & \cellcolor{blue!12}7/2 & \cellcolor{blue!12}2e-2 & \cellcolor{blue!12}1e-3 & \cellcolor{blue!12}1e-2 & \cellcolor{blue!12}1e-3 & \cellcolor{blue!12}10 & \cellcolor{blue!12}300 & \cellcolor{blue!12}50000-50255 & \cellcolor{blue!12}64 & \cellcolor{blue!12}iPNDM & \cellcolor{blue!12}22 \\  
& & \cellcolor{green!12}5 & \cellcolor{green!12}poly/uni & \cellcolor{green!12}7/2 & \cellcolor{green!12}2e-2 & \cellcolor{green!12}1e-3 & \cellcolor{green!12}1e-2 & \cellcolor{green!12}1e-3 & \cellcolor{green!12}10 & \cellcolor{green!12}300 & \cellcolor{green!12}50000-50255 & \cellcolor{green!12}64 & \cellcolor{green!12}iPNDM & \cellcolor{green!12}21 \\ 
& & \cellcolor{yellow!12}6 & \cellcolor{yellow!12}poly/uni & \cellcolor{yellow!12}7/2 & \cellcolor{yellow!12}2e-2 & \cellcolor{yellow!12}1e-3 & \cellcolor{yellow!12}1e-2 & \cellcolor{yellow!12}1e-3 & \cellcolor{yellow!12}10 & \cellcolor{yellow!12}300 & \cellcolor{yellow!12}50000-50255 & \cellcolor{yellow!12}64 & \cellcolor{yellow!12}iPNDM & \cellcolor{yellow!12}21 \\ 
& & \cellcolor{orange!12}7 & \cellcolor{orange!12}poly/uni & \cellcolor{orange!12}7/2 & \cellcolor{orange!12}2e-2 & \cellcolor{orange!12}1e-3 & \cellcolor{orange!12}1e-2 & \cellcolor{orange!12}1e-3 & \cellcolor{orange!12}10 & \cellcolor{orange!12}300 & \cellcolor{orange!12}50000-50255 & \cellcolor{orange!12}64 & \cellcolor{orange!12}iPNDM & \cellcolor{orange!12}25 \\  \cline{3-15}
& \multirow{4}{*}{iPNDM}  
& \cellcolor{blue!12}4 & \cellcolor{blue!12}poly/uni & \cellcolor{blue!12}7/2 & \cellcolor{blue!12}2e-2 & \cellcolor{blue!12}1e-3 & \cellcolor{blue!12}1e-2 & \cellcolor{blue!12}1e-3 & \cellcolor{blue!12}10 & \cellcolor{blue!12}300 & \cellcolor{blue!12}50000-50255 & \cellcolor{blue!12}64 & \cellcolor{blue!12}iPNDM & \cellcolor{blue!12}22 \\
& & \cellcolor{green!12}5 & \cellcolor{green!12}poly/uni & \cellcolor{green!12}7/2 & \cellcolor{green!12}2e-2 & \cellcolor{green!12}1e-3 & \cellcolor{green!12}1e-2 & \cellcolor{green!12}1e-3 & \cellcolor{green!12}10 & \cellcolor{green!12}300 & \cellcolor{green!12}50000-50255 & \cellcolor{green!12}64 & \cellcolor{green!12}iPNDM & \cellcolor{green!12}21 \\
& & \cellcolor{yellow!12}6 & \cellcolor{yellow!12}poly/uni & \cellcolor{yellow!12}7/2 & \cellcolor{yellow!12}2e-2 & \cellcolor{yellow!12}1e-3 & \cellcolor{yellow!12}1e-2 & \cellcolor{yellow!12}1e-3 & \cellcolor{yellow!12}10 & \cellcolor{yellow!12}300 & \cellcolor{yellow!12}50000-50255 & \cellcolor{yellow!12}64 & \cellcolor{yellow!12}iPNDM & \cellcolor{yellow!12}21 \\
& & \cellcolor{orange!12}7 & \cellcolor{orange!12}poly/uni & \cellcolor{orange!12}7/2 & \cellcolor{orange!12}2e-2 & \cellcolor{orange!12}1e-3 & \cellcolor{orange!12}1e-2 & \cellcolor{orange!12}1e-3 & \cellcolor{orange!12}10 & \cellcolor{orange!12}300 & \cellcolor{orange!12}50000-50255 & \cellcolor{orange!12}64 & \cellcolor{orange!12}iPNDM & \cellcolor{orange!12}25 \\  \cline{3-15}
& \multirow{4}{*}{DPM++(3M)} 
& \cellcolor{blue!12}4 & \cellcolor{blue!12}logsnr & \cellcolor{blue!12}-- & \cellcolor{blue!12}2e-2 & \cellcolor{blue!12}1e-3 & \cellcolor{blue!12}1e-2 & \cellcolor{blue!12}1e-3 & \cellcolor{blue!12}10 & \cellcolor{blue!12}300 & \cellcolor{blue!12}50000-50255 & \cellcolor{blue!12}64 & \cellcolor{blue!12}DPM++(3M) & \cellcolor{blue!12}22 \\
& & \cellcolor{green!12}5 & \cellcolor{green!12}logsnr & \cellcolor{green!12}-- & \cellcolor{green!12}2e-2 & \cellcolor{green!12}1e-3 & \cellcolor{green!12}1e-2 & \cellcolor{green!12}1e-3 & \cellcolor{green!12}10 & \cellcolor{green!12}300 & \cellcolor{green!12}50000-50255 & \cellcolor{green!12}64 & \cellcolor{green!12}DPM++(3M) & \cellcolor{green!12}21 \\
& & \cellcolor{yellow!12}6 & \cellcolor{yellow!12}logsnr & \cellcolor{yellow!12}-- & \cellcolor{yellow!12}2e-2 & \cellcolor{yellow!12}1e-3 & \cellcolor{yellow!12}1e-2 & \cellcolor{yellow!12}1e-3 & \cellcolor{yellow!12}10 & \cellcolor{yellow!12}300 & \cellcolor{yellow!12}50000-50255 & \cellcolor{yellow!12}64 & \cellcolor{yellow!12}DPM++(3M) & \cellcolor{yellow!12}21 \\
& & \cellcolor{orange!12}7 & \cellcolor{orange!12}logsnr & \cellcolor{orange!12}-- & \cellcolor{orange!12}2e-2 & \cellcolor{orange!12}1e-3 & \cellcolor{orange!12}1e-2 & \cellcolor{orange!12}1e-3 & \cellcolor{orange!12}10 & \cellcolor{orange!12}300 & \cellcolor{orange!12}50000-50255 & \cellcolor{orange!12}64 & \cellcolor{orange!12}DPM++(3M) & \cellcolor{orange!12}25 \\
\midrule
\multirow{12}{*}{\textbf{ImageNet64}} 
& \multirow{4}{*}{DDIM} 
& \cellcolor{blue!12}4 & \cellcolor{blue!12}poly/uni & \cellcolor{blue!12}7/2 & \cellcolor{blue!12}1e-3 & \cellcolor{blue!12}1e-3 & \cellcolor{blue!12}1e-2 & \cellcolor{blue!12}1e-3 & \cellcolor{blue!12}10 & \cellcolor{blue!12}300 & \cellcolor{blue!12}50000-51023 & \cellcolor{blue!12}64 & \cellcolor{blue!12}iPNDM & \cellcolor{blue!12}22 \\
& & \cellcolor{green!12}5 & \cellcolor{green!12}poly/uni & \cellcolor{green!12}7/2 & \cellcolor{green!12}1e-3 & \cellcolor{green!12}1e-3 & \cellcolor{green!12}1e-2 & \cellcolor{green!12}1e-3 & \cellcolor{green!12}10 & \cellcolor{green!12}300 & \cellcolor{green!12}50000-51023 & \cellcolor{green!12}64 & \cellcolor{green!12}iPNDM & \cellcolor{green!12}21 \\
& & \cellcolor{yellow!12}6 & \cellcolor{yellow!12}poly/uni & \cellcolor{yellow!12}7/2 & \cellcolor{yellow!12}1e-3 & \cellcolor{yellow!12}1e-3 & \cellcolor{yellow!12}1e-2 & \cellcolor{yellow!12}1e-3 & \cellcolor{yellow!12}10 & \cellcolor{yellow!12}300 & \cellcolor{yellow!12}50000-51023 & \cellcolor{yellow!12}64 & \cellcolor{yellow!12}iPNDM & \cellcolor{yellow!12}21 \\
& & \cellcolor{orange!12}7 & \cellcolor{orange!12}poly/uni & \cellcolor{orange!12}7/2 & \cellcolor{orange!12}1e-3 & \cellcolor{orange!12}1e-3 & \cellcolor{orange!12}1e-2 & \cellcolor{orange!12}1e-3 & \cellcolor{orange!12}10 & \cellcolor{orange!12}300 & \cellcolor{orange!12}50000-51023 & \cellcolor{orange!12}64 & \cellcolor{orange!12}iPNDM & \cellcolor{orange!12}25 \\  \cline{3-15}
& \multirow{4}{*}{iPNDM} 
& \cellcolor{blue!12}4 & \cellcolor{blue!12}poly/uni & \cellcolor{blue!12}7/2 & \cellcolor{blue!12}1e-3 & \cellcolor{blue!12}1e-3 & \cellcolor{blue!12}1e-2 & \cellcolor{blue!12}1e-3 & \cellcolor{blue!12}10 & \cellcolor{blue!12}300 & \cellcolor{blue!12}50000-51023 & \cellcolor{blue!12}64 & \cellcolor{blue!12}iPNDM & \cellcolor{blue!12}22 \\
& & \cellcolor{green!12}5 & \cellcolor{green!12}poly/uni & \cellcolor{green!12}7/2 & \cellcolor{green!12}1e-3 & \cellcolor{green!12}1e-3 & \cellcolor{green!12}1e-2 & \cellcolor{green!12}1e-3 & \cellcolor{green!12}10 & \cellcolor{green!12}300 & \cellcolor{green!12}50000-51023 & \cellcolor{green!12}64 & \cellcolor{green!12}iPNDM & \cellcolor{green!12}21 \\
& & \cellcolor{yellow!12}6 & \cellcolor{yellow!12}poly/uni & \cellcolor{yellow!12}7/2 & \cellcolor{yellow!12}1e-3 & \cellcolor{yellow!12}1e-3 & \cellcolor{yellow!12}1e-2 & \cellcolor{yellow!12}1e-3 & \cellcolor{yellow!12}10 & \cellcolor{yellow!12}300 & \cellcolor{yellow!12}50000-51023 & \cellcolor{yellow!12}64 & \cellcolor{yellow!12}iPNDM & \cellcolor{yellow!12}21 \\
& & \cellcolor{orange!12}7 & \cellcolor{orange!12}poly/uni & \cellcolor{orange!12}7/2 & \cellcolor{orange!12}1e-3 & \cellcolor{orange!12}1e-3 & \cellcolor{orange!12}1e-2 & \cellcolor{orange!12}1e-3 & \cellcolor{orange!12}10 & \cellcolor{orange!12}300 & \cellcolor{orange!12}50000-51023 & \cellcolor{orange!12}64 & \cellcolor{orange!12}iPNDM & \cellcolor{orange!12}25 \\  \cline{3-15}
& \multirow{4}{*}{DPM++(3M)} 
& \cellcolor{blue!12}4 & \cellcolor{blue!12}logsnr & \cellcolor{blue!12}-- & \cellcolor{blue!12}1e-3 & \cellcolor{blue!12}1e-3 & \cellcolor{blue!12}1e-2 & \cellcolor{blue!12}1e-3 & \cellcolor{blue!12}10 & \cellcolor{blue!12}300 & \cellcolor{blue!12}50000-51023 & \cellcolor{blue!12}64 & \cellcolor{blue!12}DPM++(3M) & \cellcolor{blue!12}22 \\
& & \cellcolor{green!12}5 & \cellcolor{green!12}logsnr & \cellcolor{green!12}-- & \cellcolor{green!12}1e-3 & \cellcolor{green!12}1e-3 & \cellcolor{green!12}1e-2 & \cellcolor{green!12}1e-3 & \cellcolor{green!12}10 & \cellcolor{green!12}300 & \cellcolor{green!12}50000-51023 & \cellcolor{green!12}64 & \cellcolor{green!12}DPM++(3M) & \cellcolor{green!12}21 \\
& & \cellcolor{yellow!12}6 & \cellcolor{yellow!12}logsnr & \cellcolor{yellow!12}-- & \cellcolor{yellow!12}1e-3 & \cellcolor{yellow!12}1e-3 & \cellcolor{yellow!12}1e-2 & \cellcolor{yellow!12}1e-3 & \cellcolor{yellow!12}10 & \cellcolor{yellow!12}300 & \cellcolor{yellow!12}50000-51023 & \cellcolor{yellow!12}64 & \cellcolor{yellow!12}DPM++(3M) & \cellcolor{yellow!12}21 \\
& & \cellcolor{orange!12}7 & \cellcolor{orange!12}logsnr & \cellcolor{orange!12}-- & \cellcolor{orange!12}1e-3 & \cellcolor{orange!12}1e-3 & \cellcolor{orange!12}1e-2 & \cellcolor{orange!12}1e-3 & \cellcolor{orange!12}10 & \cellcolor{orange!12}300 & \cellcolor{orange!12}50000-51023 & \cellcolor{orange!12}64 & \cellcolor{orange!12}DPM++(3M) & \cellcolor{orange!12}25 \\
\midrule
\multirow{12}{*}{\textbf{LSUN Bedroom}} 
& \multirow{4}{*}{DDIM}
& \cellcolor{blue!12}4 & \cellcolor{blue!12}poly/uni & \cellcolor{blue!12}7/1 & \cellcolor{blue!12}5e-2 & \cellcolor{blue!12}5e-2 & \cellcolor{blue!12}5e-3 & \cellcolor{blue!12}5e-5 & \cellcolor{blue!12}10 & \cellcolor{blue!12}150 & \cellcolor{blue!12}50000-50255 & \cellcolor{blue!12}64 & \cellcolor{blue!12}iPNDM & \cellcolor{blue!12}22 \\
& & \cellcolor{green!12}5 & \cellcolor{green!12}poly/uni & \cellcolor{green!12}7/1 & \cellcolor{green!12}5e-2 & \cellcolor{green!12}5e-2 & \cellcolor{green!12}5e-3 & \cellcolor{green!12}5e-5 & \cellcolor{green!12}10 & \cellcolor{green!12}150 & \cellcolor{green!12}50000-50255 & \cellcolor{green!12}64 & \cellcolor{green!12}iPNDM & \cellcolor{green!12}21 \\
& & \cellcolor{yellow!12}6 & \cellcolor{yellow!12}poly/uni & \cellcolor{yellow!12}7/1 & \cellcolor{yellow!12}5e-2 & \cellcolor{yellow!12}5e-2 & \cellcolor{yellow!12}5e-3 & \cellcolor{yellow!12}5e-5 & \cellcolor{yellow!12}10 & \cellcolor{yellow!12}150 & \cellcolor{yellow!12}50000-50255 & \cellcolor{yellow!12}64 & \cellcolor{yellow!12}iPNDM & \cellcolor{yellow!12}21 \\
& & \cellcolor{orange!12}7 & \cellcolor{orange!12}poly/uni & \cellcolor{orange!12}7/1 & \cellcolor{orange!12}5e-2 & \cellcolor{orange!12}5e-2 & \cellcolor{orange!12}5e-3 & \cellcolor{orange!12}5e-5 & \cellcolor{orange!12}10 & \cellcolor{orange!12}150 & \cellcolor{orange!12}50000-50255 & \cellcolor{orange!12}64 & \cellcolor{orange!12}iPNDM & \cellcolor{orange!12}25 \\ \cline{3-15}
& \multirow{4}{*}{iPNDM} 
& \cellcolor{blue!12}4 & \cellcolor{blue!12}poly/uni & \cellcolor{blue!12}7/1 & \cellcolor{blue!12}5e-2 & \cellcolor{blue!12}5e-2 & \cellcolor{blue!12}5e-3 & \cellcolor{blue!12}5e-5 & \cellcolor{blue!12}10 & \cellcolor{blue!12}150 & \cellcolor{blue!12}50000-50255 & \cellcolor{blue!12}64 & \cellcolor{blue!12}iPNDM & \cellcolor{blue!12}22 \\
& & \cellcolor{green!12}5 & \cellcolor{green!12}poly/uni & \cellcolor{green!12}7/1 & \cellcolor{green!12}5e-2 & \cellcolor{green!12}5e-2 & \cellcolor{green!12}5e-3 & \cellcolor{green!12}5e-5 & \cellcolor{green!12}10 & \cellcolor{green!12}150 & \cellcolor{green!12}50000-50255 & \cellcolor{green!12}64 & \cellcolor{green!12}iPNDM & \cellcolor{green!12}21 \\
& & \cellcolor{yellow!12}6 & \cellcolor{yellow!12}poly/uni & \cellcolor{yellow!12}7/1 & \cellcolor{yellow!12}5e-2 & \cellcolor{yellow!12}5e-2 & \cellcolor{yellow!12}5e-3 & \cellcolor{yellow!12}5e-5 & \cellcolor{yellow!12}10 & \cellcolor{yellow!12}150 & \cellcolor{yellow!12}50000-50255 & \cellcolor{yellow!12}64 & \cellcolor{yellow!12}iPNDM & \cellcolor{yellow!12}21 \\
& & \cellcolor{orange!12}7 & \cellcolor{orange!12}poly/uni & \cellcolor{orange!12}7/1 & \cellcolor{orange!12}5e-2 & \cellcolor{orange!12}5e-2 & \cellcolor{orange!12}5e-3 & \cellcolor{orange!12}5e-5 & \cellcolor{orange!12}10 & \cellcolor{orange!12}150 & \cellcolor{orange!12}50000-50255 & \cellcolor{orange!12}64 & \cellcolor{orange!12}iPNDM & \cellcolor{orange!12}25 \\ \cline{3-15}
& \multirow{4}{*}{DPM++(3M)} 
& \cellcolor{blue!12}4 & \cellcolor{blue!12}logsnr & \cellcolor{blue!12}-- & \cellcolor{blue!12}5e-2 & \cellcolor{blue!12}5e-2 & \cellcolor{blue!12}5e-3 & \cellcolor{blue!12}5e-5 & \cellcolor{blue!12}10 & \cellcolor{blue!12}150 & \cellcolor{blue!12}50000-50255 & \cellcolor{blue!12}64 & \cellcolor{blue!12}DPM++(3M) & \cellcolor{blue!12}22 \\
& & \cellcolor{green!12}5 & \cellcolor{green!12}logsnr & \cellcolor{green!12}-- & \cellcolor{green!12}5e-2 & \cellcolor{green!12}5e-2 & \cellcolor{green!12}5e-3 & \cellcolor{green!12}5e-5 & \cellcolor{green!12}10 & \cellcolor{green!12}150 & \cellcolor{green!12}50000-50255 & \cellcolor{green!12}64 & \cellcolor{green!12}DPM++(3M) & \cellcolor{green!12}21 \\
& & \cellcolor{yellow!12}6 & \cellcolor{yellow!12}logsnr & \cellcolor{yellow!12}-- & \cellcolor{yellow!12}5e-2 & \cellcolor{yellow!12}5e-2 & \cellcolor{yellow!12}5e-3 & \cellcolor{yellow!12}5e-5 & \cellcolor{yellow!12}10 & \cellcolor{yellow!12}150 & \cellcolor{yellow!12}50000-50255 & \cellcolor{yellow!12}64 & \cellcolor{yellow!12}DPM++(3M) & \cellcolor{yellow!12}21 \\
& & \cellcolor{orange!12}7 & \cellcolor{orange!12}logsnr & \cellcolor{orange!12}-- & \cellcolor{orange!12}5e-2 & \cellcolor{orange!12}5e-2 & \cellcolor{orange!12}5e-3 & \cellcolor{orange!12}5e-5 & \cellcolor{orange!12}10 & \cellcolor{orange!12}150 & \cellcolor{orange!12}50000-50255 & \cellcolor{orange!12}64 & \cellcolor{orange!12}DPM++(3M) & \cellcolor{orange!12}25 \\
\midrule
\multirow{8}{*}{\textbf{MS COCO}} 
& \multirow{4}{*}{DDIM}
& \cellcolor{blue!12}4 & \cellcolor{blue!12}discrete & \cellcolor{blue!12}1 & \cellcolor{blue!12}3e-2 & \cellcolor{blue!12}2e-2 & \cellcolor{blue!12}1e-2 & \cellcolor{blue!12}1e-3 & \cellcolor{blue!12}10 & \cellcolor{blue!12}200 & \cellcolor{blue!12}0-255 & \cellcolor{blue!12}64 & \cellcolor{blue!12}DPM++(2M) & \cellcolor{blue!12}22 \\
& & \cellcolor{green!12}5 & \cellcolor{green!12}discrete & \cellcolor{green!12}1 & \cellcolor{green!12}3e-2 & \cellcolor{green!12}2e-2 & \cellcolor{green!12}1e-2 & \cellcolor{green!12}1e-3 & \cellcolor{green!12}10 & \cellcolor{green!12}200 & \cellcolor{green!12}0-255 & \cellcolor{green!12}64 & \cellcolor{green!12}DPM++(2M) & \cellcolor{green!12}21 \\
& & \cellcolor{yellow!12}6 & \cellcolor{yellow!12}discrete & \cellcolor{yellow!12}1 & \cellcolor{yellow!12}3e-2 & \cellcolor{yellow!12}2e-2 & \cellcolor{yellow!12}1e-2 & \cellcolor{yellow!12}1e-3 & \cellcolor{yellow!12}10 & \cellcolor{yellow!12}200 & \cellcolor{yellow!12}0-255 & \cellcolor{yellow!12}64 & \cellcolor{yellow!12}DPM++(2M) & \cellcolor{yellow!12}21 \\
& & \cellcolor{orange!12}7 & \cellcolor{orange!12}discrete & \cellcolor{orange!12}1 & \cellcolor{orange!12}3e-2 & \cellcolor{orange!12}2e-2 & \cellcolor{orange!12}1e-2 & \cellcolor{orange!12}1e-3 & \cellcolor{orange!12}10 & \cellcolor{orange!12}200 & \cellcolor{orange!12}0-255 & \cellcolor{orange!12}64 & \cellcolor{orange!12}DPM++(2M) & \cellcolor{orange!12}25 \\ \cline{3-15}
& \multirow{4}{*}{DPM++(2M)} 
& \cellcolor{blue!12}4 & \cellcolor{blue!12}discrete & \cellcolor{blue!12}1 & \cellcolor{blue!12}3e-2 & \cellcolor{blue!12}2e-2 & \cellcolor{blue!12}1e-3 & \cellcolor{blue!12}1e-4 & \cellcolor{blue!12}10 & \cellcolor{blue!12}200 & \cellcolor{blue!12}0-255 & \cellcolor{blue!12}64 & \cellcolor{blue!12}DPM++(2M) & \cellcolor{blue!12}22 \\
& & \cellcolor{green!12}5 & \cellcolor{green!12}discrete & \cellcolor{green!12}1 & \cellcolor{green!12}3e-2 & \cellcolor{green!12}2e-2 & \cellcolor{green!12}1e-3 & \cellcolor{green!12}1e-4 & \cellcolor{green!12}10 & \cellcolor{green!12}200 & \cellcolor{green!12}0-255 & \cellcolor{green!12}64 & \cellcolor{green!12}DPM++(2M) & \cellcolor{green!12}21 \\
& & \cellcolor{yellow!12}6 & \cellcolor{yellow!12}discrete & \cellcolor{yellow!12}1 & \cellcolor{yellow!12}3e-2 & \cellcolor{yellow!12}2e-2 & \cellcolor{yellow!12}1e-3 & \cellcolor{yellow!12}1e-4 & \cellcolor{yellow!12}10 & \cellcolor{yellow!12}200 & \cellcolor{yellow!12}0-255 & \cellcolor{yellow!12}64 & \cellcolor{yellow!12}DPM++(2M) & \cellcolor{yellow!12}21 \\
& & \cellcolor{orange!12}7 & \cellcolor{orange!12}discrete & \cellcolor{orange!12}1 & \cellcolor{orange!12}3e-2 & \cellcolor{orange!12}2e-2 & \cellcolor{orange!12}1e-3 & \cellcolor{orange!12}1e-4 & \cellcolor{orange!12}10 & \cellcolor{orange!12}200 & \cellcolor{orange!12}0-255 & \cellcolor{orange!12}64 & \cellcolor{orange!12}DPM++(2M) & \cellcolor{orange!12}25 \\
\bottomrule
\end{tabular}}%
\label{tab:exp-set}
\end{table*}

\begin{table}[tb]
  \centering
  \begin{minipage}[t]{0.61\textwidth}
    \centering
    \caption{Datasets, Checkpoints, and Licenses}
    \centering

    \begin{tabular}{ll}
    \toprule
    \multicolumn{2}{c}{{\color[HTML]{000000} Datasets}} \\
    \midrule
    Name & License \\
    \midrule
    CIFAR-10~\cite{cifar10} & \textbackslash{} \\
    FFHQ 64X64~\cite{ffhq} & CC BY-NC-SA 4.0 \\
    ImageNet 64~\cite{imagenet} & Custom \\
    ImageNet 256~\cite{imagenet} & Custom \\
    LSUN Bedroom~\cite{lsun} & \textbackslash{} \\
    MS-COCO~\cite{coco} & Custom \\
    \midrule
    \multicolumn{2}{c}{Checkpoints} \\
    \midrule
    Name & License \\
    \midrule
    edm-cifar10-32x32-uncond-vp.pkl & \\
    edm-ffhq-64x64-uncond-vp.pkl & \\
    edm-imagenet-64x64-cond-adm.pkl & \multirow{-3}{*}{CC BY-NC-SA 4.0} \\
    DiT-XL-2-256X256.pt & CC BY-NC 4.0 \\
    edm\_bedroom256\_ema.pt & \textbackslash{} \\
    v1-5-pruned-emaonly.ckpt & \textbackslash{} \\
    \bottomrule
    \end{tabular}
    \label{tab:checkpoints}
        
  \end{minipage}
  \hfill
  \begin{minipage}[t]{0.33\textwidth}
    \centering
        \caption{Status Files}
            \begin{tabular}{l}
            \toprule
            \multicolumn{1}{c}{Status Files} \\
            \midrule
            Name \\
            \midrule
            cifar10-32x32.npz \\
            ffhq-64x64.npz \\
            imagenet-64x64.npz \\
            lsun\_bedroom-256x256.npz \\
            ms\_coco-512x512.npz \\
            VIRTUAL\_imagenet64\\
            \_labeled.npz \\
            VIRTUAL\_imagenet256\\
            \_labeled.npz \\
            \bottomrule
            \end{tabular}
            \label{tab:status}
  \end{minipage}
\end{table}




\section{Time Estimates Compared to Distillation}

\subsection{Data Sources}
Except for DMD~\cite{dmd} and ECM~\cite{ect}, all training-time data are taken from~\cite{AMED}, which reports times on an A100 GPU. (DMD does not specify its training-time overhead.) According to ECM’s Table 8, training took 24 hours on one A6000 for CIFAR-10, and 8.5 hours on four H100s for ImageNet-64.

\subsection{GPU Time Conversions}
\subsubsection{A100 to RTX 3090.}
We profiled MTEO on CIFAR-10: on an RTX 3090 it required 0.59, 0.52, 0.54, 0.53 hours for 3, 4, 5, and 6 NFE, respectively; on an A100 these were 0.069, 0.097, 0.105, 0.118 hours.  The ratios (A100/RTX) are roughly 0.59, 0.52, 0.54, 0.53, averaging 0.54.  Thus we approximate \(T_{\mathrm{RTX3090}}\approx T_{\mathrm{A100}}/0.54\).

\subsubsection{H100/A6000 to RTX 3090.}
We assume H100 and A6000 are about 2.25× and 1.3× faster than the RTX 3090, respectively, and scale the reported times accordingly.




\section{Ablation Study}

\subsection{Batch Size}
We evaluated the impact of training batch size on MTEO. The default is 64; we also tried 16, 32, and 128. Experiments on three ODE samplers (see \cref{tab:cifar-batch}) show that batch size has only a minor effect on final performance.

\subsection{Training Set Size}
The default MTEO uses 256 trajectories (seeds 50000–50255). We also trained with 64 trajectories (seeds 50000–50063) and 128 trajectories (50000–50127). As shown in \cref{tab:cifar-seeds}, larger training sets generally yield slightly better results, indicating that MTEO benefits modestly from more diverse reference trajectories.

\begin{table}[tb]
  \centering
  \begin{minipage}[t]{0.49\textwidth}
    \centering
    \caption{Ablation on CIFAR10 based on FID. train seeds denote training datasets generated by corresponding random seeds.}
        \centering
        \small
        \setlength{\tabcolsep}{0.5pt}
        \newcommand{\mteo}[1]{\cellcolor{gray!15}#1}
        \newcommand{\single}[1]{#1}
        
        \begin{tabular}{l c rrr}
        \toprule
        \multirow{2}{*}{Method} & \multirow{2}{*}{NFE} &
        \multicolumn{3}{c}{Train Seeds}  \\
        \cmidrule(lr){3-5}
         & & 50000-50063 & 50127 & 50255 \\
        \midrule
        \multirow{4}{*}{DDIM}
         & 3 & 5.99 & \single{\textbf{4.92}}  & \mteo{5.29}\\
         & 4 & 4.53 & \single{\textbf{3.77}}  & \mteo{3.91}\\
         & 5 & 3.84 & \single{3.17}   & \mteo{\textbf{3.02}}\\
         & 6 & 3.36 & \single{2.93}   & \mteo{\textbf{2.87}}\\
        \midrule
        \multirow{4}{*}{iPNDM}
         & 3 & 7.54 & \single{6.66} & \mteo{\textbf{6.63}}\\
         & 4 & 5.38 & \single{4.92} & \mteo{\textbf{4.81}}\\
         & 5 & 4.86 & \single{4.30}  & \mteo{\textbf{4.20}}\\
         & 6 & 3.67 & \single{3.27}   & \mteo{\textbf{3.26}}\\
        \midrule
        \multirow{4}{*}{DPM++(3M)}
         & 3 & 7.22 & \single{\textbf{5.35}} & \mteo{5.54} \\
         & 4 & 4.33 & \single{4.04} & \mteo{\textbf{3.87}} \\
         & 5 & 4.13 & \single{3.73}  & \mteo{\textbf{3.38}} \\
         & 6 & 3.46 & \single{\textbf{3.34}}  & \mteo{3.60} \\
        \bottomrule
        \end{tabular}
        \label{tab:cifar-seeds}

  \end{minipage}
  \hfill
  \begin{minipage}[t]{0.48\textwidth}
    \centering
        \caption{Ablation on CIFAR10 based on FID. Batch size denote  the batch size per update during the training process.}
        \centering
        \small
        \setlength{\tabcolsep}{0.5pt}
        \newcommand{\mteo}[1]{\cellcolor{gray!15}#1}
        \newcommand{\single}[1]{#1}
        
        \begin{tabular}{l c rrrr}
        \toprule
        \multirow{2}{*}{Method} & \multirow{2}{*}{NFE} &
        \multicolumn{4}{c}{Batch Size}  \\
        \cmidrule(lr){3-6}
         & & 16 & 32 & 64 & 128\\
        \midrule
        \multirow{4}{*}{DDIM}
         & 3 & \textbf{4.85} & \single{5.02}  & \mteo{5.29} & 5.60\\
         & 4 & 3.96 & \textbf{3.76}  & \mteo{3.91} & 4.10\\
         & 5 & 3.13 & \single{3.06}   & \mteo{\textbf{3.02}} & 3.18\\
         & 6 & 3.00 & \textbf{2.75}   & \mteo{2.87} & 2.97\\
        \midrule
        \multirow{4}{*}{iPNDM}
         & 3 & \textbf{6.43} & \single{6.62} & \mteo{6.63} & 6.92\\
         & 4 & 5.02 & \textbf{4.70} & \mteo{4.81} & 4.97\\
         & 5 & 4.43 & \textbf{4.18} & \mteo{4.20} & 4.30\\
         & 6 & 3.48 & \textbf{3.25} & \mteo{3.26} & 3.26\\
        \midrule
        \multirow{4}{*}{DPM++(3M)}
         & 3 & 5.12 & \textbf{5.10} & \mteo{5.54} & 6.29\\
         & 4 & 4.80 & \single{4.62} & \mteo{\textbf{3.87}} & 4.73\\
         & 5 & 4.23 & \single{3.79}  & \mteo{\textbf{3.38}} & 4.39\\
         & 6 & 3.84 & \single{3.82}  & \mteo{3.60} & \textbf{3.59}\\
        \bottomrule
        \end{tabular}
        \label{tab:cifar-batch}
  \end{minipage}
\end{table}
\subsection{Training Heuristics}
\label{app:train-heuristics}

We further evaluate the proposed training heuristics by comparing them against a simple \emph{fixed-epoch} baseline. Specifically, using DDIM as the underlying sampler, we train each step’s embeddings for a fixed number of epochs ($E_{max} \in \{100, 300\}$) and contrast the resulting quality--cost trade-off with our adaptive scheme (\cref{alg:train-full}), which allocates training effort per step based on convergence.

As shown in~\cref{tab:cifar_fixed_steps}, the fixed-epoch strategy exhibits a clear inefficiency: using a small budget (100 epochs) tends to under-train difficult steps and leaves performance on the table, while a large budget (300 epochs) improves quality but incurs unnecessary computation on steps that converge quickly. In contrast, our heuristics achieve a more favorable Pareto trade-off by (i) terminating updates once the relative loss improvement becomes negligible, while (ii) ensuring that later denoising steps are not prematurely stopped. Overall, the proposed scheme preserves the sample quality of the higher-budget fixed-epoch setting while substantially reducing the total training time, validating its role in balancing training overhead and final performance.

\subsection{Sensitivity to Teacher Steps}
We varied the length of the teacher trajectory. \cref{tab:tea-steps-nfe} shows that increasing or decreasing the teacher sampling steps by a small amount changes the FID only slightly. Even halving the teacher steps in the 3-NFE setting worsened FID by only 0.31. This suggests our main training configuration is well balanced.
\begin{algorithm}[!h]
    \renewcommand{\algorithmicrequire}{\textbf{Input:}}
    \renewcommand{\algorithmicensure}{\textbf{Output:}}
    \caption{Training MTE via Trajectory Distillation (specific)}
    \label{alg:train-full}
    \begin{algorithmic}[1]
        \REQUIRE Teacher trajectory $\{\hat{x}_{t_i}\}_{i=0}^{N-1}$, time schedule $\{t_i\}_{i=0}^{N-1}$, ODE solver $S$,
        frozen denoiser $\epsilon_\theta$, distance metric $\mathcal{D}$,
        layer-wise embeddings $\{\Phi_i\}_{i=0}^{N-1}$ with $\Phi_i=\{\phi_{i,\ell}\}_{\ell=1}^{L}$, Threshold $\epsilon$, Patience $P$, maximum epoch $E_{max}$.
        \ENSURE Optimized embeddings $\{\Phi_i\}_{i=0}^{N-1}$.
        \STATE $x_{t_0} \leftarrow \hat{x}_{t_0}$
        \FOR{$i = 0$ \textbf{to} $N-2$}
            \STATE $c=0,p=0$
            \WHILE{$p<P \ and \ c < E_{max} $}
                \STATE $x_{t_{i+1}} \leftarrow S(x_{t_i},\, \epsilon_\theta,\, t_i,\, t_{i+1},\, \Phi_i)$
                \STATE $\mathcal{L} \leftarrow \mathcal{D}\!\big(x_{t_{i+1}},\, \hat{x}_{t_{i+1}}\big)$
                \STATE Update $\Phi_i$ using $\nabla_{\Phi_i}\mathcal{L}$
                \IF{$c == 0$}
                    \STATE $\mathcal{L}_{pre} \leftarrow \mathcal L$ 
                \ELSIF{$(\mathcal{L}_{pre} -\mathcal L)/\mathcal{L}_{pre}<\epsilon$ }
                    \STATE  $p=p+1$
                \ELSE
                    \STATE $p=0$
                \ENDIF
                \STATE $c=c+1$
            \ENDWHILE
            \STATE $x_{t_{i+1}} \leftarrow S(x_{t_i},\, \epsilon_\theta,\, t_i,\, t_{i+1},\, \Phi_i)$

        \ENDFOR
        \RETURN $\{\Phi_i\}_{i=0}^{N-1}$
    \end{algorithmic}
\end{algorithm}

\begin{table}[!htbp]
  \centering
  \renewcommand{\arraystretch}{0.86}
  \begin{minipage}[t]{0.55\textwidth}
    \centering
    \captionsetup{width=17em}
    \caption{Ablation of training heuristics on CIFAR-10. We use +DDIM as baseline and compare with fixed training epochs 100 and 300. To demonstrate the efficiency of our mechanism, we also report training time based on RTX3090 GPU hours.}
    \label{tab:cifar_fixed_steps}
  \setlength{\tabcolsep}{0.9em}
  \begin{tabular}{cccc}
    \toprule
    Epochs & NFE & FID & Hours \\
    \midrule
    100 & \multirow{3}{*}{3} & 5.37 & 0.10 \\
    300 & & 4.96 & 0.30 \\
    Ours & & 5.29 & 0.13 \\
    \midrule
    100 & \multirow{3}{*}{4} & 4.14 & 0.13 \\
    300 & & 3.84 & 0.40 \\
    Ours & & 3.91 & 0.18 \\
    \midrule
    100 & \multirow{3}{*}{5} & 3.26 & 0.17 \\
    300 & & 3.18 & 0.50 \\
    Ours & & 3.02 & 0.19 \\
    \midrule
    100 & \multirow{3}{*}{6} & 3.03 & 0.20 \\
    300 & & 2.99 & 0.61 \\
    Ours & & 2.87 & 0.22 \\
    \bottomrule
  \end{tabular}

  \end{minipage}
  \hfill
  \renewcommand{\arraystretch}{0.89}
  \begin{minipage}[t]{0.4\textwidth}
    \centering
        \setlength{\tabcolsep}{0.8em}
          \caption{Comparation across different Teacher Steps on CIFAR-10. We use +DDIM as baseline.}
          \label{tab:tea-steps-nfe}
          \begin{tabular}{ccc}
            \toprule
            Teacher Steps & NFE & FID \\
            \midrule
            28 & \multirow{4}{*}{3} & 5.19 \\
            22 (Ours) & & 5.29 \\
            16 & & 5.27 \\
            13 & & 5.60 \\
            \midrule
            29 & \multirow{4}{*}{4} & 3.85 \\
            21 (Ours) & & 3.91 \\
            17 & & 3.96 \\
            13 & & 4.26 \\
            \midrule
            31 & \multirow{4}{*}{5} & 3.02 \\
            21 (Ours) & & 3.02 \\
            16 & & 3.26 \\
            11 & & 4.30 \\
            \midrule
            31 & \multirow{4}{*}{6} & 2.86 \\
            25 (Ours) & & 2.87 \\
            19 & & 2.98 \\
            13 & & 3.53 \\
            \bottomrule
          \end{tabular}

  \end{minipage}
\end{table}

\section{Application to Stable Diffusion and DiT}

\subsection{Stable Diffusion}
Stable Diffusion exhibits two main issues: (1) In the medium-step regime, increasing NFE often yields diminishing or negative returns, indicating a sampling threshold. (2) In the very-low-step regime, the model can converge to suboptimal modes. For example, with 4 steps it converges early, and even adding many mid-steps between 3 and 4 NFEs does not escape this mode; only after \(\sim\)15 NFEs does it reach the optimal mode. We attribute this to the complexity of the conditional generation task, which may limit MTEO’s corrective range. To mitigate this, we propose two extensions: (i) jointly optimizing the “unconditional\_condition” embedding for each step alongside MTE, providing stronger guidance; (ii) combining MTEO with a timestep search strategy to steer the trajectory toward the optimal mode. In this work we implement (i) and defer (ii) to future work.

\subsection{DiT (Diffusion Transformers)}
The DiT codebase supports only low-order samplers (DDPM/DDIM), making medium-step results poor. We trained MTEO using a teacher trajectory of 100→3, 101→4, 101→5, 103→6 and corresponding student NFEs of 4, 5, 6, 7. The overall time-conditioning pipeline remains the same (with AdaLayerNorm replacing FiLM). We trained each step’s embeddings for 300 epochs (LR $2\times10^{-5}$, batch 64) with no early stopping or extra heuristics, also including the “unconditional\_condition” embedding. Under DiT, MTEO achieves exceptionally strong results, which we attribute to the Transformer’s inherent layer-wise design synergizing with our multi-layer embeddings.

\section{Discussion}
\label{sec:discussion}


\subsection{On the Effective Dimensionality of Timestep Embeddings}
As discussed in~\cref{subsec:vartoemb} and visualized in~\cref{fig:pca_emb}, the original timestep embeddings occupy a highly low-dimensional subset of the embedding space.
Concretely, PCA on the embeddings of 121 timesteps embeddings shows that the first two principal components explain \textbf{77.72\%} and \textbf{19.41\%} of the variance, accounting for \textbf{97.14\%} in total.
In contrast, the FiLM modulation parameters that directly affect layer-wise feature processing, i.e., $(\alpha_\ell,\beta_\ell)$, exhibit substantially higher effective dimensionality: even the first 30 principal components explain only \textbf{90.49\%} of the variance.

A natural interpretation is that the conventional timestep embedding is generated from a \emph{scalar} variable $t$.
Although sinusoidal encoding and an MLP map $t$ to a high-dimensional vector, the resulting embeddings across time lie on a low-dimensional manifold induced by a one-dimensional input; PCA then recovers a very low-dimensional linear approximation of this trajectory.
This creates a potential representational bottleneck when the downstream FiLM parameters vary in a richer, higher-dimensional manner across layers and time.

We further apply the same PCA analysis to the learned MTE embeddings.
As shown in~\cref{fig:pca_emb}, MTE substantially increases the effective dimensionality of the embedding space: the first 30 principal components explain only \textbf{84.53\%} of the variance.
This indicates that the optimized embeddings are no longer constrained to a narrow low-dimensional subset, which may partially explain the strong empirical performance of MTEO in few-step sampling.

Interestingly, we also observe a small but consistent change in the geometry of the optimized FiLM parameters:
after optimization, the cumulative explained variance of the first 30 principal components increases from \textbf{90.49\%} to \textbf{93.49\%}.
This suggests that MTEO may make the effective modulation patterns slightly more structured (i.e., variance concentrated in fewer components), potentially reflecting a more coherent use of FiLM degrees of freedom.

\subsection{Implications for Interpreting Diffusion Networks}
Neural networks demonstrate remarkable capabilities, yet their internal mechanisms often remain opaque.
An intriguing empirical finding from our experiments is that MTEO can reach sampling quality comparable to distillation-based acceleration methods while training fewer than \textbf{0.2\%} additional parameters, whereas many distillation approaches require updating a large fraction (often all) of the model parameters.

From a mechanistic perspective, MTEO primarily acts by adjusting feature-wise affine modulation (FiLM) within the network: it changes how intermediate features are scaled and shifted at different layers and time steps.
Since different layers are known to encode different types of information and operate at different spatial/semantic granularities, even simple gain-control-like adjustments can have surprisingly large downstream effects on the denoising trajectory.
This observation suggests a promising direction for future work: analyzing the learned modulation patterns across layers and time (\eg., which blocks and which denoising stages are most sensitive) may provide actionable clues toward understanding diffusion models beyond a purely black-box view.

\subsection{A Direct FiLM-Parameter Variant: MTEO-deep}
As discussed throughout the paper, MTEO ultimately influences sampling by changing the effective FiLM parameters $(\alpha_\ell,\beta_\ell)$.
This motivates a natural extension: directly optimizing $(\alpha_\ell,\beta_\ell)$ instead of learning the embeddings that generate them.
We call this variant \textbf{MTEO-deep} and include its experimental results in~\cref{tab:cifar-deep}.

We do not adopt MTEO-deep as the main method for two practical reasons.
First, the parameterization and injection locations of FiLM-style modulation vary across diffusion backbones and implementations (\eg, some blocks use bias-only conditioning, while others use scale-and-shift), which makes direct parameter optimization less portable and less unified across frameworks.
Second, directly parameterizing $(\alpha_\ell,\beta_\ell)$ can introduce a larger, architecture-dependent parameter footprint, whereas MTE leverages a compact embedding interface that is largely consistent across diffusion models (sinusoidal encoding followed by a small MLP).
For simplicity and broad applicability, we therefore use MTEO as our primary approach and treat MTEO-deep as an optional extension.

\begin{table}[!htbp]
\captionsetup{width=.5\textwidth}
\caption{Comparison on CIFAR10 between MTEO and MTEO-deep. We train both methods with same configuration}
\centering
\renewcommand{\arraystretch}{1.1}
\newcommand{\mteo}[1]{\cellcolor{gray!15}\textbf{#1}}
\newcommand{\single}[1]{\cellcolor{gray!5}#1}

\begin{tabular}{l c rr}
\toprule
\multirow{2}{*}{Method} & \multirow{2}{*}{NFE} &
\multicolumn{2}{c}{FID $\downarrow$}  \\
\cmidrule(lr){3-4}
 & & +MTEO  & +MTEO-deep \\
\midrule
\multirow{4}{*}{DDIM}
 & 3 & 5.29   & \mteo{5.18}\\
 & 4 & 3.91   & \mteo{3.93}\\
 & 5 & 3.02   & \mteo{3.48}\\
 & 6 & 2.87   & \mteo{3.05}\\
\midrule
\multirow{4}{*}{iPNDM}
 & 3 & 6.63  & \mteo{6.57}\\
 & 4 & 4.81  & \mteo{5.44}\\
 & 5 & 4.20   & \mteo{4.81}\\
 & 6 & 3.26   & \mteo{3.90}\\
\midrule
\multirow{4}{*}{DPM++(3M)}
 & 3 & 5.54  & \mteo{11.52} \\
 & 4 & 3.87  & \mteo{7.87} \\
 & 5 & 3.38   & \mteo{3.60} \\
 & 6 & 3.60   & \mteo{2.81} \\
\bottomrule
\end{tabular}
\label{tab:cifar-deep}
\end{table}



\section{Additional Trajectory Analyses}

\begin{figure*}[htbp]
    \centering

    \begin{subfigure}[b]{0.48\textwidth}
        \centering
        \includegraphics[width=\textwidth]{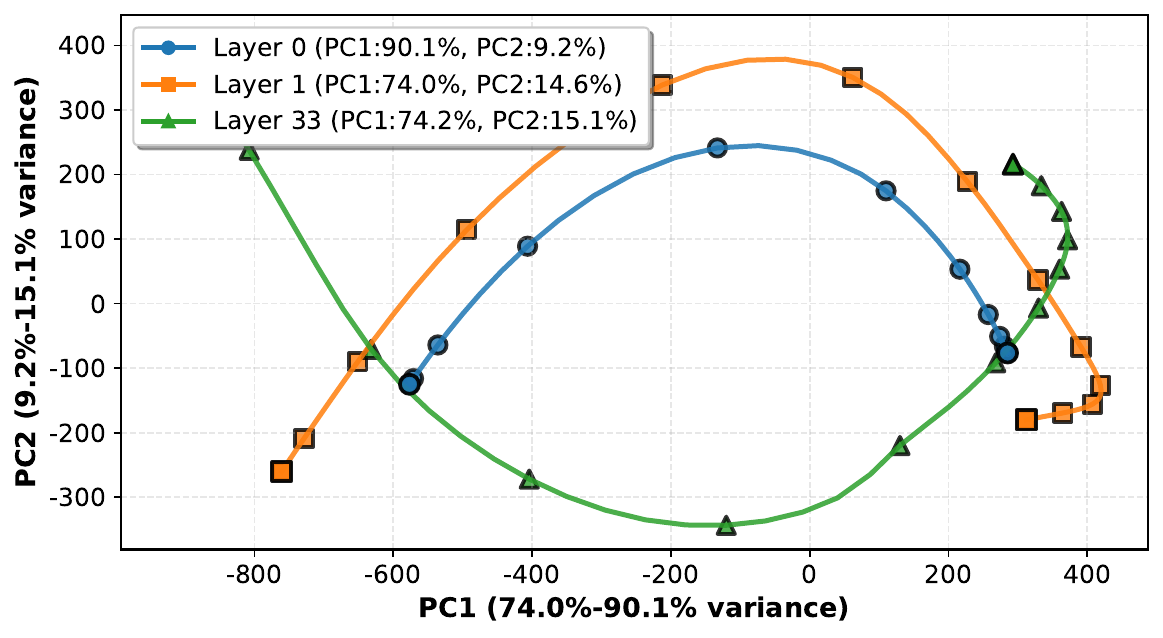}
        \caption{Multi-Layer Trajectory PCA: Sample 0 (2D)}
    \end{subfigure}
    \hfill
    \begin{subfigure}[b]{0.48\textwidth}
        \centering
        \includegraphics[width=\textwidth]{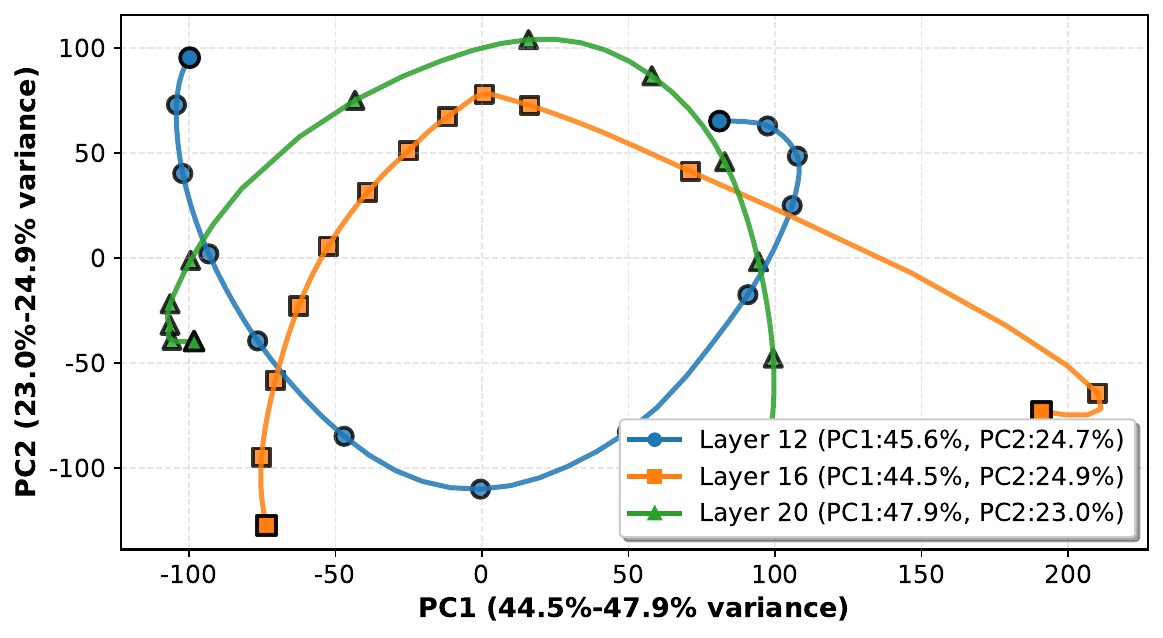}
        \caption{Multi-Layer Trajectory PCA: Sample 0 (2D)}
    \end{subfigure}

    \vskip\baselineskip
    \begin{subfigure}[b]{0.49\textwidth}
        \centering
        \includegraphics[width=\textwidth]{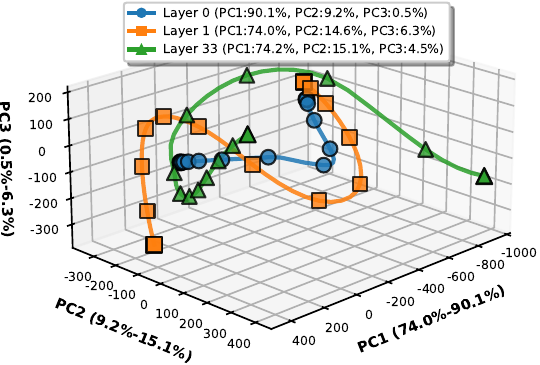}
        \caption{Multi-Layer Trajectory PCA: Sample 0 (3D)}
    \end{subfigure}
    \hfill
    \begin{subfigure}[b]{0.49\textwidth}
        \centering
        \includegraphics[width=\textwidth]{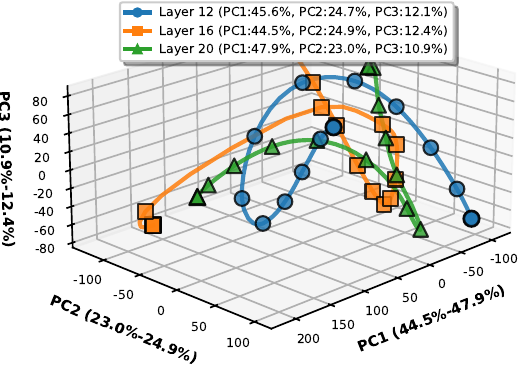}
        \caption{Multi-Layer Trajectory PCA: Sample 0 (3D)}
    \end{subfigure}

    \caption{Visualization of feature trajectory. See more example in appendix.}
    \label{fig:featuretraj}
\end{figure*}

\begin{figure*}[htbp]

    \centering
    
    \begin{subfigure}[b]{0.32\textwidth}
        \centering
        \includegraphics[width=\textwidth]{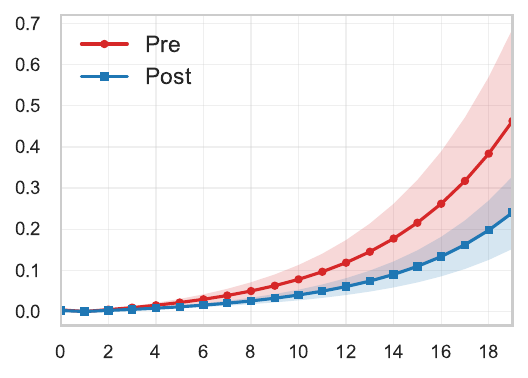}
        \caption{Idx 0 to 19. Seeds0-63}
    \end{subfigure}
    \hfill
    \begin{subfigure}[b]{0.32\textwidth}
        \centering
        \includegraphics[width=\textwidth]{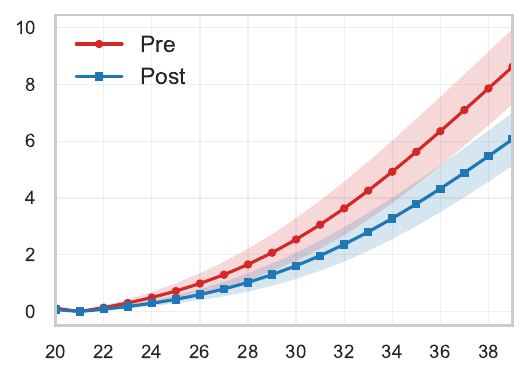}
        \caption{Idx 20 to 39. Seeds0-63}
    \end{subfigure}
    \hfill
    \begin{subfigure}[b]{0.32\textwidth}
        \centering
        \includegraphics[width=\textwidth]{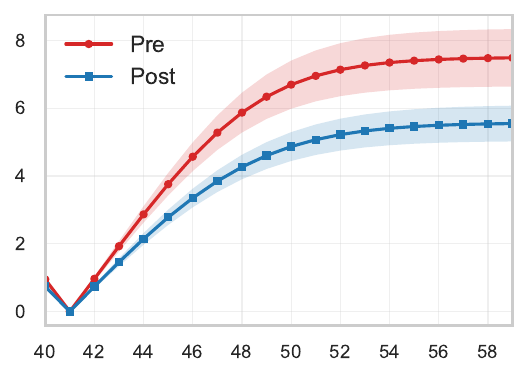}
        \caption{Idx 40 to 59. Seeds0-63}
    \end{subfigure}
    
    \vskip\baselineskip 
    \begin{subfigure}[b]{0.32\textwidth}
        \centering
        \includegraphics[width=\textwidth]{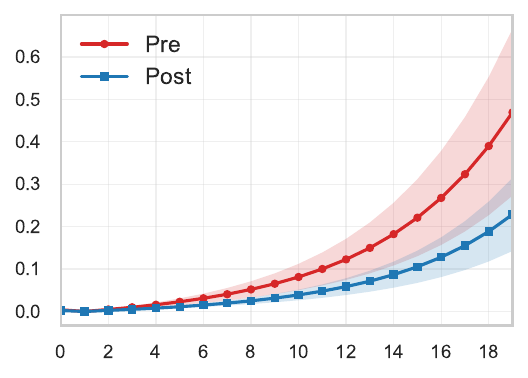}
        \caption{Idx 0 to 19. Seeds64-127}
    \end{subfigure}
    \hfill
    \begin{subfigure}[b]{0.32\textwidth}
        \centering
        \includegraphics[width=\textwidth]{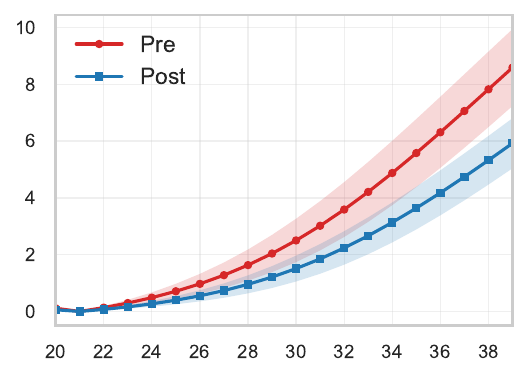}
        \caption{Idx 20 to 39. Seeds64-127}
    \end{subfigure}
    \hfill
    \begin{subfigure}[b]{0.32\textwidth}
        \centering
        \includegraphics[width=\textwidth]{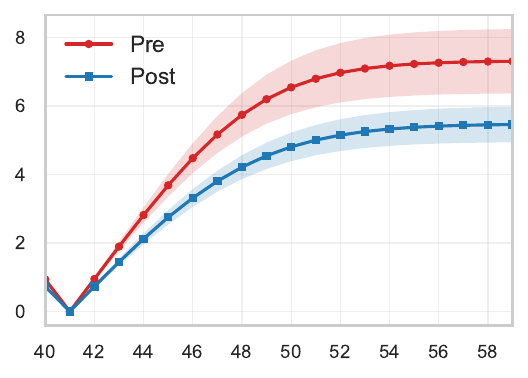}
        \caption{Idx 40 to 59. Seeds64-127}
    \end{subfigure}

    \caption{\textbf{Additional Experiments on FiLM’s Channel Correction Capability.} We conducted experiments on two batches of data, 0-63 and 64-127, and reported the mean values before and after correction, along with the methods used. First, we generated the true trajectory using iPNDM with 60 NFE. Based on this, we simulated a 4-steps sampling process, corresponding to the steps from Idx0 $\to$ Idx19 $\to$ Idx39 $\to$ Idx59.}
    \label{fig:film-extra}
\end{figure*}

\begin{figure}[!htbp]
    \centering

    \begin{subfigure}[b]{0.49\textwidth}
        \centering
        \includegraphics[width=\textwidth]{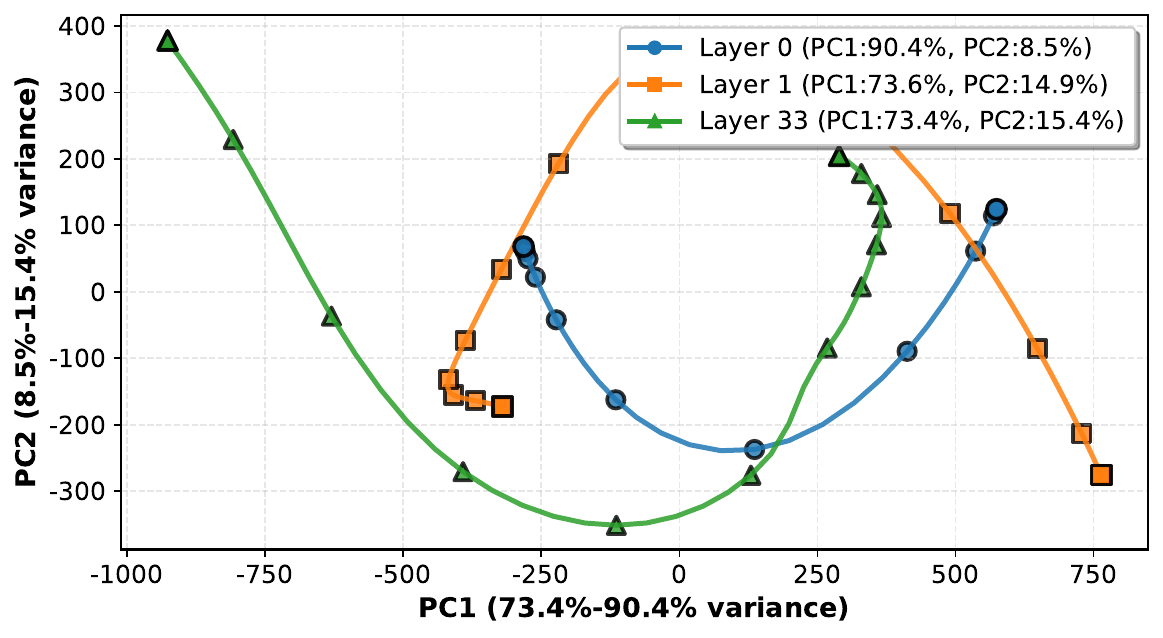}
        \caption{Multi-Layer Trajectory PCA: Sample 1 (2D)}
    \end{subfigure}
    \hfill
    \begin{subfigure}[b]{0.49\textwidth}
        \centering
        \includegraphics[width=\textwidth]{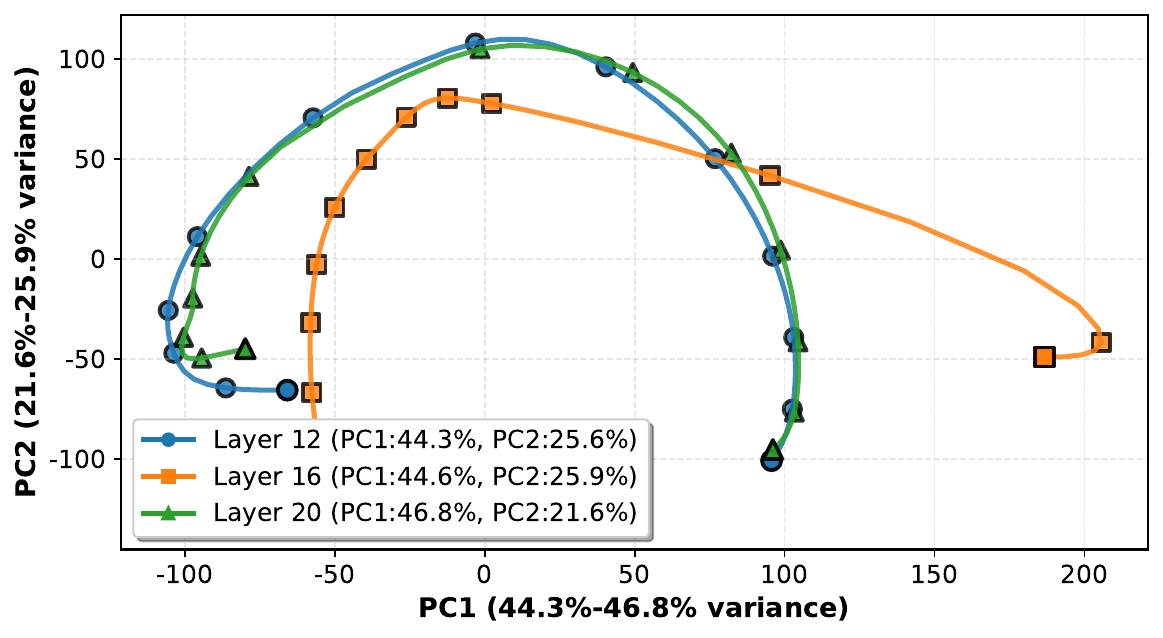}
        \caption{Multi-Layer Trajectory PCA: Sample 1 (2D)}
    \end{subfigure}

    \vskip\baselineskip
    \begin{subfigure}[b]{0.49\textwidth}
        \centering
        \includegraphics[width=\textwidth]{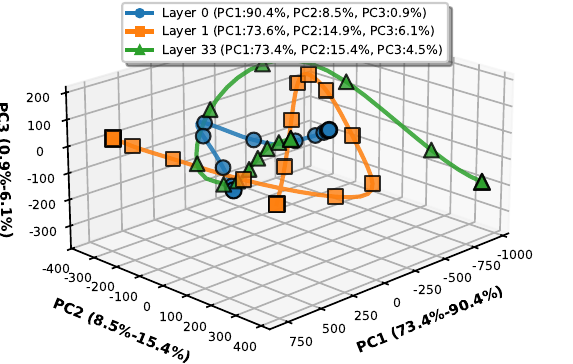}
        \caption{Multi-Layer Trajectory PCA: Sample 1 (3D)}
    \end{subfigure}
    \hfill
    \begin{subfigure}[b]{0.49\textwidth}
        \centering
        \includegraphics[width=\textwidth]{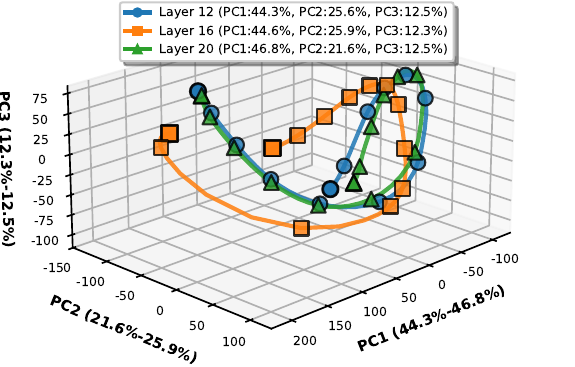}
        \caption{Multi-Layer Trajectory PCA: Sample 1 (3D)}
    \end{subfigure}

    \caption{Additional visualization of feature trajectory with trajectory 1}
    \label{fig:featuretraj-ex}
\end{figure}

\begin{figure}[!htbp]

    \centering
    
    \begin{subfigure}[b]{0.45\textwidth}
        \centering
        \includegraphics[width=\textwidth]{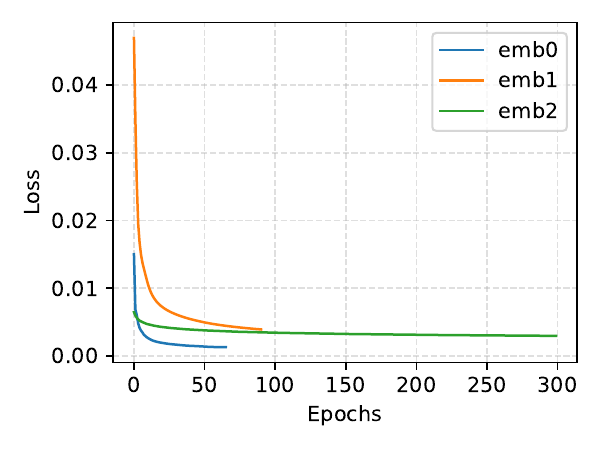}
    \end{subfigure}
    \begin{subfigure}[b]{0.45\textwidth}
        \centering
        \includegraphics[width=\textwidth]{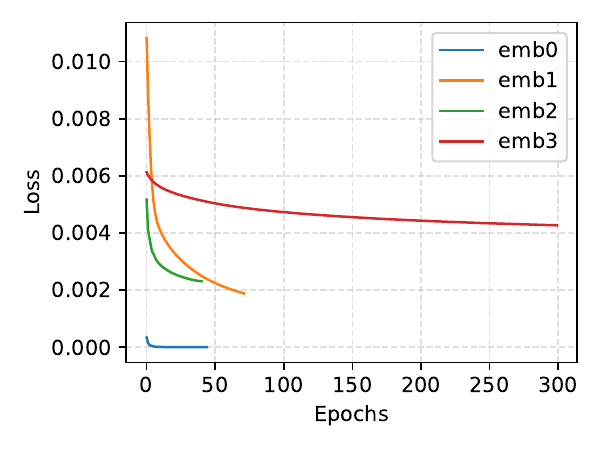}
    \end{subfigure}
    
    \vskip\baselineskip 
    \begin{subfigure}[b]{0.45\textwidth}
        \centering
        \includegraphics[width=\textwidth]{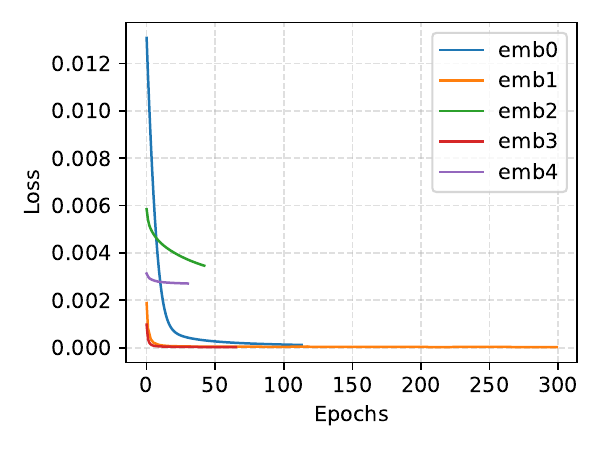}
    \end{subfigure}
    \begin{subfigure}[b]{0.45\textwidth}
        \centering
        \includegraphics[width=\textwidth]{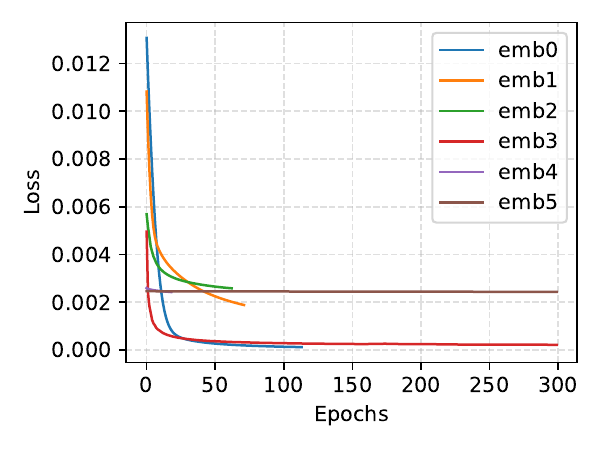}
    \end{subfigure}
    \captionsetup{width=1.\textwidth}
    \caption{\textbf{Trainging loss curve.} We display the trainging procedure of MTEO+DDIM on CIFAR10 with step 4, 5, 6, 7.}
    \label{fig:loss}
\end{figure}

\begin{figure*}[!htbp]
    \centering
    \includegraphics[width=1.\linewidth]{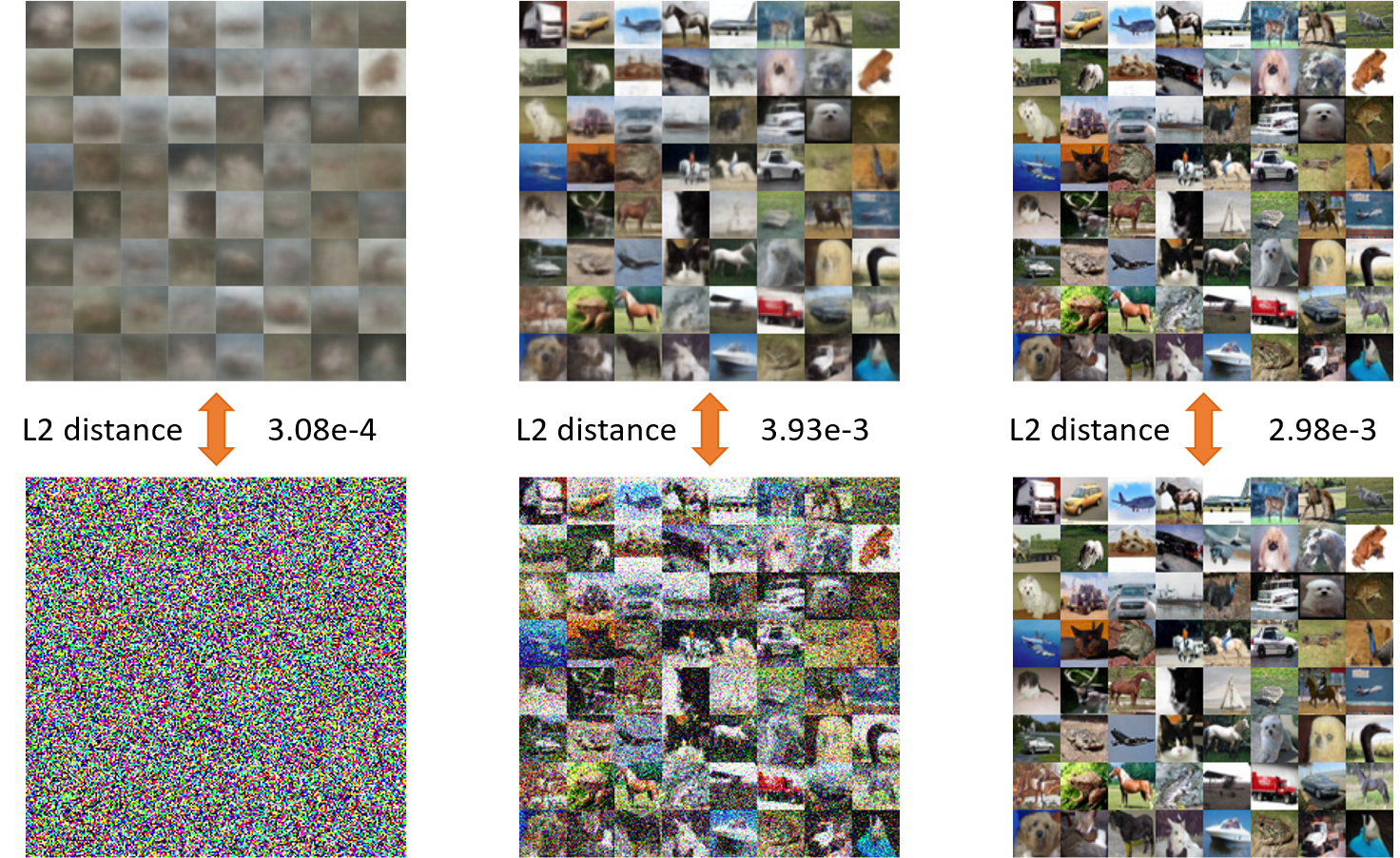}
    \caption{The intermediate visualization results of training procedures with 4-step MTEO+DDIM.}
    \label{fig:mid}
\end{figure*}

In this section, we provide further results on feature trajectories. This includes direct visualizations of trajectories(~\cref{fig:featuretraj,fig:featuretraj-ex}), detailed PCA variance tables for each layer’s features(~\cref{tab:pca-full}), and more experiments validating FiLM’s corrective power across additional layers and inputs(~\cref{fig:film-extra}). We also provide the training loss curve(~\cref{fig:loss}) and visualization of intermediate training process(~\cref{fig:mid}).

\begin{table}[!htbp]
  \centering

  \begin{minipage}[t]{0.49\textwidth}
    \centering
    \renewcommand{\arraystretch}{0.85}
    \caption{Cumulative explained variance (\%) of trajectory 0.}
    
    \begin{tabular}{lS[table-format=2.2]S[table-format=2.2]S[table-format=2.2]S[table-format=2.2]S[table-format=2.2]}
        \toprule
        \textbf{ Index} & {PC$_1$} & {PC$_{1:2}$} & {PC$_{1:3}$} & {PC$_{1:4}$} & {PC$_{1:5}$} \\
        \midrule
        0  & \textbf{90.13} & 99.33 & 99.82 & 99.94 & 99.98 \\  \midrule
        1  & 73.99 & 88.59 & \textbf{94.93} & 98.36 & 99.29 \\  \midrule
        2  & 67.04 & 82.43 & \textbf{90.70} & 96.40 & 98.30 \\  \midrule
        3  & 59.60 & 78.60 & 88.29 & \textbf{94.59} & 97.48 \\  \midrule
        4  & 53.04 & 75.56 & 86.07 & \textbf{92.05} & 96.24 \\  \midrule
        5  & 51.04 & 72.83 & 84.53 & \textbf{92.07} & 96.18 \\  \midrule
        6  & 48.45 & 70.89 & 83.98 & \textbf{91.69} & 95.92 \\  \midrule
        7  & 48.06 & 70.72 & 83.65 & \textbf{91.71} & 95.61 \\ \midrule
        8  & 46.51 & 69.54 & 83.20 & \textbf{91.22} & 95.13 \\ \midrule
        9  & 44.48 & 67.86 & 81.72 & \textbf{90.07} & 94.47 \\ \midrule
        10 & 45.38 & 70.74 & 83.13 & \textbf{91.21} & 94.79 \\ \midrule
        11 & 46.06 & 70.93 & 83.23 & \textbf{91.11} & 94.78 \\ \midrule
        12 & 45.63 & 70.34 & 82.45 & \textbf{90.45} & 94.41 \\ \midrule
        13 & 44.99 & 69.76 & 81.85 & \textbf{90.07} & 94.26 \\ \midrule
        14 & 42.10 & 68.67 & 83.16 & \textbf{91.31} & 94.92 \\ \midrule
        15 & 40.08 & 65.92 & 80.99 & 89.44 & \textbf{94.00} \\ \midrule
        16 & 44.46 & 69.35 & 81.80 & 89.77 & \textbf{93.77} \\ \midrule
        17 & 41.51 & 69.27 & 83.16 & 89.77 & \textbf{93.50} \\ \midrule
        18 & 47.53 & 70.22 & 82.53 & \textbf{90.01} & 93.67 \\ \midrule
        19 & 47.97 & 71.44 & 82.29 & \textbf{89.97} & 93.67 \\ \midrule
        20 & 47.87 & 70.88 & 81.75 & 89.79 & \textbf{93.39} \\ \midrule
        21 & 47.08 & 69.57 & 81.64 & 89.96 & \textbf{93.39} \\ \midrule
        22 & 47.40 & 68.79 & 81.87 & \textbf{90.12} & 93.54 \\ \midrule
        23 & 45.90 & 66.46 & 80.18 & 88.34 & \textbf{92.70} \\ \midrule
        24 & 45.06 & 65.38 & 79.18 & 87.63 & \textbf{92.32} \\ \midrule
        25 & 44.96 & 64.49 & 79.45 & 87.79 & \textbf{92.58} \\ \midrule
        26 & 42.52 & 63.03 & 78.57 & 87.08 & \textbf{92.35} \\ \midrule
        27 & 44.36 & 64.57 & 79.87 & 87.59 & \textbf{92.78} \\ \midrule
        28 & 42.33 & 62.65 & 79.79 & 87.44 & \textbf{92.67} \\ \midrule
        29 & 44.28 & 68.00 & 80.93 & 88.30 & \textbf{93.40} \\ \midrule
        30 & 44.64 & 70.60 & 81.29 & 89.13 & \textbf{93.55} \\ \midrule
        31 & 50.39 & 75.44 & 84.50 & \textbf{90.69} & 94.21 \\ \midrule
        32 & 53.16 & 75.76 & 85.36 & \textbf{91.48} & 94.68 \\ \midrule
        33 & 74.25 & 89.34 & \textbf{93.85} & 96.68 & 97.95 \\ 
        \bottomrule
    \end{tabular}

  \end{minipage}
  \hfill
  \begin{minipage}[t]{0.49\textwidth}
    \centering
          \caption{Cumulative explained variance (\%) of trajectory 1.}
        \renewcommand{\arraystretch}{0.85}
    \begin{tabular}{lS[table-format=2.2]S[table-format=2.2]S[table-format=2.2]S[table-format=2.2]S[table-format=2.2]}
        \toprule
        \textbf{Index} & {PC$_1$} & {PC$_{1:2}$} & {PC$_{1:3}$} & {PC$_{1:4}$} & {PC$_{1:5}$} \\
        \midrule
        0  & \textbf{90.39} & 98.88 & 99.73 & 99.93 & 99.97 \\ \midrule
        1  & 73.61 & 88.47 & \textbf{94.53} & 98.19 & 99.28 \\ \midrule
        2  & 66.91 & 83.40 & \textbf{91.06} & 96.28 & 98.29 \\ \midrule
        3  & 59.49 & 79.36 & 88.66 & \textbf{93.94} & 97.21 \\ \midrule
        4  & 52.92 & 76.13 & 86.66 & \textbf{92.01} & 95.95 \\ \midrule
        5  & 51.36 & 74.57 & 85.70 & \textbf{91.92} & 95.96 \\ \midrule
        6  & 46.86 & 71.34 & 84.31 & \textbf{91.36} & 95.64 \\ \midrule
        7  & 46.43 & 70.28 & 83.80 & \textbf{91.39} & 95.39 \\ \midrule
        8  & 45.51 & 69.84 & 83.34 & \textbf{91.26} & 95.36 \\ \midrule
        9  & 43.97 & 68.71 & 82.43 & \textbf{90.40} & 94.72 \\ \midrule
        10 & 45.04 & 70.72 & 83.14 & \textbf{91.64} & 95.30 \\ \midrule
        11 & 45.20 & 70.59 & 83.04 & \textbf{91.44} & 95.20 \\ \midrule
        12 & 44.30 & 69.89 & 82.42 & \textbf{90.83} & 94.92 \\ \midrule
        13 & 43.17 & 68.31 & 81.14 & \textbf{90.17} & 94.40 \\ \midrule
        14 & 40.13 & 66.88 & 83.68 & \textbf{91.41} & 94.97 \\ \midrule
        15 & 39.28 & 68.70 & 83.35 & \textbf{90.73} & 94.45 \\ \midrule
        16 & 44.58 & 70.50 & 82.83 & 89.84 & \textbf{94.14} \\ \midrule
        17 & 41.49 & 69.74 & 83.09 & 89.95 & \textbf{93.98} \\ \midrule
        18 & 46.34 & 69.15 & 82.46 & \textbf{90.02} & 93.50 \\ \midrule
        19 & 47.65 & 69.31 & 80.99 & 89.49 & \textbf{93.56} \\ \midrule
        20 & 46.76 & 68.38 & 80.85 & 88.94 & \textbf{93.28} \\ \midrule
        21 & 47.33 & 68.20 & 81.41 & 89.41 & \textbf{93.49} \\ \midrule
        22 & 47.84 & 66.40 & 81.53 & 89.44 & \textbf{93.34} \\ \midrule
        23 & 47.95 & 66.31 & 80.31 & 88.30 & \textbf{92.80} \\ \midrule
        24 & 46.68 & 65.11 & 80.14 & 87.94 & \textbf{92.78} \\ \midrule
        25 & 45.86 & 64.80 & 80.36 & 88.21 & \textbf{92.99} \\ \midrule
        26 & 44.01 & 66.16 & 80.80 & 88.43 & \textbf{93.15} \\ \midrule
        27 & 43.21 & 66.09 & 81.13 & 88.47 & \textbf{93.21} \\ \midrule
        28 & 42.71 & 65.96 & 81.34 & 88.95 & \textbf{93.34} \\ \midrule
        29 & 44.80 & 71.35 & 82.82 & \textbf{90.14} & 93.87 \\ \midrule
        30 & 45.62 & 72.61 & 82.64 & \textbf{90.30} & 93.95 \\ \midrule
        31 & 50.74 & 75.26 & 85.46 & \textbf{91.10} & 94.94 \\ \midrule
        32 & 53.38 & 76.02 & 85.12 & \textbf{91.42} & 94.71 \\ \midrule
        33 & 73.36 & 88.76 & \textbf{93.23} & 96.38 & 97.68 \\
        \bottomrule
    \end{tabular}
    \label{tab:pca-full}

  \end{minipage}
\end{table}

\begin{figure*}[!htbp]
    \centering

    \begin{subfigure}[h]{0.25\textwidth}
        \centering
        \includegraphics[width=\textwidth]{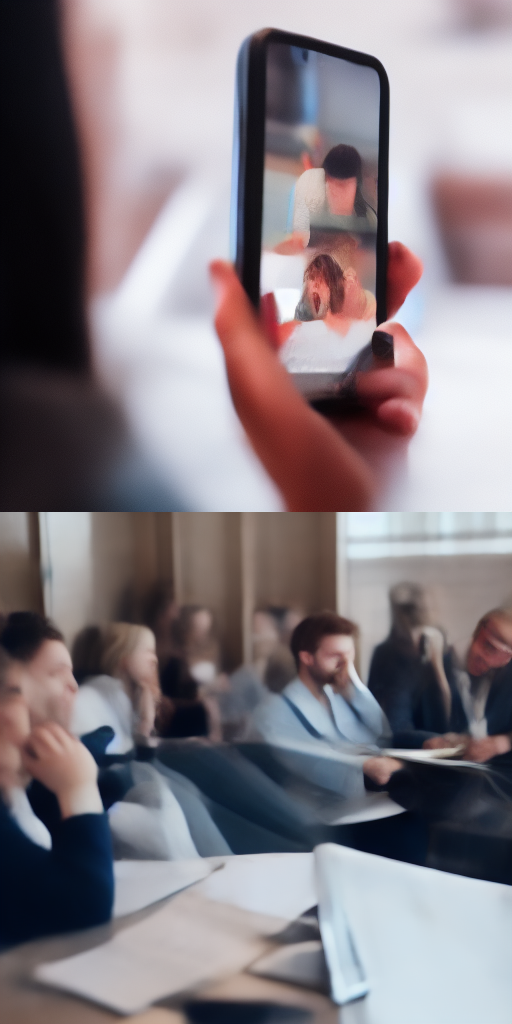}
        \caption{DDIM, 4 steps}
    \end{subfigure}
    \hfill
    \begin{subfigure}[h]{0.25\textwidth}
        \centering
        \includegraphics[width=\textwidth]{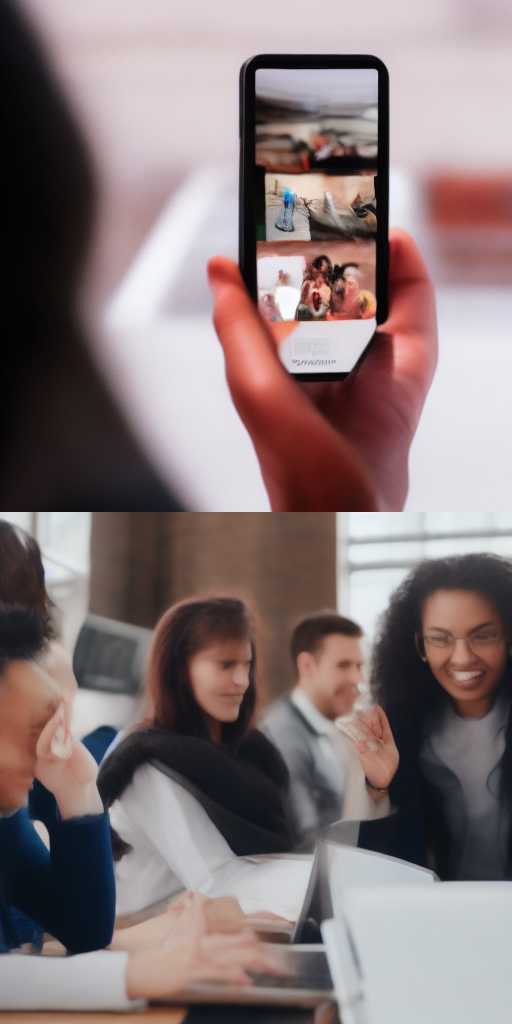}
        \caption{DDIM, 7 steps}
    \end{subfigure}
    \hfill
    \begin{subfigure}[h]{0.25\textwidth}
        \centering
        \includegraphics[width=\textwidth]{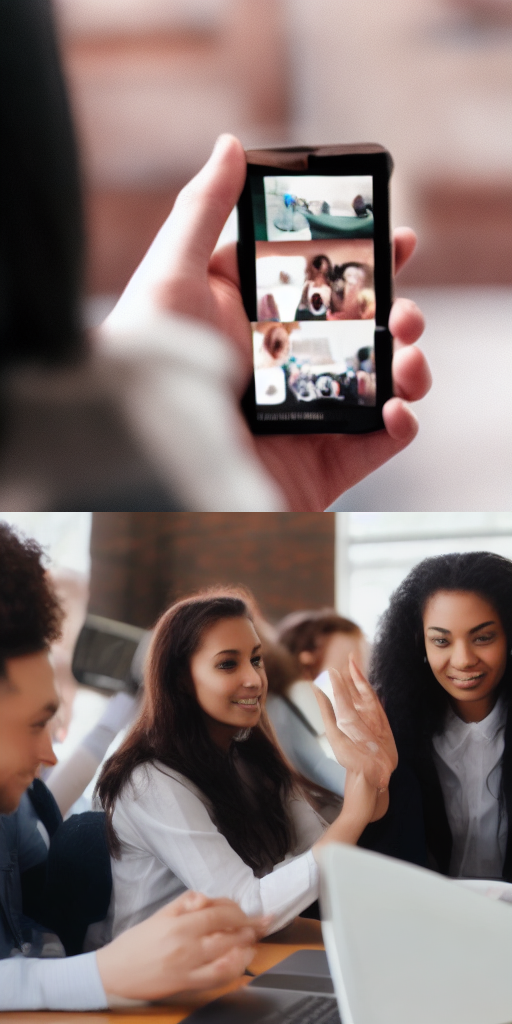}
        \caption{DDIM, 11 steps}
    \end{subfigure}

    \caption{The phenomenon of mode non-convergence in Stable Diffusion sampling. As shown in the figure, the sampling result at 4 steps exhibits a pattern that is completely different from that at 11 steps, while the result at 7 steps demonstrates an intermediate transition.}
    \label{fig:mode-un}
\end{figure*}

\clearpage
\section{Visualization}

\begin{figure*}[!htbp]

    \centering
    
    \begin{subfigure}[b]{0.45\textwidth}
        \centering
        \includegraphics[width=\textwidth]{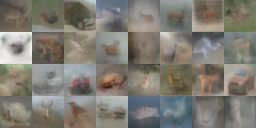}
        \caption{DDIM}
    \end{subfigure}
    \hfill
    \begin{subfigure}[b]{0.45\textwidth}
        \centering
        \includegraphics[width=\textwidth]{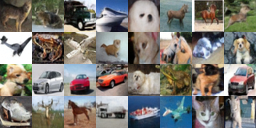}
        \caption{DDIM+METO}
    \end{subfigure}
    \hfill
    
    \vskip\baselineskip 
    \begin{subfigure}[b]{0.45\textwidth}
        \centering
        \includegraphics[width=\textwidth]{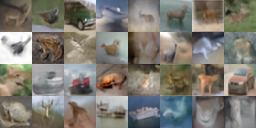}
        \caption{iPNDM}
    \end{subfigure}
    \hfill
    \begin{subfigure}[b]{0.45\textwidth}
        \centering
        \includegraphics[width=\textwidth]{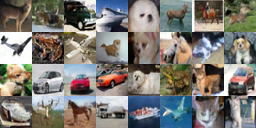}
        \caption{iPNDM+METO}
    \end{subfigure}
    \hfill

    \caption{Samples on CIFAR10 32×32 with 3 NFE.}
 
\end{figure*}

\begin{figure*}[!htbp]

    \centering
    \begin{subfigure}[b]{0.45\textwidth}
        \centering
        \includegraphics[width=\textwidth]{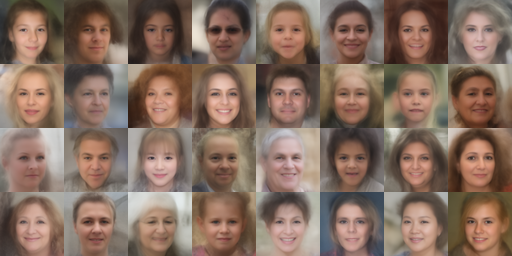}
        \caption{DDIM}
    \end{subfigure}
    \hfill
    \begin{subfigure}[b]{0.45\textwidth}
        \centering
        \includegraphics[width=\textwidth]{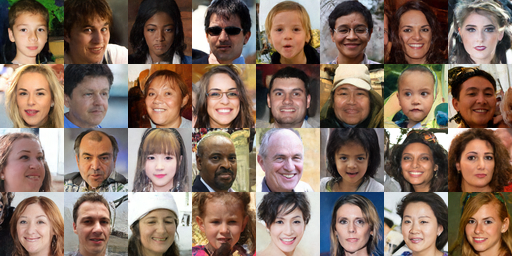}
        \caption{DDIM+METO}
    \end{subfigure}
    \hfill
    
    \vskip\baselineskip 
    \begin{subfigure}[b]{0.45\textwidth}
        \centering
        \includegraphics[width=\textwidth]{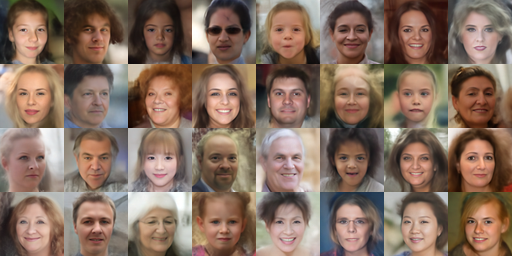}
        \caption{iPNDM}
    \end{subfigure}
    \hfill
    \begin{subfigure}[b]{0.45\textwidth}
        \centering
        \includegraphics[width=\textwidth]{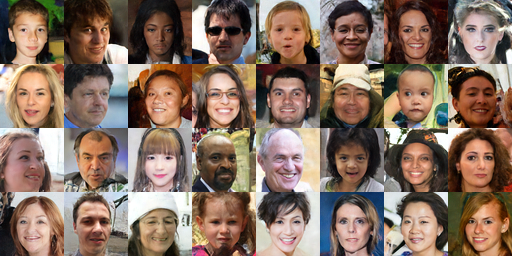}
        \caption{iPNDM+METO}
    \end{subfigure}
    \hfill

    \caption{Samples on FFHQ 64×64 with 3 NFE.}
\end{figure*}

\begin{figure*}[!htbp]

    \centering
    \begin{subfigure}[b]{0.43\textwidth}
        \centering
        \includegraphics[width=\textwidth]{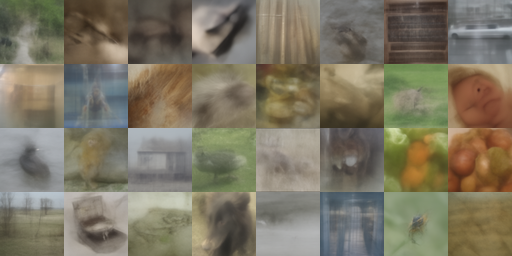}
        \caption{DDIM}
    \end{subfigure}
    \hfill
    \begin{subfigure}[b]{0.43\textwidth}
        \centering
        \includegraphics[width=\textwidth]{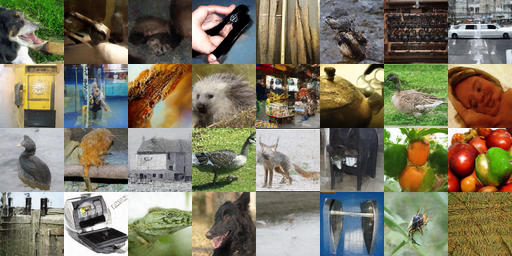}
        \caption{DDIM+METO}
    \end{subfigure}
    \hfill
    
    \vskip\baselineskip 
    \begin{subfigure}[b]{0.43\textwidth}
        \centering
        \includegraphics[width=\textwidth]{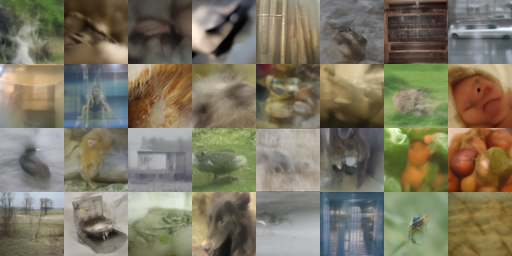}
        \caption{iPNDM}
    \end{subfigure}
    \hfill
    \begin{subfigure}[b]{0.43\textwidth}
        \centering
        \includegraphics[width=\textwidth]{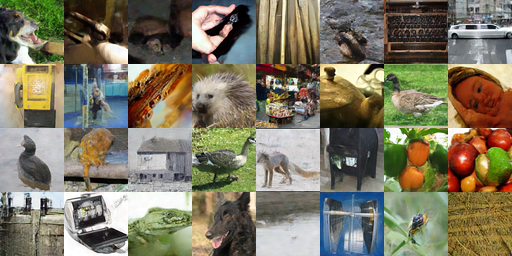}
        \caption{iPNDM+METO}
    \end{subfigure}
    \hfill

    \caption{Samples on ImageNet 64×64 with 3 NFE.}
\end{figure*}

\begin{figure*}[!htbp]

    \centering
    \begin{subfigure}[b]{0.43\textwidth}
        \centering
        \includegraphics[width=\textwidth]{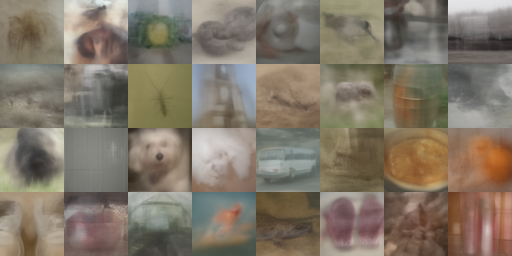}
        \caption{DDIM}
    \end{subfigure}
    \hfill
    \begin{subfigure}[b]{0.43\textwidth}
        \centering
        \includegraphics[width=\textwidth]{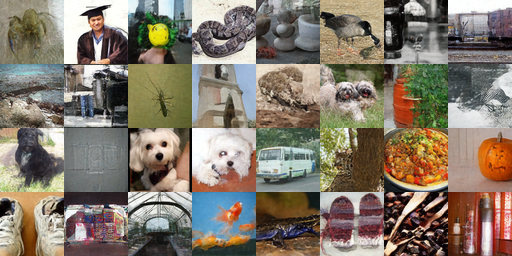}
        \caption{DDIM+METO}
    \end{subfigure}
    \hfill
    
    \vskip\baselineskip 
    \begin{subfigure}[b]{0.43\textwidth}
        \centering
        \includegraphics[width=\textwidth]{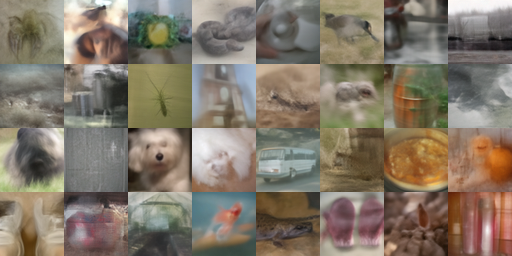}
        \caption{iPNDM}
    \end{subfigure}
    \hfill
    \begin{subfigure}[b]{0.43\textwidth}
        \centering
        \includegraphics[width=\textwidth]{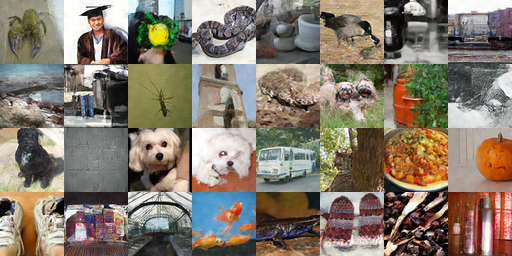}
        \caption{iPNDM+METO}
    \end{subfigure}
    \hfill

    \caption{Samples on ImageNet 64×64 with 3 NFE.}
\end{figure*}

\begin{figure*}[!htbp]

    \centering
    \begin{subfigure}[b]{0.45\textwidth}
        \centering
        \includegraphics[width=\textwidth]{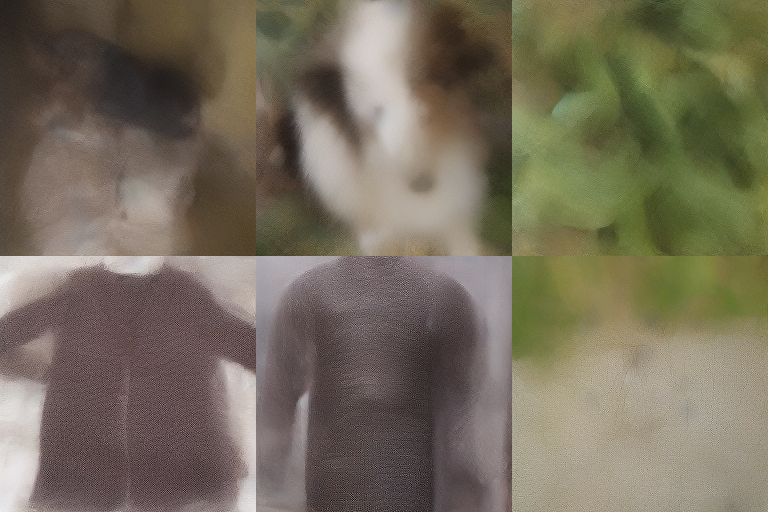}
        \caption{DDIM}
    \end{subfigure}
    \hfill
    \begin{subfigure}[b]{0.45\textwidth}
        \centering
        \includegraphics[width=\textwidth]{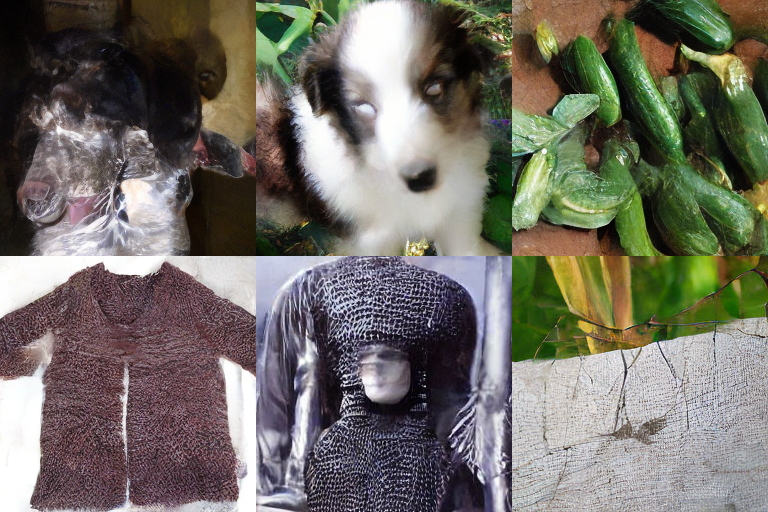}
        \caption{DDIM+METO}
    \end{subfigure}
    \hfill
    
    \vskip\baselineskip 
    \begin{subfigure}[b]{0.45\textwidth}
        \centering
        \includegraphics[width=\textwidth]{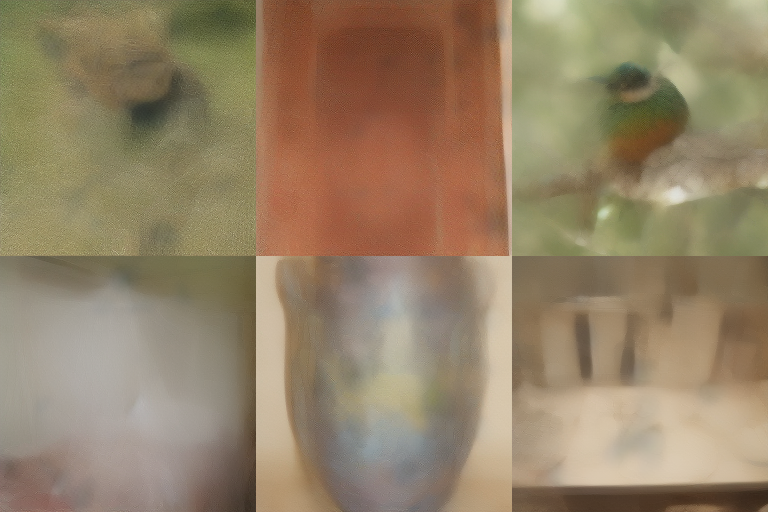}
        \caption{DDIM}
    \end{subfigure}
    \hfill
    \begin{subfigure}[b]{0.45\textwidth}
        \centering
        \includegraphics[width=\textwidth]{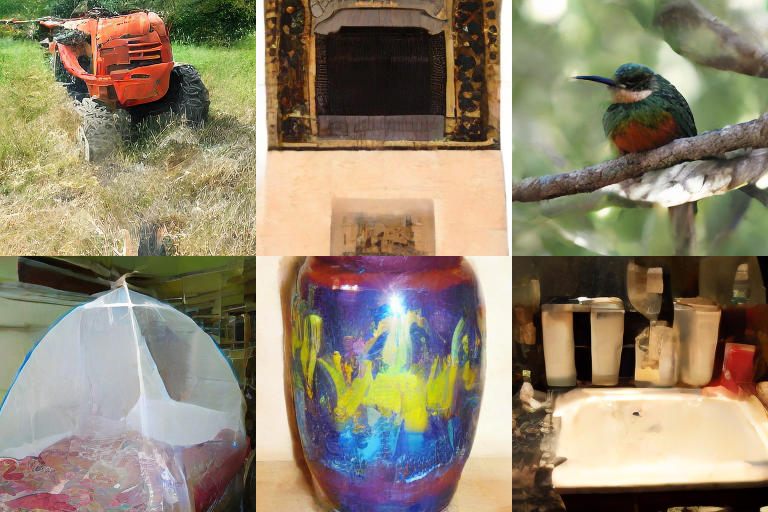}
        \caption{DDIM+METO}
    \end{subfigure}
    \hfill

    \caption{Samples on ImageNet-256 256×256 with 3 NFE.}
\end{figure*}

\begin{figure*}[!htbp]

    \centering
    \begin{subfigure}[b]{0.45\textwidth}
        \centering
        \includegraphics[width=\textwidth]{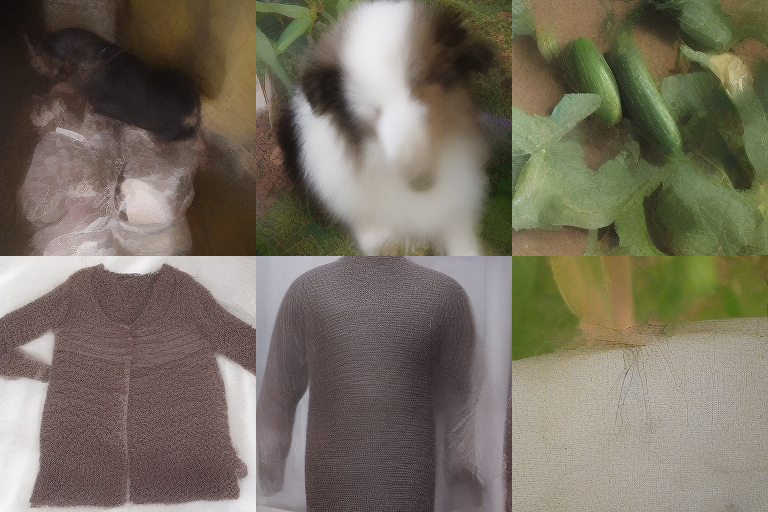}
        \caption{DDIM}
    \end{subfigure}
    \hfill
    \begin{subfigure}[b]{0.45\textwidth}
        \centering
        \includegraphics[width=\textwidth]{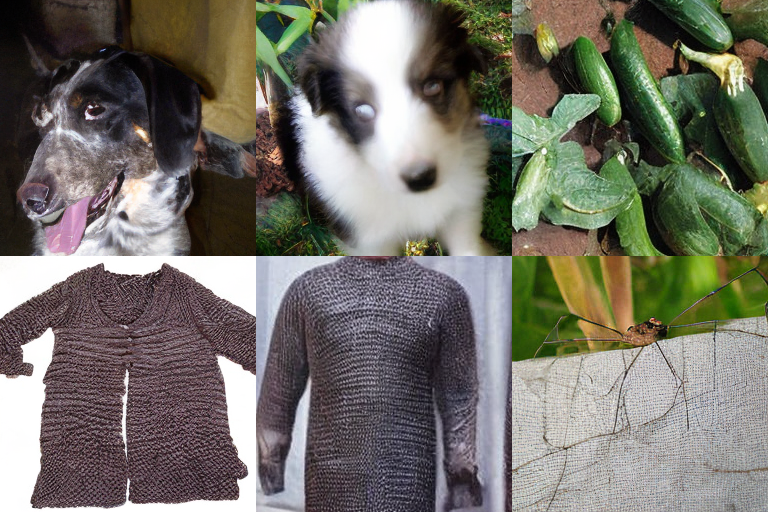}
        \caption{DDIM+METO}
    \end{subfigure}
    \hfill
    
    \vskip\baselineskip 
    \begin{subfigure}[b]{0.45\textwidth}
        \centering
        \includegraphics[width=\textwidth]{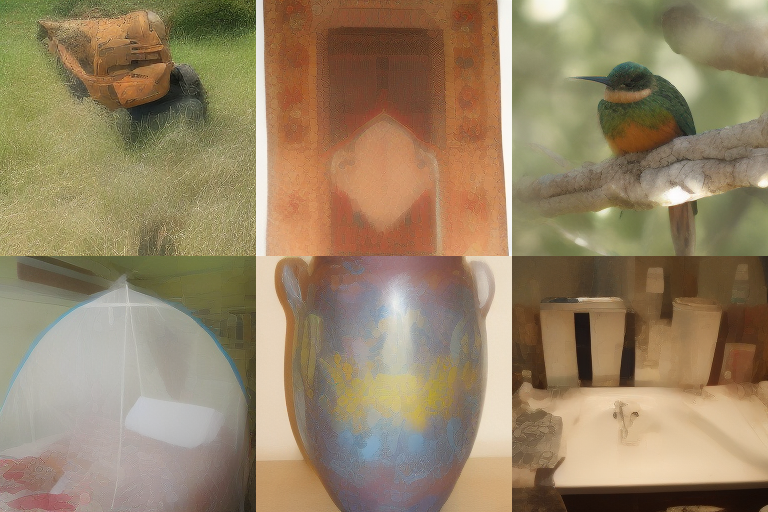}
        \caption{DDIM}
    \end{subfigure}
    \hfill
    \begin{subfigure}[b]{0.45\textwidth}
        \centering
        \includegraphics[width=\textwidth]{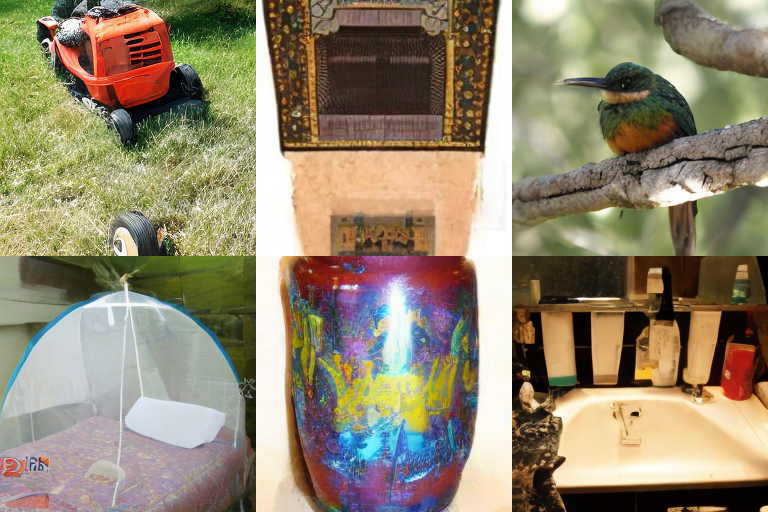}
        \caption{DDIM+METO}
    \end{subfigure}
    \hfill

    \caption{Samples on ImageNet-256 256×256 with 6 NFE.}
\end{figure*}

\begin{figure*}[!htbp]

    \centering
    \begin{subfigure}[b]{0.42\textwidth}
        \centering
        \includegraphics[width=\textwidth]{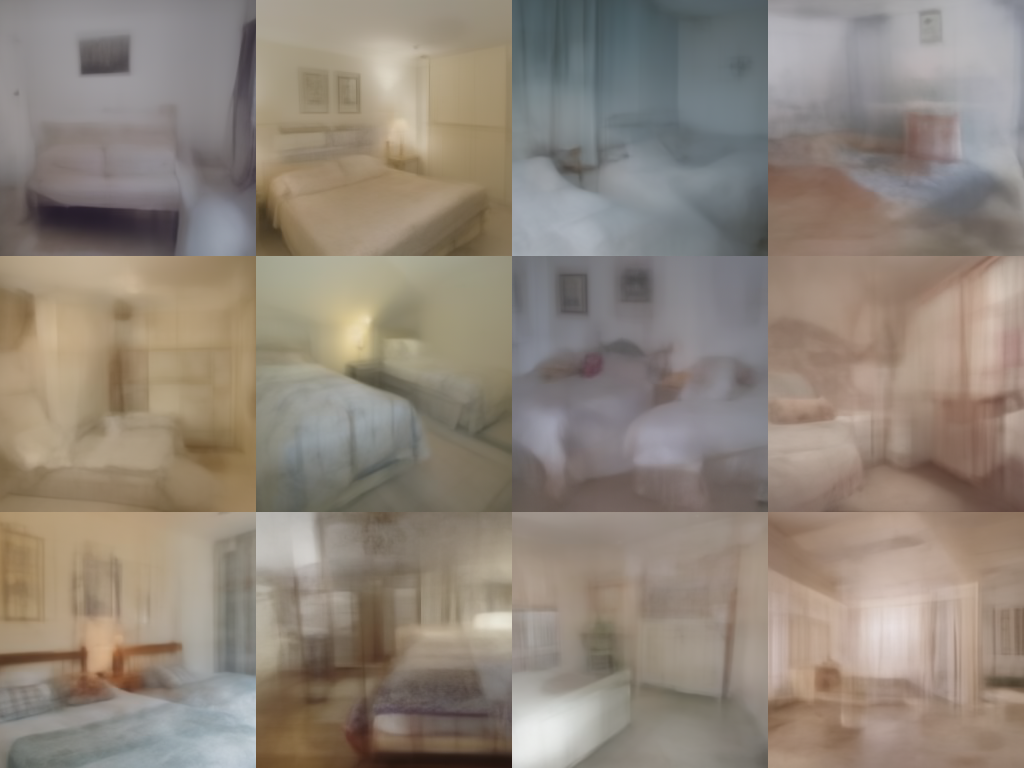}
        \caption{DDIM}
    \end{subfigure}
    \hfill
    \begin{subfigure}[b]{0.42\textwidth}
        \centering
        \includegraphics[width=\textwidth]{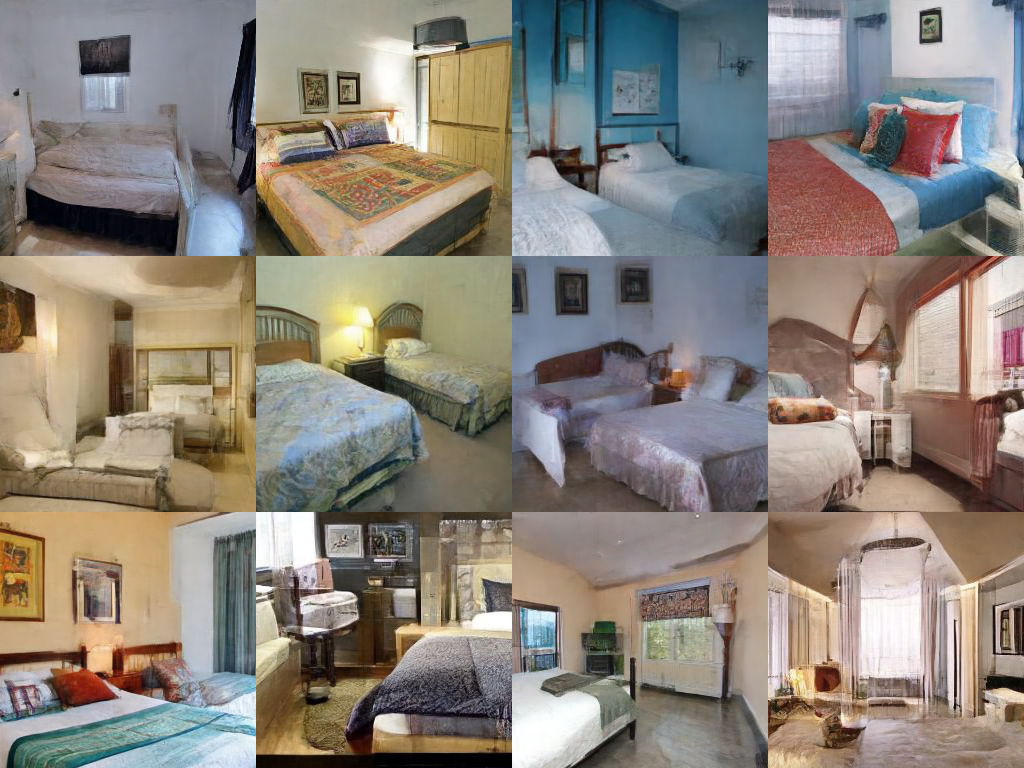}
        \caption{DDIM+METO}
    \end{subfigure}
    \hfill
    
    \vskip\baselineskip 
    \begin{subfigure}[b]{0.42\textwidth}
        \centering
        \includegraphics[width=\textwidth]{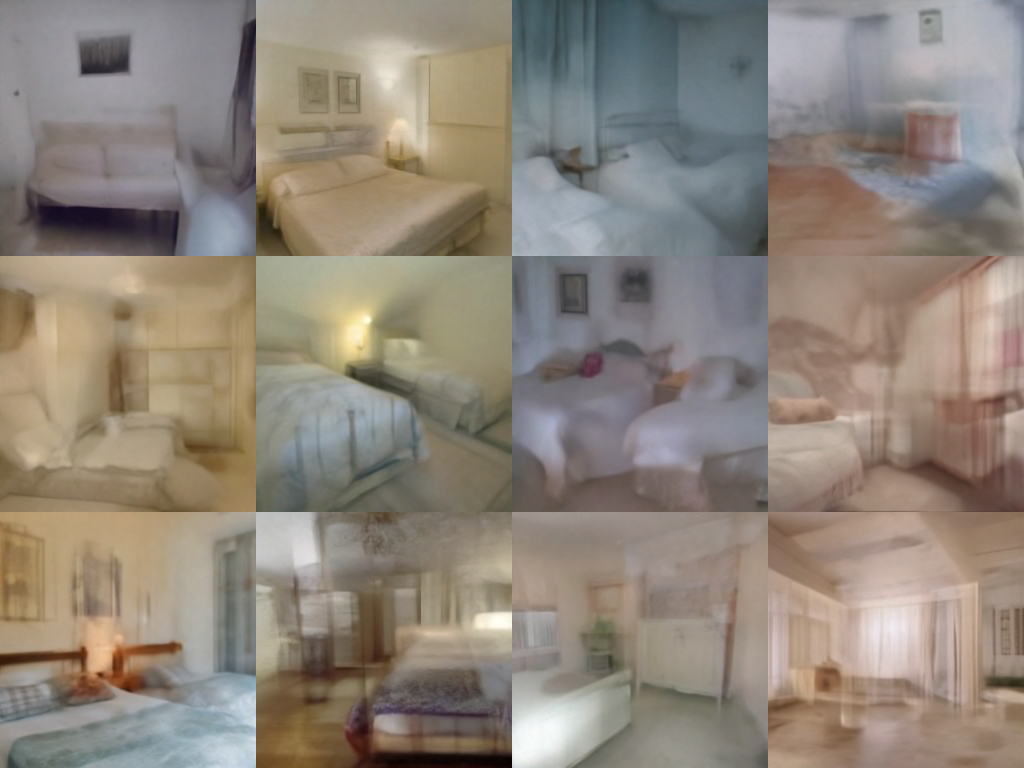}
        \caption{iPNDM}
    \end{subfigure}
    \hfill
    \begin{subfigure}[b]{0.42\textwidth}
        \centering
        \includegraphics[width=\textwidth]{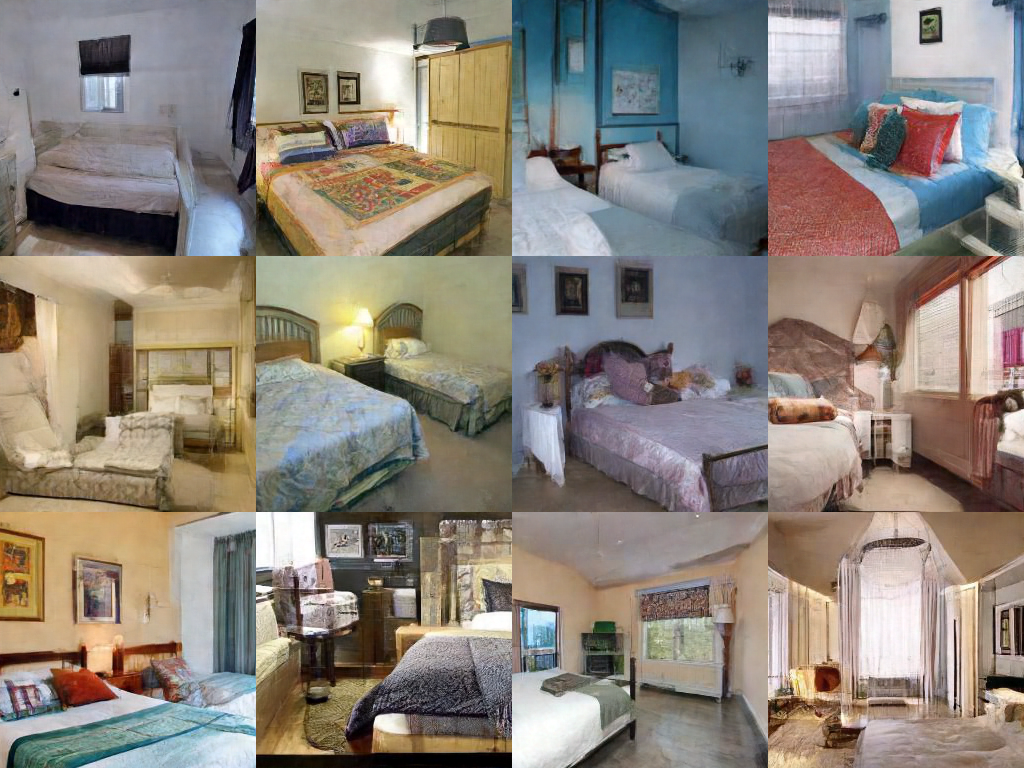}
        \caption{iPNDM+METO}
    \end{subfigure}
    \hfill

    \caption{Samples on LSUN Bedroom 256×256 with 3 NFE.}
\end{figure*}

\begin{figure*}[!htbp]

    \centering
    \begin{subfigure}[b]{0.45\textwidth}
        \centering
        \includegraphics[width=\textwidth]{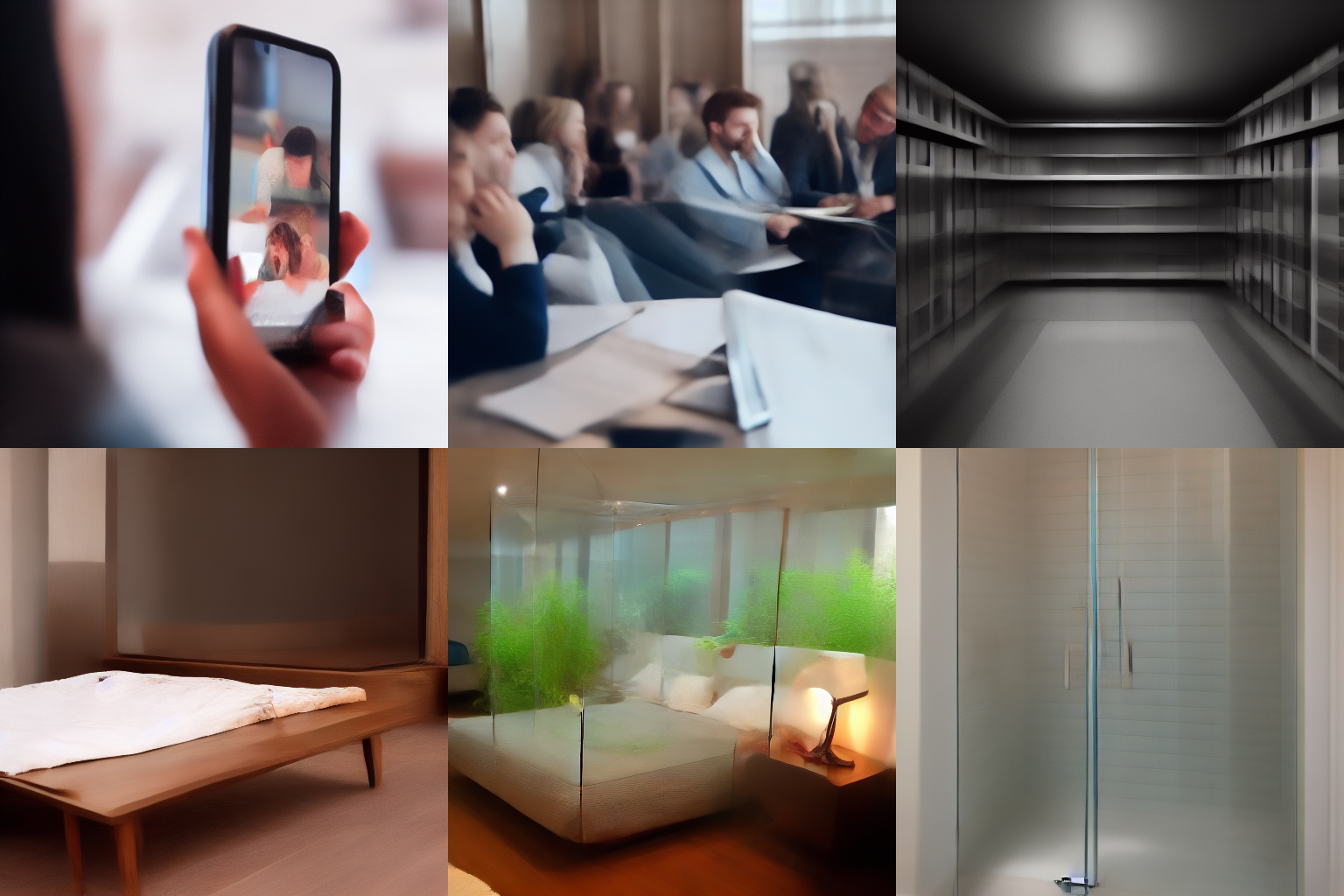}
        \caption{DDIM}
    \end{subfigure}
    \hfill
    \begin{subfigure}[b]{0.45\textwidth}
        \centering
        \includegraphics[width=\textwidth]{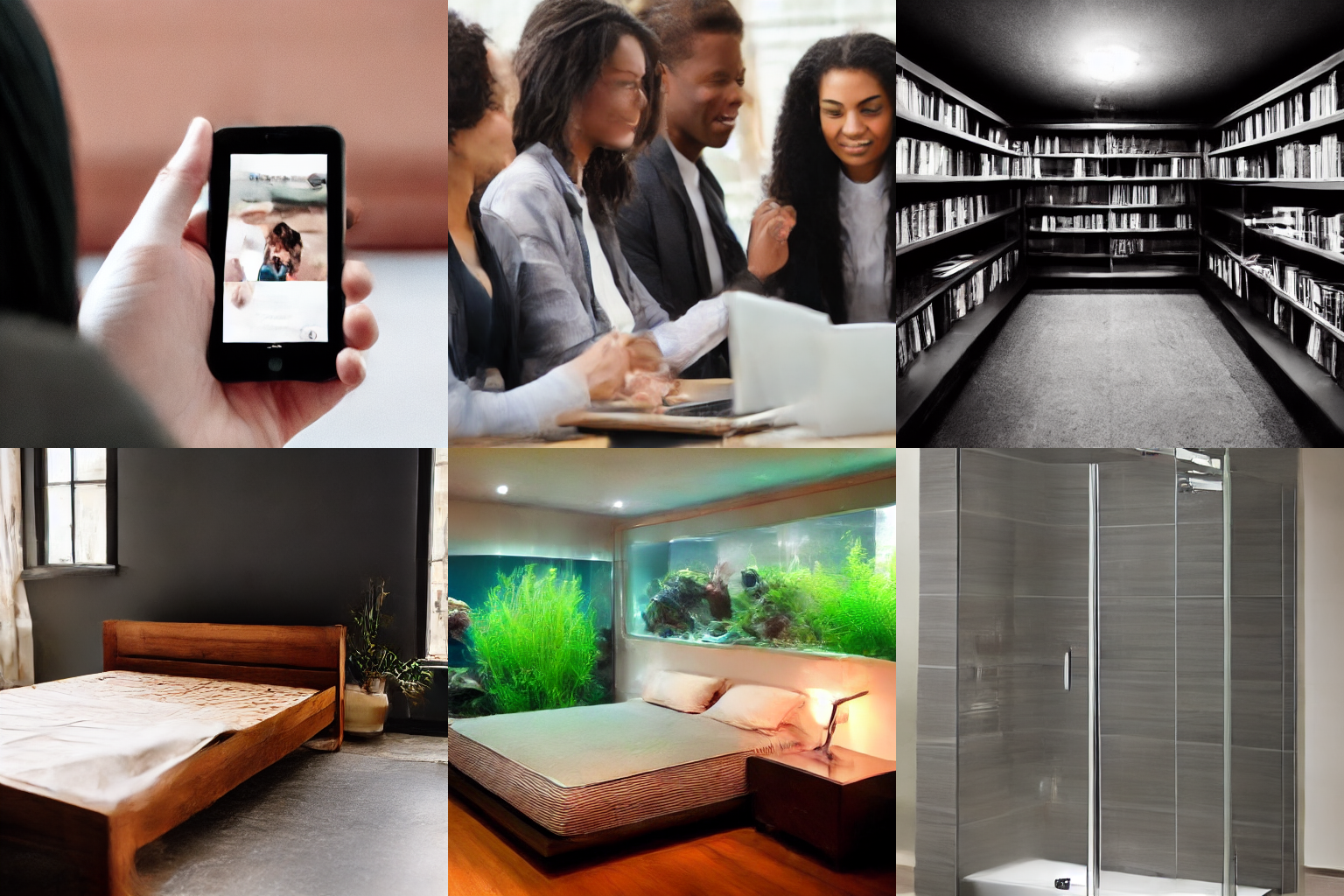}
        \caption{DDIM+METO}
    \end{subfigure}
    \hfill
    
    \vskip\baselineskip 
    \begin{subfigure}[b]{0.45\textwidth}
        \centering
        \includegraphics[width=\textwidth]{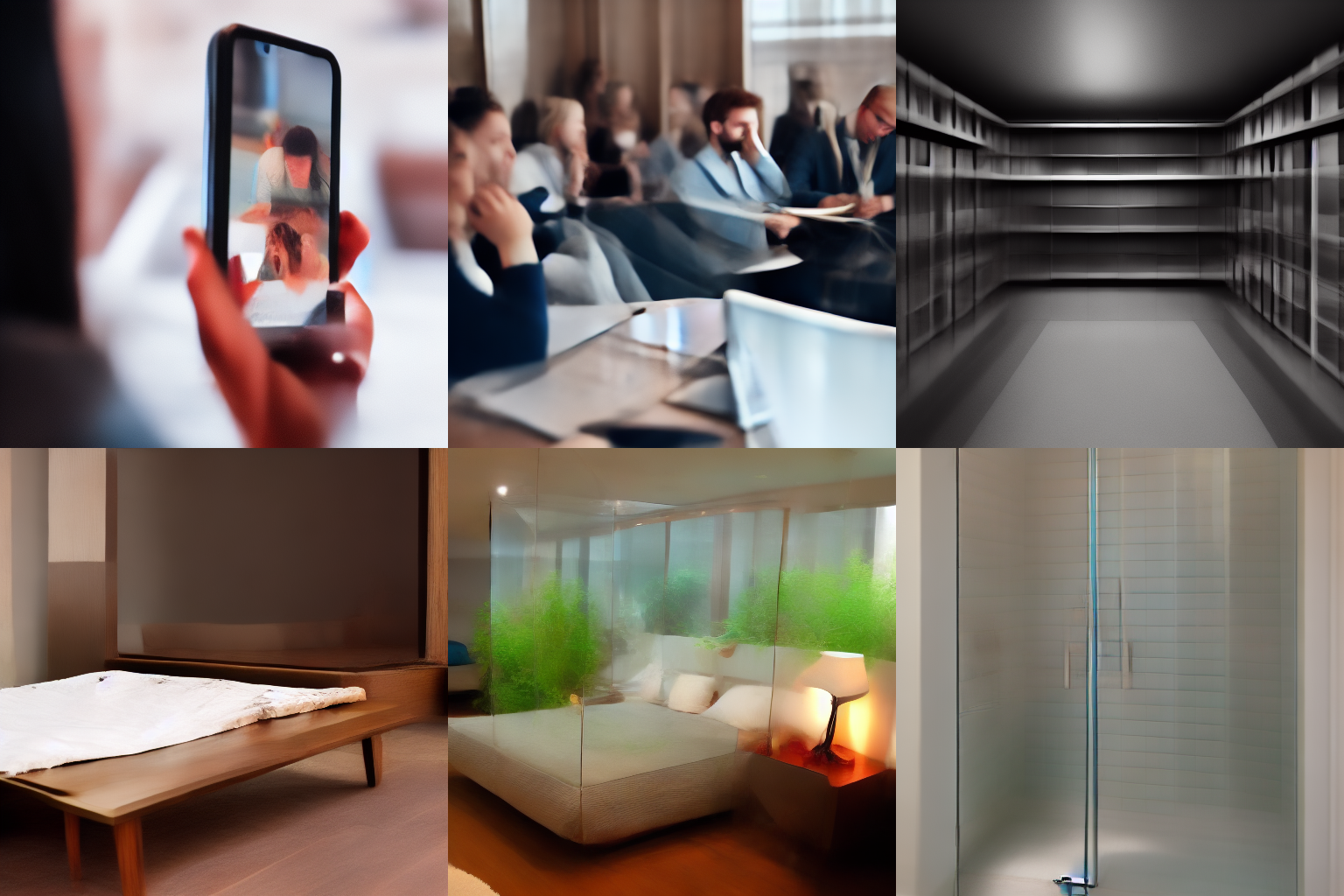}
        \caption{DPM++(2M)}
    \end{subfigure}
    \hfill
    \begin{subfigure}[b]{0.45\textwidth}
        \centering
        \includegraphics[width=\textwidth]{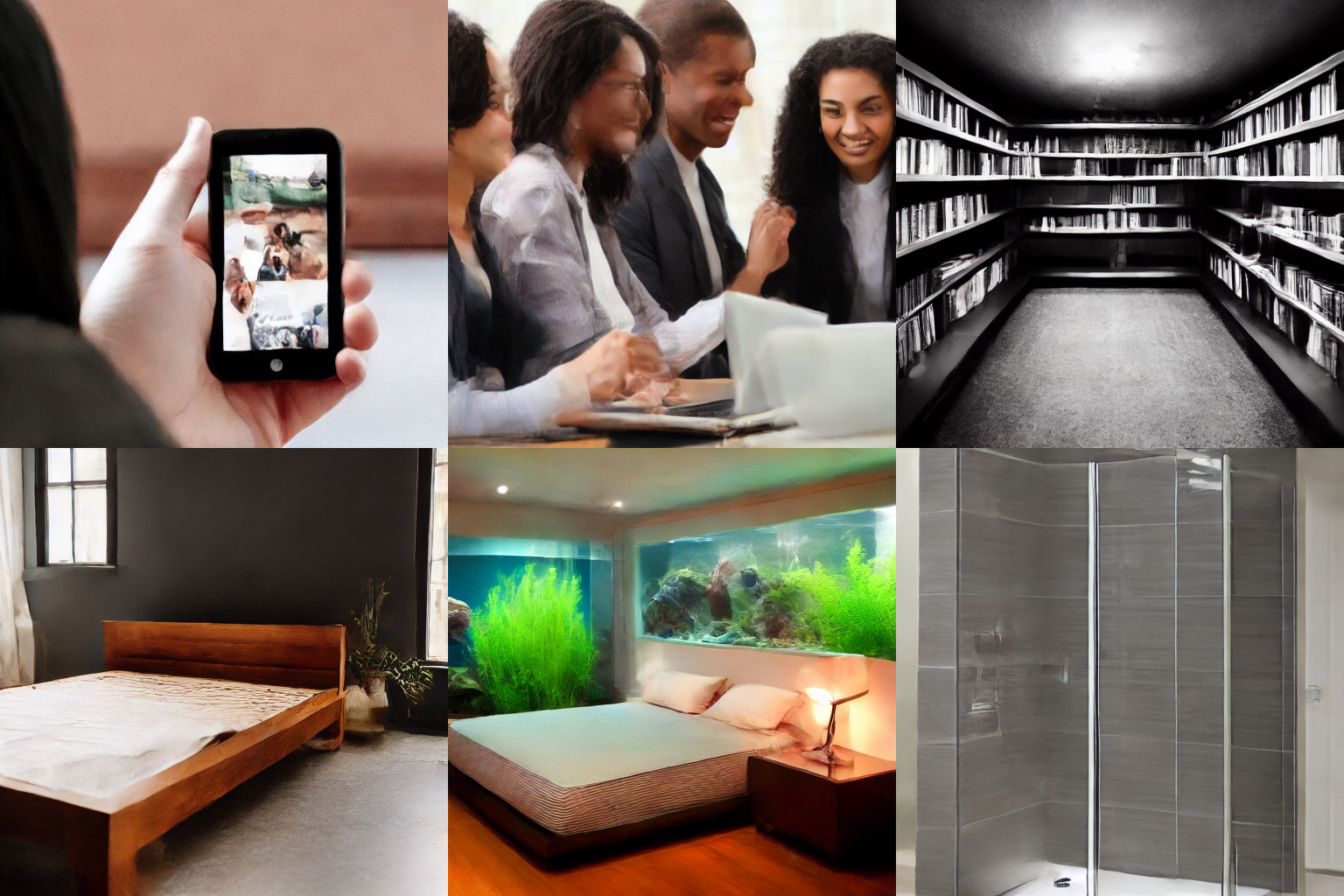}
        \caption{DPM++(2M)+METO}
    \end{subfigure}
    \hfill

    \caption{Samples on MS COCO 512×512 with 3 NFE.}
\end{figure*}


\end{document}